\documentclass[lettersize,journal]{IEEEtran}
\usepackage{amsmath,amsfonts,amssymb}
\usepackage{algorithmic}
\usepackage{algorithm}
\usepackage{array}
\usepackage[caption=false,font=normalsize,labelfont=sf,textfont=sf]{subfig}
\usepackage{textcomp}
\usepackage{stfloats}
\usepackage{url}
\usepackage{verbatim}
\usepackage{graphicx}
\usepackage{siunitx}
\usepackage{gensymb}
\usepackage{cite}
\usepackage{booktabs}
\usepackage{multirow}
\usepackage{pifont} 
\usepackage[hidelinks]{hyperref} % remove the color
\usepackage{dsfont}

\usepackage{listings}
\usepackage{xcolor}

\usepackage{tikz}

\usepackage{makecell}

\newcommand{\newcontent}[1]{\textcolor{black}{#1}} % red

\begin{document}

\title{A Comprehensive Survey and Systematic Real-World Evaluation of Embodied Vision-and-Language Navigation}

\author{
    Liuyi Wang,~\IEEEmembership{Student Member,~IEEE}, Kai Sheng, Zongtao He, Jinlong Li, Yongrui Qin, Haojie Dai, \\ Xiangyi Wang,
    Jingwei Yang, Qingqing Yan, Chengju Liu$^\dagger$,~\IEEEmembership{Member,~IEEE}, Qijun Chen$^\dagger$,~\IEEEmembership{Senior Member,~IEEE}
        % <-this % stops a space
\thanks{Authors are with College of Electronic and Information Engineering, Tongji University, Shanghai, China. (Email: wly@tongji.edu.cn, qjchen@tongji.edu.cn). This paper is supported by the National Natural Science Foundation of China under Grants (62233013, 62473295, 62333017, 624B2105). $^\dagger$ presents the corresponding author.}
}% <-this % stops a space

% The paper headers
\markboth{Journal of \LaTeX\ Class Files,~Vol.~14, No.~8, August~2021}%
{Shell \MakeLowercase{\textit{et al.}}: A Sample Article Using IEEEtran.cls for IEEE Journals}

% \IEEEpubid{0000--0000/00\$00.00~\copyright~2021 IEEE}
% Remember, if you use this you must call \IEEEpubidadjcol in the second
% column for its text to clear the IEEEpubid mark.

\maketitle

\begin{abstract}

    Navigation is a fundamental capability of autonomous systems, yet most existing approaches rely on highly structured models and strong prior assumptions, limiting their robustness in open and uncertain real-world environments. Vision-and-Language Navigation (VLN) offers a promising direction by enabling robots to integrate natural language understanding with visual perception in a data-driven manner. Although VLN has attracted increasing research attention, systematic methodological taxonomy and real-world validation remain limited. This survey presents a comprehensive review of VLN research. Specifically, state-of-the-art methods are organized along two orthogonal dimensions: action paradigms, including hierarchical and monolithic frameworks, and model paradigms, including discriminative and generative approaches. A critical analysis of their respective strengths and limitations is provided. Additionally, \newcontent{we conduct a systematic real-world evaluation of representative VLN system configurations on a physical robotic platform.} Experiments across ten diverse real-world scenes \newcontent{show a substantial performance gap between simulation and real-world deployment under the tested configurations}: a representative monolithic RGB-only method achieves 61\% success in simulation but drops to 22\% in real-world deployment, while a hierarchical framework \newcontent{achieves a higher real-world success rate of 51\%, suggesting stronger robustness in our evaluation setting.} 
    Finally, we highlight key challenges in perception, decision-making, and control that must be addressed in future research. 
\end{abstract}


\setlength{\textfloatsep}{5pt}
\setlength{\abovecaptionskip}{5pt}
\setlength{\belowcaptionskip}{5pt}

\setlength{\abovedisplayskip}{3pt}
\setlength{\belowdisplayskip}{3pt}

\begin{quote}
    \emph{``For the things we have to learn before we can do them, we learn by doing them.''} \\
    \null\hfill --- Aristotle
    \end{quote}

\section{Introduction}
\label{sec_introduction}

\begin{figure}[t]
    \centering
    \includegraphics[width=\linewidth]{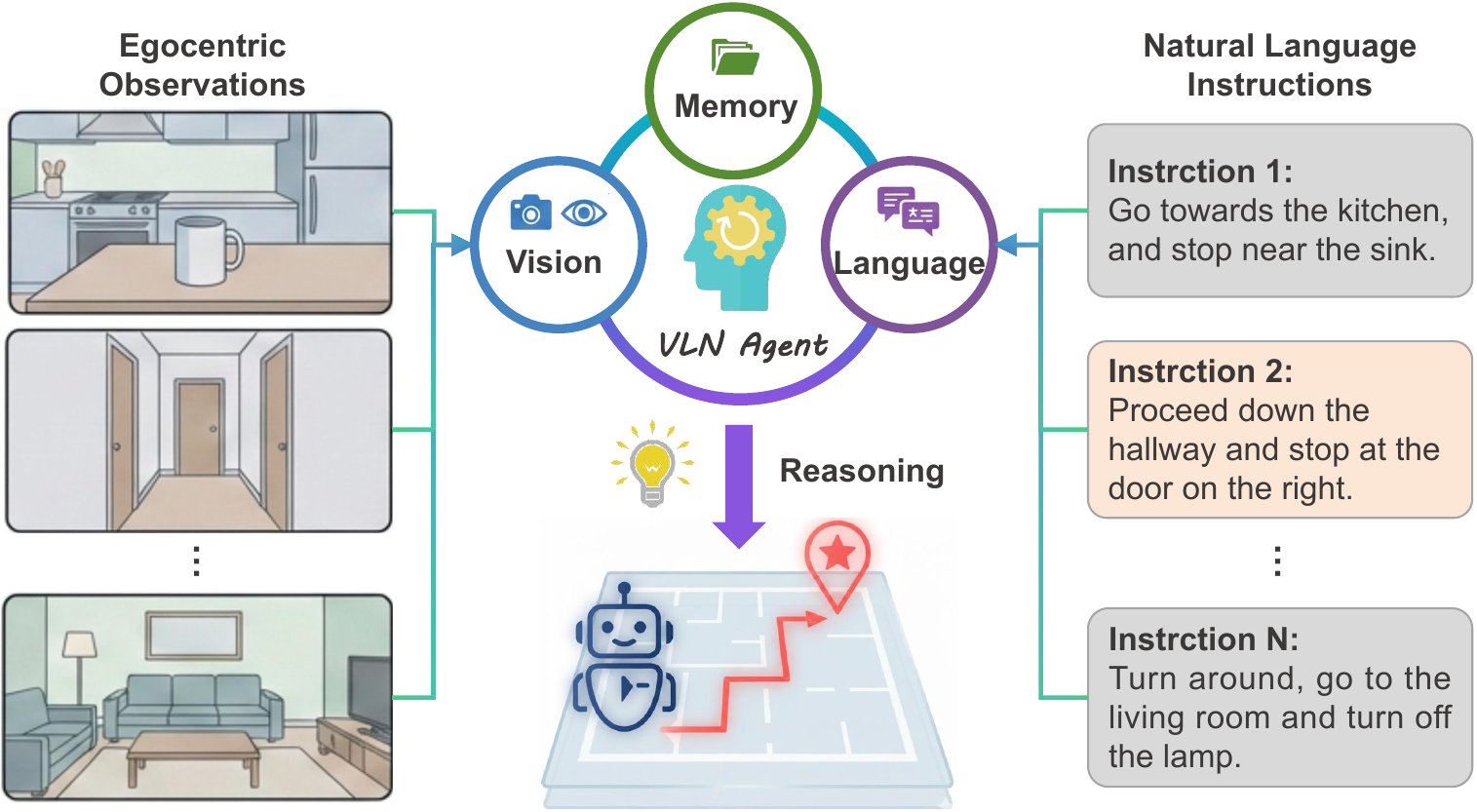}
    \caption{Overview of the vision-and-language navigation (VLN) research.}
    \label{fig_overview}
\end{figure}

\IEEEPARstart{N}{avigation} is a foundational capability of autonomous systems. Most existing navigation approaches adopt highly model-driven paradigms that rely on strong prior assumptions about environmental structure and task objectives~\cite{Smith1986,Smith1990,1315094,orb_slam2}, which fundamentally limit their ability to generalize to open-ended scenarios and cope with complex uncertainty. One particularly promising direction draws inspiration from human navigation, which integrates visual perception and natural language to reason and act under uncertainty. 
Given a natural language instruction such as \textit{“go towards the kitchen, and stop near the sink,”} Vision-and-Language Navigation (VLN)~\cite{anderson2018vision} formalizes this problem by requiring an agent to follow free-form language instructions to navigate previously unseen, mapless environments using only egocentric visual inputs (Fig.~\ref{fig_overview}).
By enabling natural interaction and embodying core elements of human intelligence, vision-language fusion establishes a principled foundation for robust operation in diverse and unstructured environments.

\begin{figure}[t]
    \centering
    \includegraphics[width=\linewidth]{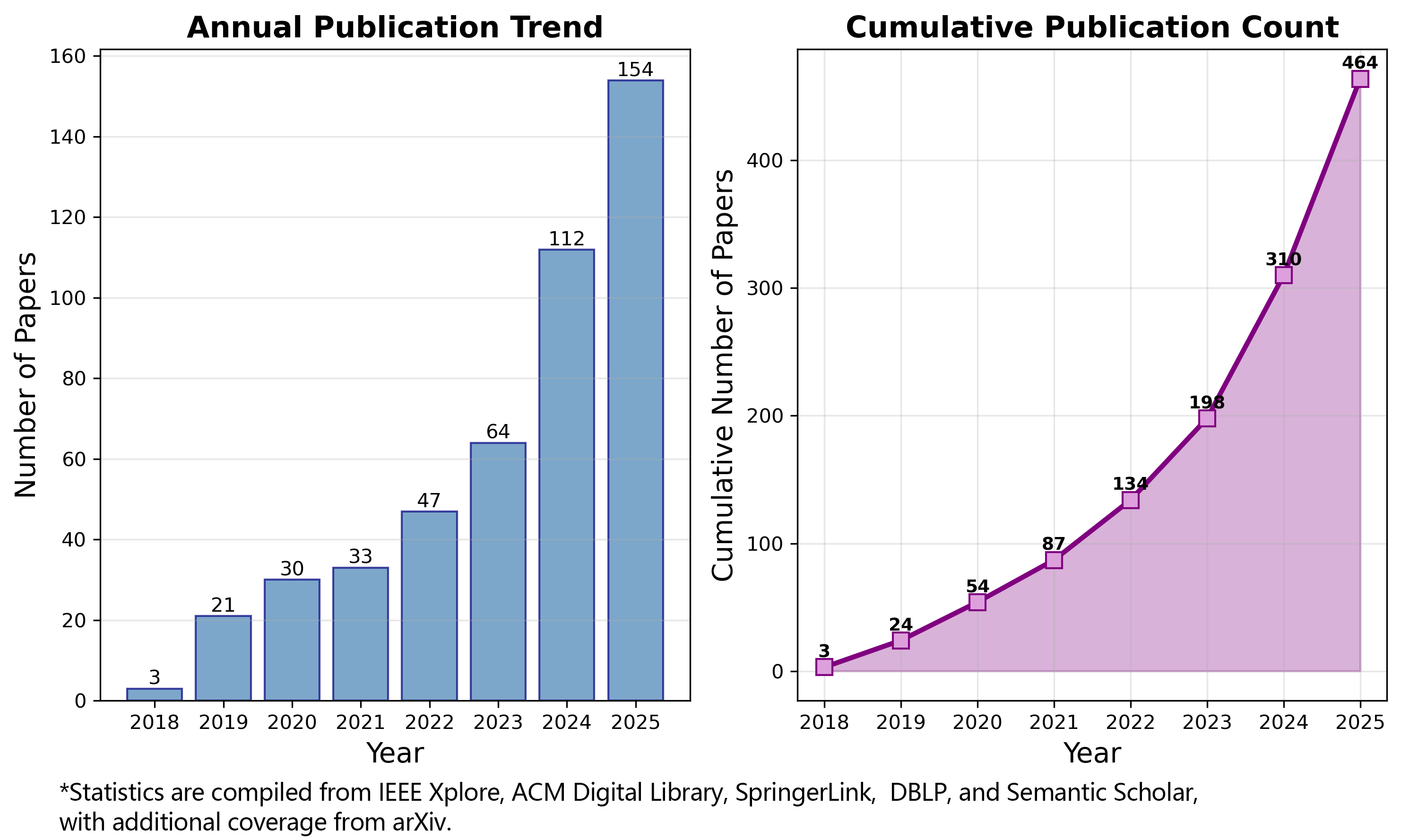}
    \caption{Trend of vision-and-language navigation (VLN) publications from 2018 to 2025.}
    \label{fig_publications}
\end{figure}

Fig.~\ref{fig_publications} illustrates the rapid growth of VLN research over recent years, reflecting the field's increasing vitality and expanding research community. Over the past several years, the VLN community has witnessed remarkable progress, driven by the development of large-scale datasets~\cite{anderson2018vision,jain2019stay,ku2020room,krantz_beyond_2020,song2025towards}, high-fidelity simulators~\cite{anderson2018vision,szot2021habitat,kolve2017ai2thor,li2024behavior,wang2025rethinking}, and increasingly sophisticated algorithms leveraging deep learning~\cite{fried2018speaker,tan2019learning,hong2021vln,chen2022think,wang2025g3d}, reinforcement learning~\cite{krantz_beyond_2020,qi2025vln,tan2025source,Xu_2025_ICCV}, and more recently, large foundation models~\cite{zhang2024navid,zhang2024uni,cheng2024navila,wei2025streamvln,wang2025clash,zeng2025janusvln}. These advances have yielded substantial performance gains on established VLN benchmarks, such as Room-to-Room (R2R)~\cite{anderson2018vision} and its continuous counterpart VLN-CE~\cite{krantz_beyond_2020}, with state-of-the-art (SoTA) methods achieving success rates (SR) exceeding 85\% in idealized simulated unseen environments~\cite{wangbootstrapping}.

Over time, VLN research has converged toward two principal action-space architectures: hierarchical and monolithic frameworks. Hierarchical approaches decompose navigation into high-level waypoint planning and low-level control~\cite{krantz2021waypoint,chen2022think,an2024etpnav,wang2025clash,internvla-n1}, whereas monolithic approaches model an end-to-end mapping from perceptual observations to actions within a unified architecture~\cite{krantz_beyond_2020,he2024mee,zhang2024navid,zhang2024uni,cheng2024navila}.
Orthogonal to action-space design, existing methods can also be categorized by their model paradigms. Discriminative methods select actions from a predefined discrete candidate set, typically via a classification head~\cite{hong2021vln,raychaudhuri2021language,wang2025g3d}. In contrast, generative methods produce actions through autoregressive generation in a continuous or open-ended space, commonly implemented with an autoregressive decoder~\cite{zhou2024navgpt,lin2025navcot,wang2025dynam3d}.

Based on a systematic analysis of methodological trends and empirical performance, this survey identifies a clear shift in the field's evolution. Early VLN research was predominantly characterized by hierarchical, discriminative models trained on task-specific datasets~\cite{tan2019learning,chen2021history,hong2021vln,wang2024causal,wang2024lookahead}. In contrast, recent advances increasingly emphasize generative foundation models enabled by large-scale pre-training~\cite{zhou2024navgpt,he2023learning,wang2025dynam3d,zhang2024navid,cheng2024navila}. Leveraging the representational and reasoning capabilities of large language models (LLMs) and large vision-language models (LVLMs)~\cite{openai2022chatgpt,Qwen2.5-VL,chen2024internvl}, contemporary studies have focused on system-level prompt engineering for zero-shot navigation as well as task-specific fine-tuning. Many recent studies have also moved beyond performance comparisons in simulation and increasingly emphasize deployment and evaluation on real robotic platforms~\cite{cheng2024navila,zhang2025embodied,zeng2025janusvln}.

\newcontent{Recent surveys have reviewed VLN and embodied navigation from complementary perspectives, including foundation-model-based VLN~\cite{zhang2024vision}, embodied intelligence~\cite{gao2024vision}, LLM-based navigation~\cite{lin2023advances}, safety and security~\cite{wang2025safety}, aerial VLN~\cite{yao2025aeroverse}, and broader VLA models for embodied AI and robotics~\cite{ma2026survey,kawaharazuka2025vla}. These works provide valuable summaries of specific methodological trends, application scenarios, or model families.}
Despite substantial progress, VLN research remains largely benchmark-driven and simulation-bound, with rapid methodological diversification obscuring the practical effectiveness of existing approaches. This survey addresses this gap through a comprehensive and structured review of VLN methods, complemented by systematic real-world evaluation. Interestingly, our findings reveal that current VLN algorithms suffer from a pronounced simulation-to-real (sim-to-real) gap, exhibit high sensitivity to instruction types, and lack sufficient robustness in obstacle avoidance and safety. This work aims to serve as a clear and comprehensive reference for understanding the VLN landscape, its evolution, key challenges, and promising directions for future research.

\begin{figure*}[thb]
    \centering
    \includegraphics[width=0.9\linewidth]{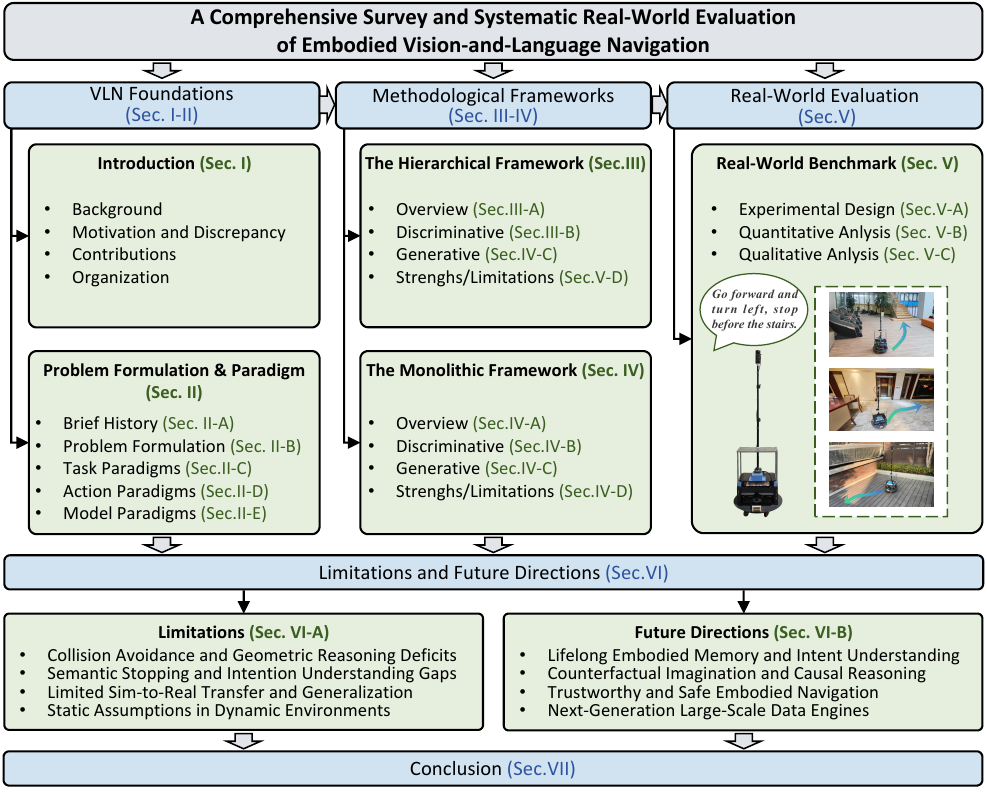}
    \caption{Overview of this survey. First, the origins of VLN are traced, followed by the problem definition and paradigm classification. Hierarchical and monolithic methods are then comprehensively reviewed, with their respective strengths and limitations analyzed. Systematic real-world experiments are conducted to reveal key challenges in deploying existing methods in physical environments. Finally, promising directions for future VLN research are outlined.}
    \label{fig_overall}
\end{figure*}

\begin{figure*}[t]
\centering
\begin{tikzpicture}
    \node[anchor=south west, inner sep=0] (image) at (0,0) {
        \includegraphics[width=\linewidth]{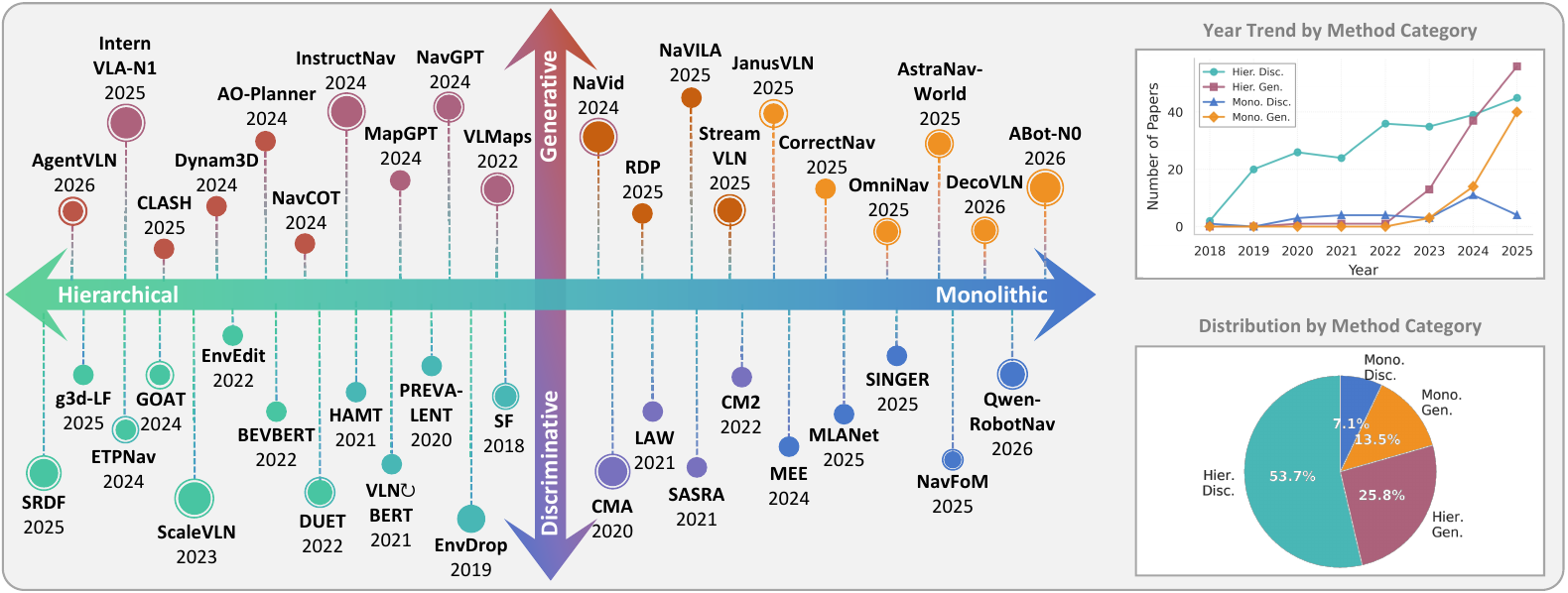}
    };

    \begin{scope}[x={(image.south east)}, y={(image.north west)}]

        % ===== Hierarchical + Generative =====
        \node[anchor=west] at (0.05, 0.686) {\scriptsize\textcolor{black!60}{\cite{xin2026agentvln}}}; % agentvln
        \node[anchor=west] at (0.09, 0.885) {\scriptsize\textcolor{black!60}{\cite{internvla-n1}}}; % Internnav-n1
        \node[anchor=west] at (0.109, 0.615) {\scriptsize\textcolor{black!60}{\cite{wang2025clash}}}; % clash
        \node[anchor=west] at (0.14, 0.686) {\scriptsize\textcolor{black!60}{\cite{wang2025dynam3d}}}; % dynam3D
        \node[anchor=west] at (0.172, 0.8) {\scriptsize\textcolor{black!60}{\cite{chen2025affordances}}}; % ao-planner
        \node[anchor=west] at (0.198, 0.624) {\scriptsize\textcolor{black!60}{\cite{lin2025navcot}}}; % navcot
        \node[anchor=west] at (0.225, 0.858) {\scriptsize\textcolor{black!60}{\cite{long2024instructnav}}}; % instructnav
        \node[anchor=west] at (0.258, 0.731) {\scriptsize\textcolor{black!60}{\cite{chen2024mapgpt}}}; % mapgpt
        \node[anchor=west] at (0.29, 0.86) {\scriptsize\textcolor{black!60}{\cite{zhou2024navgpt}}}; % navgpt
        % \node[anchor=west] at (0.271, 0.668) {\scriptsize\textcolor{black!60}{\cite{yu2023l3mvn}}}; % l3mvn
        % \node[anchor=west] at (0.294, 0.885) {\scriptsize\textcolor{black!60}{\cite{shah2023lm}}}; % LM-Nav
        \node[anchor=west] at (0.305, 0.8) {\scriptsize\textcolor{black!60}{\cite{huang23vlmaps}}}; % VLMaps

        % ===== Monolithic + Generatiev 分界 =====
        \node[anchor=west] at (0.389, 0.858) {\scriptsize\textcolor{black!60}{\cite{zhang2024navid},}}; % navid
        \node[anchor=west] at (0.389, 0.828) {\scriptsize\textcolor{black!60}{\cite{zhang2024uni}}}; % uninavid
        % \node[anchor=west] at (0.413, 0.674) {\scriptsize\textcolor{black!60}{\cite{zhang2024uni}}}; % uninavid
        \node[anchor=west] at (0.413, 0.674) {\scriptsize\textcolor{black!60}{\cite{wang2025rethinking}}}; % rdp
        \node[anchor=west] at (0.45, 0.915) {\scriptsize\textcolor{black!60}{\cite{cheng2024navila}}}; % navila
        \node[anchor=west] at (0.465, 0.724) {\scriptsize\textcolor{black!60}{\cite{wei2025streamvln}}}; % streamvln
        \node[anchor=west] at (0.51, 0.89) {\scriptsize\textcolor{black!60}{\cite{zeng2025janusvln}}}; % janusvln
        \node[anchor=west] at (0.53, 0.717) {\scriptsize\textcolor{black!60}{\cite{yu2025correctnav}}}; % correctnav
        % \node[anchor=west] at (0.57, 0.6) {\scriptsize\textcolor{black!60}{\cite{zhang2025embodied}}}; % navfom
        \node[anchor=west] at (0.57, 0.647) {\scriptsize\textcolor{black!60}{\cite{xue2025omninav}}}; % omninav
        \node[anchor=west] at (0.608, 0.84) {\scriptsize\textcolor{black!60}{\cite{hu2025astranav}}}; % astranav-world
        \node[anchor=west] at (0.632, 0.65) {\scriptsize\textcolor{black!60}{\cite{cvpr2026decovln}}}; % decoVLN
        \node[anchor=west] at (0.67, 0.732) {\scriptsize\textcolor{black!60}{\cite{chu2026abot}}}; % abot-n0

        % ===== Hierarchical + Discriminative =====
        \node[anchor=west] at (0.0343, 0.153) {\scriptsize\textcolor{black!60}{\cite{wangbootstrapping}}}; % SRDF
        \node[anchor=west] at (0.0515, 0.356) {\scriptsize\textcolor{black!60}{\cite{wang2025g3d}}}; % g3dlf
        \node[anchor=west] at (0.09, 0.23) {\scriptsize\textcolor{black!60}{\cite{an2024etpnav}}}; % etpnav
        \node[anchor=west] at (0.11, 0.324) {\scriptsize\textcolor{black!60}{\cite{wang2024causal}}}; % goat
        \node[anchor=west] at (0.142, 0.105) {\scriptsize\textcolor{black!60}{\cite{wang2023scaling}}}; % scalevln
        \node[anchor=west] at (0.16, 0.4) {\scriptsize\textcolor{black!60}{\cite{li2022envedit}}}; % envedit
        \node[anchor=west] at (0.18, 0.225) {\scriptsize\textcolor{black!60}{\cite{an2022bevbert}}}; % bevbert
        \node[anchor=west] at (0.184, 0.035) {\scriptsize\textcolor{black!60}{\cite{chen2022think}}}; % duet
        \node[anchor=west] at (0.22, 0.333) {\scriptsize\textcolor{black!60}{\cite{chen2021history}}}; % hamt
        \node[anchor=west] at (0.256, 0.153) {\scriptsize\textcolor{black!60}{\cite{hong2021vln}}}; % vlnbert
        \node[anchor=west] at (0.253, 0.223) {\scriptsize\textcolor{black!60}{\cite{hao2020towards}}}; % prevalent
        \node[anchor=west] at (0.303, 0.038) {\scriptsize\textcolor{black!60}{\cite{tan2019learning}}}; % envdrop
        \node[anchor=west] at (0.299, 0.215) {\scriptsize\textcolor{black!60}{\cite{fried2018speaker}}}; % speaker-follower

        % ===== Monolithic + Discriminative =====
        \node[anchor=west] at (0.393, 0.145) {\scriptsize\textcolor{black!60}{\cite{krantz_beyond_2020}}}; % cma
        \node[anchor=west] at (0.42, 0.265) {\scriptsize\textcolor{black!60}{\cite{raychaudhuri2021language}}}; % law
        \node[anchor=west] at (0.455, 0.164) {\scriptsize\textcolor{black!60}{\cite{irshad2021sasra}}}; % sasra
        \node[anchor=west] at (0.475, 0.323) {\scriptsize\textcolor{black!60}{\cite{georgakis2022cross}}}; % cm2
        \node[anchor=west] at (0.505, 0.2) {\scriptsize\textcolor{black!60}{\cite{he2024mee}}}; % mee
        \node[anchor=west] at (0.550, 0.263) {\scriptsize\textcolor{black!60}{\cite{he2023mlanet}}}; % mlanet
        
        \node[anchor=west] at (0.583, 0.359) {\scriptsize\textcolor{black!60}{\cite{adang2025singer}}}; % singer
        \node[anchor=west] at (0.622, 0.190) {\scriptsize\textcolor{black!60}{\cite{zhang2025embodied}}}; % navfom
        \node[anchor=west] at (0.655, 0.323) {\scriptsize\textcolor{black!60}{\cite{zhang2026qwen}}}; % qwen-robotnav

    \end{scope}
\end{tikzpicture}
\caption{Timeline of representative VLN methods categorized by action space and model paradigm. The four quadrants correspond to different combinations of action paradigms and model paradigms. The right panel presents the temporal trends by method category, and the distribution of methods across categories.}
\label{fig_timeline}
\end{figure*}

The remainder of this paper is organized as follows. Sec.~\ref{sec_problem_formulation} formalizes the VLN problem, reviews the evolution of visual navigation, and introduces task paradigms, datasets, simulators, and evaluation metrics. Sec.~\ref{sec_hierarchical_framework} and~\ref{sec_monolithic_framework} systematically survey hierarchical and monolithic VLN approaches, respectively, analyzing their architectural principles, representative methods, and comparative characteristics. Sec.~\ref{sec_real_world} presents our empirical analysis of real-world evluation, reports extensive experimental results across 10 diverse scenes, and provides both quantitative and qualitative analysis of navigation performance, failure modes, and the sim-to-real gap. Sec.~\ref{sec_limitations} discusses limitations of current approaches and identifies future research directions. Finally, Sec.~\ref{sec_conclusion} concludes the paper. Fig.~\ref{fig_overall} provides a visual overview of the survey structure. Fig.~\ref{fig_timeline} illustrates the taxonomy of VLN methods, along with their publication timeline and statistical distribution.

\section{Problem Formulation and VLN Paradigms}
\label{sec_problem_formulation}

\subsection{A Brief History of Visual Navigation}

Visual navigation, which requires interpreting the environment through optical sensors to determine a viable path, has been a long-standing challenge in robotics and artificial intelligence.
Early work in the 1980s, such as Carnegie Mellon's Navlab system \cite{navlab_visnav} and modular architectures for autonomous vehicles \cite{Waxman_visnav_alv}, demonstrated the feasibility of vision-based navigation using cameras. 
From the late 1980s to 2000s, Simultaneous Localization and Mapping (SLAM) became the foundation for visual navigation, with seminal contributions by Smith \textit{et al.}~\cite{Smith1986, Smith1990}, Moutarlier and Chatila~\cite{moutarlier1989experimental}, and Krotkov~\cite{Krotkov1989}.
The introduction of visual odometry by Nistér \textit{et al.}~\cite{1315094} enabled real-time estimation of camera motion from image sequences.
Between 2000s and 2010s, modern VSLAM systems, including MonoSLAM \cite{4160954}, PTAM \cite{4538852}, LSD-SLAM \cite{lsd_slam}, and ORB-SLAM2 \cite{orb_slam2}, improved robustness and accuracy, supporting monocular, stereo, and RGB-D inputs.
Learning-based approaches such as DeepVO \cite{deepvo} and DROID-SLAM \cite{droid_slam} further enhanced performance and generalization by incorporating neural network priors.

Despite these advances, conventional modular SLAM-based navigation systems remain predominantly model-driven and rely on strong prior assumptions about environmental geometry, sensor fidelity, and task structure. \newcontent{Although such systems perform well in deterministic and well-structured scenarios, they can become fragile in broader uncertain settings due to accumulated localization errors, sensor noise, dynamic obstacles, and deviations from pre-built maps.} While highly effective for metric localization and mapping, such systems are less suited to open-ended tasks where the goal is specified by natural language rather than by a precise coordinate or pre-built map. \newcontent{For example, a robot may be asked to ``go to the chair beside the window,'' ``Find the printer in the office and bring me the document on the desk,'' or ``return to the corridor after passing the lounge,'' where successful navigation requires grounding semantic landmarks, resolving underspecified goals, and inferring human intent from context.} Therefore, a series of data-driven, end-to-end navigation tasks have been proposed~\cite{wijmansdd2019ddppo,zhu2017target,chaplot2020object}. Motivated by how humans navigate using both perception and language to interpret intent, resolve ambiguity, and adapt to unseen situations, VLN~\cite{anderson2018vision} was introduced to augment navigation with instruction-conditioned semantic reasoning, enabling agents to follow natural-language directives and execute task-oriented behaviors in novel environments. \newcontent{By integrating vision-language mechanisms into end-to-end decision making, data-driven navigation models provide robots with stronger semantic understanding and contextual reasoning capabilities, offering greater potential for handling dynamic unseen environments and complex instructions. Compared with traditional navigation pipelines, VLN is particularly advantageous in human-centered scenarios where users provide flexible linguistic instructions, target objects or regions are semantically defined, and the environment may be unfamiliar or only partially observable.} This shift toward language-guided, goal-driven navigation offers a promising path to improved generalization and robustness in uncertain and dynamically changing settings.

\subsection{Problem Formulation}

VLN is a challenging task where an agent is required to follow natural language instructions to navigate a visually realistic environment. The agent starts at a specific location and orientation and must determine a sequence of actions that leads to the target location described by the instruction. 

Formally, VLN task is modeled as a Partially Observable Markov Decision Process (POMDP), defined by the tuple $\mathcal{M} = \langle \mathcal{S}, \mathcal{A}, \mathcal{T}, \mathcal{O}, \Omega, \mathcal{I} \rangle$.

\subsubsection{General Framework}
At time step $t$, the agent is in a state $s_t \in \mathcal{S}$, which encapsulates the agent's position and orientation within the environment. The agent receives an observation $o_t \in \mathcal{O}$ via the observation function $o_t \sim \Omega(s_t)$. The observation typically consists of egocentric visual inputs (\textit{e.g.}, RGB-D images, panoramic views) and sensor readings. The navigation is conditioned on a natural language instruction $\mathcal{I} = \{w_1, w_2, \dots, w_L\}$, where $w_i$ represents the $i$-th token in the sequence, and $L$ denotes the length of the instruction.

The objective of the agent is to learn a policy $\pi(a_t | o_{\le t}, \mathcal{I})$ that maps the instruction and the history of observations to an action $a_t \in \mathcal{A}$. The transition function $\mathcal{T}(s_{t+1} | s_t, a_t)$ defines the dynamics of the environment. The episode terminates when the agent issues a \texttt{STOP} action or exceeds the maximum time steps. Success is determined by whether the agent's final state $s_T$ falls within a threshold distance $d_{th}$ of the ground-truth target defined by $\mathcal{I}$.

\subsubsection{Discrete Graph-based Environments}
In discrete environment settings~\cite{anderson2018vision,qi2020reverie,ku2020room}, the environment is abstracted as a navigation graph $\mathcal{G} = (\mathcal{V}, \mathcal{E})$.
\begin{itemize}
    \item \textbf{State Space:} The state space $\mathcal{S}$ corresponds to the discrete set of navigable viewpoints $\mathcal{V}$. The agent's location is restricted to these pre-defined nodes.
    \item \textbf{Action Space:} The action space $\mathcal{A}_t$ consists of adjacent neighbor nodes $\{v \mid (s_t, v) \in \mathcal{E}\}$ and the \texttt{STOP} action.
    \item \textbf{Dynamics:} The transition function $\mathcal{T}$ is deterministic and simplified. Selecting a neighbor node teleports the agent instantly to that viewpoint without intermediate physical simulation.
\end{itemize}

\subsubsection{Continuous Euclidean-based Environments}
In typical continuous environment settings~\cite{krantz_beyond_2020,song2025towards,wang2025rethinking}, the agent operates in a continuous Euclidean space $\mathcal{S} \subset \mathbb{R}^3 \times SO(2)$.
\begin{itemize}
    \item \textbf{State Space:} The state $s_t$ represents the precise continuous coordinates and heading. The agent is not bound to a pre-defined topological graph and can move freely within the navigable geometry.
    \item \textbf{Action Space:} 
    The action space $\mathcal{A}$ in continuous environments is categorized into two paradigms:
    \begin{itemize}
        \item \emph{Low-level Actions:} The agent executes atomic commands, such as:
        \begin{align*}
        \mathcal{A}_{low} = \{ 
            & \texttt{STOP}, \, \texttt{Move\_Forward}(\delta), \\
            & \texttt{Turn\_Left}(\theta), \texttt{Turn\_Right}(\theta),\}
        \end{align*}
        where $\delta$ and $\theta$ denote fixed step sizes for translation and rotation, respectively. 
        \item \emph{High-level Waypoints:} The action is formulated to predict a relative sub-goal $(r, \psi)$, where $r$ is the radial distance and $\psi$ is the heading relative to the agent's current pose. A local controller then executes the low-level actions to reach this waypoint.
    \end{itemize}
    \item \textbf{Dynamics:} The transition $\mathcal{T}$ involves a physics engine. Unlike teleportation, state transitions are affected by actuation noise and collisions: $s_{t+1} = \text{Sim}(s_t, a_t) + \epsilon$, where $\epsilon$ denotes stochastic disturbances. This setting necessitates robust and physically grounded policy learning.
\end{itemize}

\subsubsection{Optimization Goal}
Regardless of the environment type, the overarching goal is to maximize the expectation of reaching the target:
\begin{equation}
    \theta^* = \operatorname*{argmax}_\theta \mathbb{E}_{\tau \sim \pi_\theta} \left[ \mathbb{I}(\text{dist}(s_T, s_{target}) < d_{th}) \right],
\end{equation}
where $\tau$ is the trajectory generated by the policy parameterized by $\theta$, and $\mathbb{I}(\cdot)$ is the indicator function for success.

\subsection{Task Paradigms: Environment and Benchmark Ecosystem}
\subsubsection{Dataset}
\begin{figure*}[t]
    \centering
    \subfloat{%
        \includegraphics[width=0.5\textwidth]{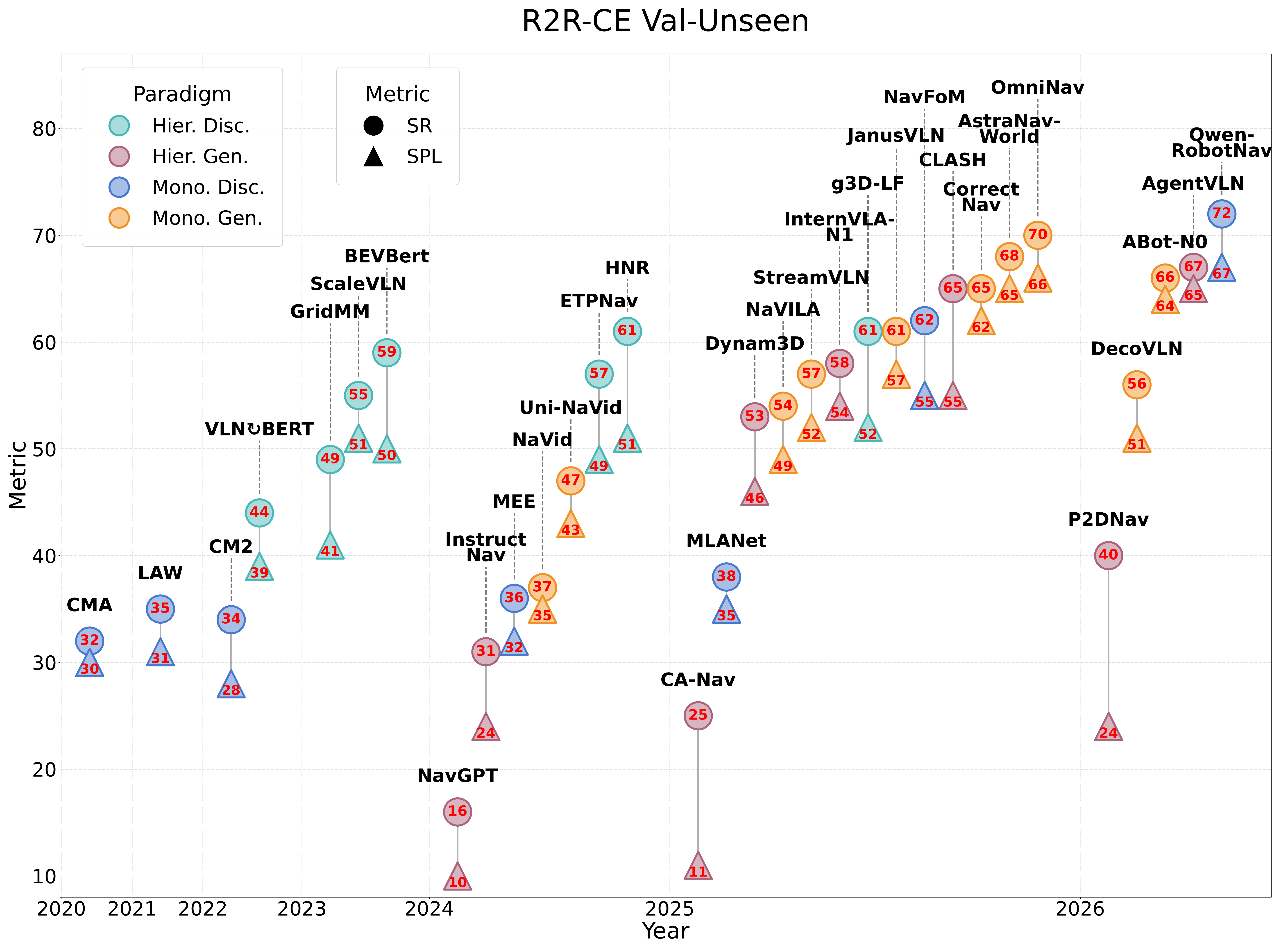}%
    }
    \hfill
    \subfloat{%
        \includegraphics[width=0.5\textwidth]{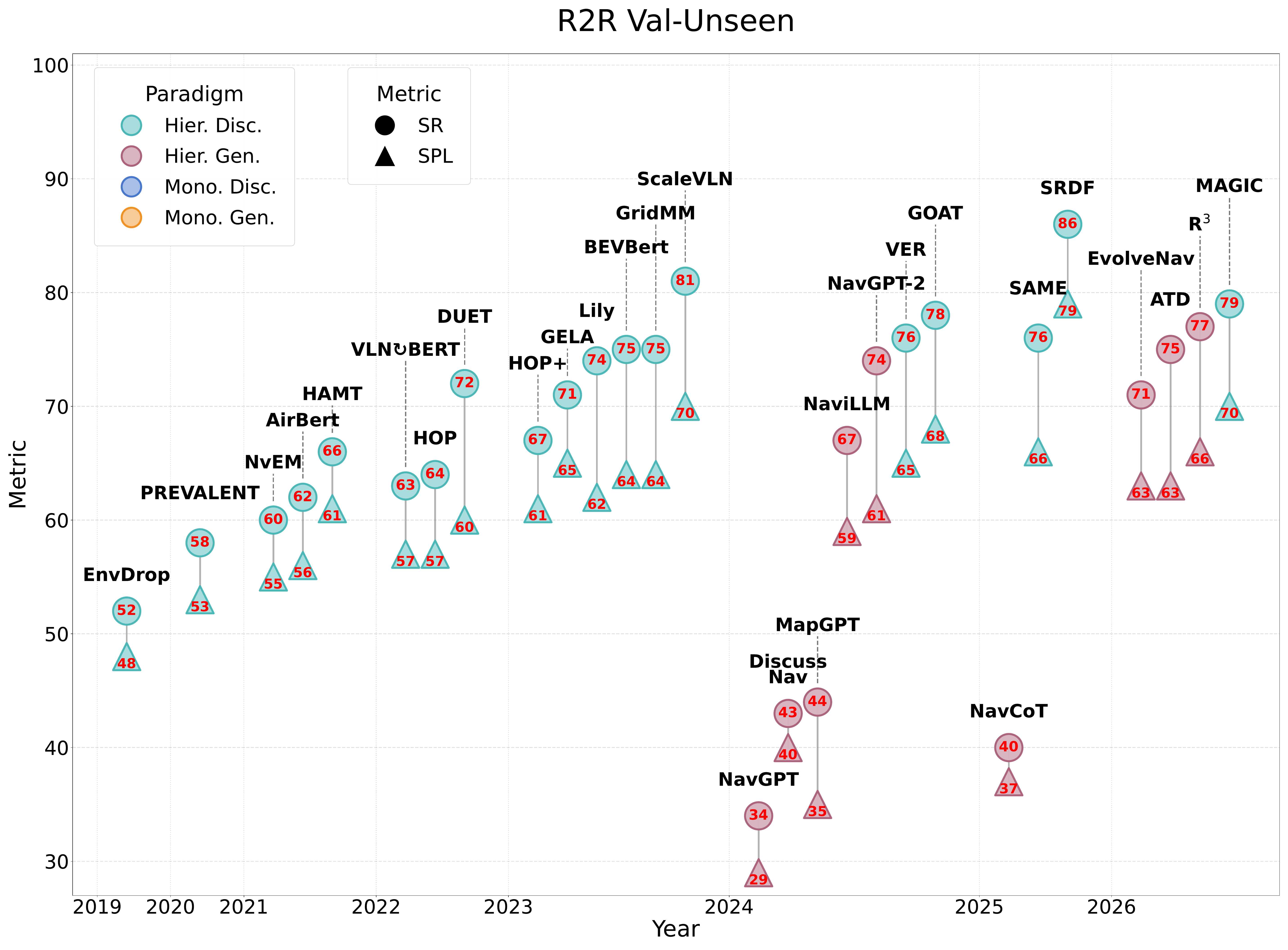}%
    }
    \caption{Performance trends of representative VLN methods on the validation unseen splits of R2R~\cite{anderson2018vision} and R2R-CE~\cite{krantz_beyond_2020}, measured by SR and SPL.}
    \label{fig:r2r_val_unseen_sr_spl}
\end{figure*}

The evolution of VLN datasets reflects the field's ongoing pursuit of realism and complexity, as summarized in Table~\ref{tab:vln_datasets}. 
\newcontent{The performance trends of representative VLN methods on widely used benchmark datasets R2R~\cite{anderson2018vision} and R2R-CE~\cite{krantz_beyond_2020} are shown in Figure~\ref{fig:r2r_val_unseen_sr_spl}.}
While foundational benchmarks primarily focused on discrete, graph-based navigation within static indoor scans, recent efforts have gravitated towards continuous environments that necessitate low-level control and physical interaction to better bridge the sim-to-real gap. This overview categorizes existing datasets by environmental domain, ranging from discrete and continuous indoor spaces to complex outdoor settings, and distinguishes them by task specifications, such as fine-grained object manipulation, multi-turn dialogue, and long-horizon aerial navigation.

\begin{table*}[htbp]
    \centering
    \caption{Comprehensive Summary of Vision-and-Language Navigation Datasets.}
    \label{tab:vln_datasets}
    \resizebox{\linewidth}{!}{%
    \begin{tabular}{l l l l c l c l}
    \toprule
    \textbf{Dataset} & \textbf{Simulator} & \textbf{Domain} & \textbf{Task Type} & \textbf{Action} & \textbf{Goal Type} & \textbf{Samples} & \textbf{Characteristics} \\
    \midrule
    \multicolumn{8}{c}{\textit{\textbf{Discrete Indoor Environments}}} \\
    \midrule
    R2R \cite{anderson2018vision} & Matterport3D & Indoor & Navigation & High-level & PointGoal & 22K & Foundational dataset; Shortest path bias \\
    R4R \cite{jain2019stay} & Matterport3D & Indoor & Navigation & High-level & PointGoal & 280K & Extension of R2R; Longer, twisted paths \\
    RxR \cite{ku2020room} & Matterport3D & Indoor & Navigation & High-level & PointGoal & 126K & Multilingual; Time-aligned pose traces \\
    CEREALBAR \cite{suhr2019executing} & -- & Indoor & Dialogue & High-level & Location/Object & 1K & Interactive instruction tuning/correction \\
    LANI/CHAI \cite{misra2018mapping} & CHALET & Indoor & Manipulation & High-level & Landmark & 8K & Block manipulation in 3D; 3D Navigation\\
    CVDN \cite{thomason2020vision} & Matterport3D & Indoor & Dialogue & High-level & PointGoal & 2K & Two-party cooperative dialogue for navigation \\
    EmbodiedQA \cite{das2018embodied} & House3D & Indoor & QA & High-level & Answer Retrieval & 5K & Answering questions requiring embodied navigation \\
    HANNA \cite{nguyen2019help} & Matterport3D & Indoor & Interactive & High-level & ObjectGoal & 103K & Request-response assistance; Error correction \\
    Just Ask \cite{chi2020just} & Matterport3D & Indoor & Dialogue & High-level & Location/Object & 22K & Dialogue to refine navigation instruction \\
    Landmark-RxR \cite{he2021landmark} & Matterport3D & Indoor & Navigation & High-level & LandmarkGoal & 167K & Navigating to visual landmarks/signs \\
    REVERIE \cite{qi2020reverie} & Matterport3D & Indoor & Remote Object & High-level & ObjectGoal & 22K & High-level instr.; Remote object grounding \\
    RoomNav \cite{wu2018building} & House3D & Indoor & Navigation & High-level & RoomGoal & 1K & Navigating to specified room categories \\
    SOON \cite{Zhu_2021_SOON} & Matterport3D & Indoor & Object Nav. & High-level & ObjectGoal & 4K & Orientation-less nodes; Scene-oriented instr. \\
    ASKNAV(VNLA) \cite{nguyen2019vision} & Matterport3D & Indoor & Navigation & High-level & PointGoal & 115K & Active language feedback; Guidance dataset \\
    XL-R2R \cite{yan2019cross} & Matterport3D & Indoor & Navigation & High-level & PointGoal & 17K & Cross-lingual extension; Longer instructions \\
    
    \midrule
    \multicolumn{8}{c}{\textit{\textbf{Continuous Indoor Environments}}} \\
    \midrule
    IQA \cite{gordon2018iqa} & AI2-THOR & Indoor & QA & Low-level & Answer Retrieval & 75K & Visual Question Answering within the scene \\
    VLN-CE \cite{krantz_beyond_2020} & Habitat & Indoor & Navigation & Low-level & PointGoal & 17K & Transferred R2R paths to continuous space \\
    ALFRED \cite{shridhar2020alfred} & AI2-THOR & Indoor & Manipulation & Low-level & Interaction & 25K & Long-horizon; Navigation and Object Interaction \\
    Robo-VLN \cite{irshad2021hierarchical} & Habitat & Indoor & Navigation & Low-level & PointGoal & 10K & Focus on lower-level control dynamics \\
    TEACh \cite{padmakumar2022teach} & AI2-THOR & Indoor & Dialogue & Low-level & Interaction & 3K & Dialogue-based complex interaction tasks \\
    REVE-CE \cite{li2022reve-ce} & Habitat & Indoor & Remote Object & Low-level & ObjectGoal & 4K & Continuous version of REVERIE; High-level instr. \\
    DialFRED \cite{gao2022dialfred} & AI2-THOR & Indoor & Dialogue & Low-level & Interaction & 53K & Dialogue system based on ALFRED tasks \\
    Behavior-1k \cite{li2024behavior} & Gibson & Indoor & Complex Housework & Low-level & Interaction & 1K & 1,000 human-centered activities; High realism \\
    IVLN \cite{krantz2023iterative} & Matterport3D\&Habitat & Indoor & Iterative & Low-level & Seq. PointGoal & 1K & Tour guide task; Memory persistence across goals \\
    GOAT-Bench \cite{khanna2024goat} & Habitat & Indoor & Iterative & Low-level & Multi-Modal & 725K & Open-vocab; Image/Text/Category goals \\
    VLN-PE \cite{wang2025rethinking} & GRUTopia & Indoor & Navigation & Low-level & PointGoal & 13K & Focuses on ``embodied gap"; Physical/Visual disparities \\
    LHPR-VLN \cite{song2025towards} & Habitat & Indoor & Long-Horizon & Low-level & Sub-goals & 3K & Multi-stage planning; NavGen platform \\
    VLNVerse \cite{lin2025vlnversebenchmarkvisionlanguagenavigation} & GRUTopia & Indoor & Navigation & Low-level & PointGoal & 85K & Large-scale, high-quality indoor 3DGS environments\\
    
    \midrule
    \multicolumn{8}{c}{\textit{\textbf{Outdoor Environments}}} \\
    \midrule
    ANDH \cite{fan2023aerial} & xView & Outdoor & Aerial Nav. & High-level & PointGoal & 6K & Aerial navigation with dialogue history \\
    AerialVLN \cite{liu2023aerialvln} & AirSim & Outdoor & Aerial Nav. & High-level & PointGoal & 25K & 3D movement; Bird's-eye view \\
    CDNLI \cite{roh2020conditional} & CARLA & Outdoor & Driving Nav. & Low-level & PointGoal & -- & Cross-domain instruction transfer (city maps) \\
    LCSD \cite{sriram2019talk} & CARLA & Outdoor & Dialogue & High-level & PointGoal & 140 & Cooperative scene description (Local to Global) \\
    GRBench (SocialNav) \cite{wang2024grutopia} & GRUTopia & Outdoor & General Robotics & Low-level & Multi-Task & 300 & City scale simulation; Diverse environment benchmark \\
    MC Collab \cite{narayan2019collaborative} & Minecraft & Outdoor & Dialogue & High-level & Object/Location & 509& Collaborative object/location finding in block environment \\
    RobotSlang \cite{banerjee2021robotslang} & Real & Outdoor & Dialogue & High-level & Location & 169 & Dialogue for outdoor navigation; Requires disambiguation \\
    SDN \cite{ma-etal-2022-dorothie} & CARLA & Outdoor & Driving Nav. & High/Low-level & PointGoal & 183 & Social driving cues in dynamic environment \\
    StreetLearn \cite{mirowski2019streetlearn} & Street View & Outdoor & Navigation & High-level & PointGoal & 114K & Large-scale, real-world Google Street View environment \\
    StreetNav \cite{Hermann2019LearningTF} & Street View & Outdoor & Navigation & High-level & PointGoal & 613K & Navigating with directional cues and landmarks \\
    Talk the Walk \cite{de2018talk} & Real & Outdoor & Dialogue & High-level & Location & 10K & Outdoor pedestrian navigation guided by human dialogue \\
    Talk2Nav \cite{vasudevan2021talk2nav} & Street View & Outdoor & Dialogue & High-level & PointGoal & 10K & Long-range nav; Dual attention; Spatial memory \\
    TOUCHDOWN \cite{chen2019touchdown} & Street View & Outdoor & Navigation & High-level & PointGoal & 9K & Real-world urban streets; Nav and SDR \\
    \bottomrule
    \end{tabular}%
    }
    \end{table*}

\subsubsection{Simulator}
For embodied tasks such as VLN, a real-time, high-fidelity simulation platform that spans diverse environments and robot types is essential. Such a platform must support on-the-fly perception and interaction, enabling agents to continuously collect multi-round data and facilitating fair, scalable, and fully online evaluation. As summarized in Table~\ref{tab:vln_simulators}, this survey reviews the most widely used VLN-related simulators, outlining their representative datasets, strengths and limitations, and providing example environment snapshots.
\begin{table*}[htbp]
    \centering
    \caption{Comparison of major embodied navigation simulators}
    \label{tab:vln_simulators}
    \resizebox{\linewidth}{!}{
    \begin{tabular}{
        m{2.4cm}<{\centering}  % Platform
        m{1.8cm}<{\centering}  % Action Space
        m{2.8cm}<{\centering}  % Scene Data
        m{3.2cm}<{\centering}  % Primary Datasets
        m{2.8cm}<{\centering}  % Advantages
        m{2.8cm}<{\centering}  % Limitations
        m{3.2cm}<{\centering}  % Example View
    }
    \toprule
    Platform & Action Space & Scene Data & Primary Embodied Datasets & Advantages & Limitations & Example View \\ 
    \midrule

    \makecell{Matterport3D \\ Simulator~\cite{anderson2018vision} \\ \scriptsize (Georgia Tech)} &
    Discrete &
    Real indoor environments &
    R2R~\cite{anderson2018vision}, REVERIE~\cite{qi2020reverie}, R4R~\cite{jain2019stay}, RxR~\cite{ku2020room}, CVDN~\cite{thomason2020vision}, etc. &
    High photorealism &
    No continuous control &
    \includegraphics[width=2cm]{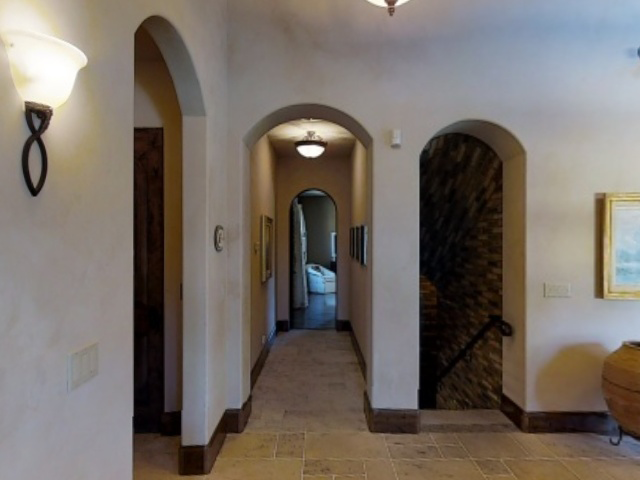} \\

    \midrule

    \makecell{Habitat~\cite{savva2019habitat,szot2021habitat} \\ \scriptsize (Meta)} &
    Continuous &
    3D reconstructions and synthetic scenes &
    VLN-CE~\cite{krantz_beyond_2020},
    LH-VLN~\cite{song2025towards}, ObjectNav~\cite{anderson2018evaluation}, HITL~\cite{puig2023habitat3}, PointGoal~\cite{anderson2018evaluation}, GOAT-bench~\cite{khanna2024goat}, etc. &
    Supports multiple navigation tasks and NPCs &
    Limited support for complex sensors and interactions &
    \includegraphics[width=2cm]{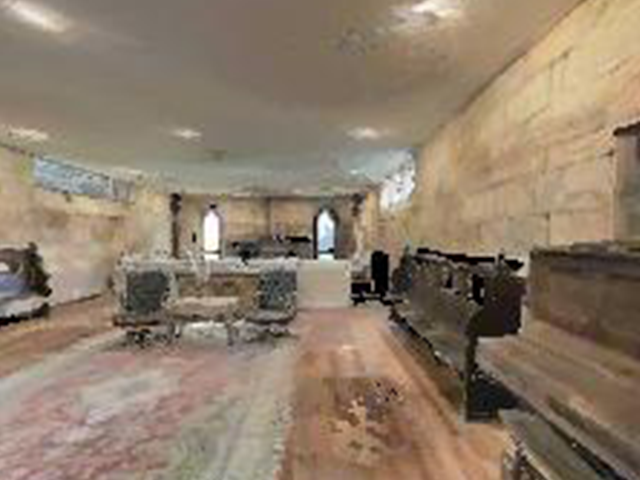} \\

    \midrule

    \makecell{Gibson~\cite{xia2020interactive} \\ \scriptsize (Stanford University)} &
    Continuous &
    Synthetic indoor scenes &
    iGibson~\cite{shen2021igibson,li2022igibson}, BEHAVIOR~\cite{srivastava2022behavior}, BEHAVIOR-1K~\cite{li2024behavior}, etc. &
    Object and scene interactivity &
    Complex task setup &
    \includegraphics[width=2cm]{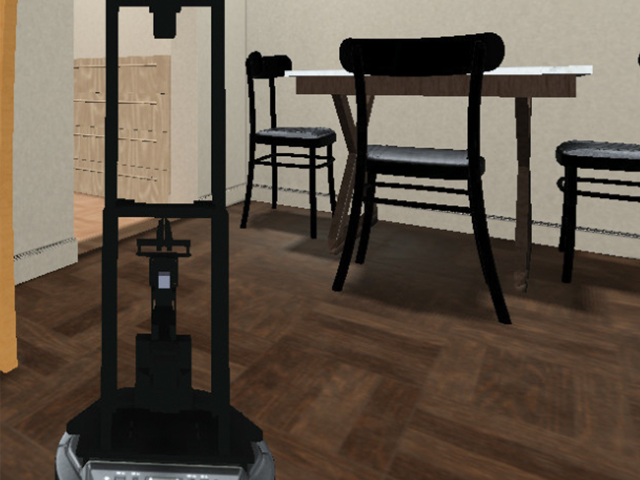} \\

    \midrule

    \makecell{AI2-THOR~\cite{kolve2017ai2thor} \\ \scriptsize (Allen Institute for AI)} &
    Continuous &
    Synthetic indoor scenes &
    ObjectNav~\cite{kolve2017ai2thor}, ALFRED~\cite{shridhar2020alfred}, EQA, ProcTHOR~\cite{deitke2022️procthor}, ManipulaTHOR~\cite{ehsani2021manipulathor}, RoboTHOR~\cite{deitke2020robothor}, etc. &
    Highly interactive environments &
    Limited photorealism &
    \includegraphics[width=2cm]{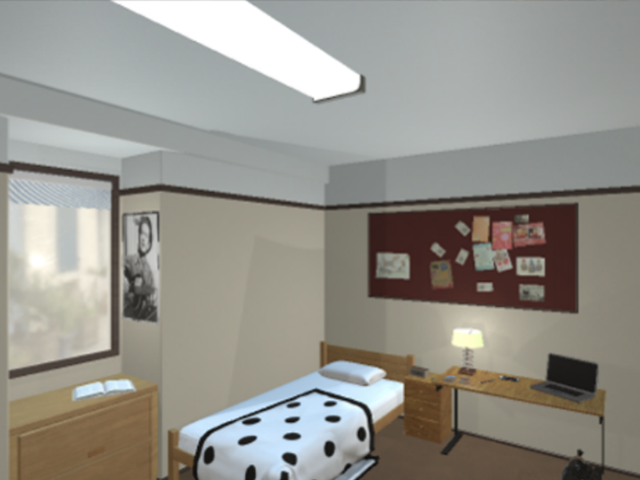} \\

    \midrule

    \makecell{GRUTopia~\cite{wang2024grutopia} \\ \scriptsize (Shanghai AI Lab)} &
    Continuous &
    3D reconstructions and synthetic scenes &
    VLN-PE~\cite{wang2025rethinking}, SocialNav~\cite{wang2024grutopia} &
    High-fidelity physics and robot simulation &
    High configuration and computational cost &
    \includegraphics[width=2cm]{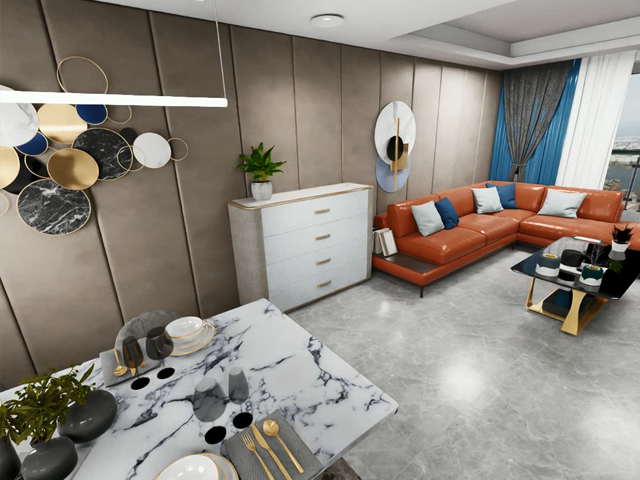} \\

    \midrule

    \makecell{CARLA~\cite{dosovitskiy2017carla} \\ \scriptsize (Intel Labs et al.)} &
    Continuous &
    Synthetic urban driving scenes &
    SDN~\cite{ma-etal-2022-dorothie}, Think2Drive~\cite{li2024think2drive}, DriveLM~\cite{sima2024drivelm}, etc. &
    Comprehensive traffic simulation with multi-sensor support &
    Focused on outdoor driving scenarios &
    \includegraphics[width=2cm]{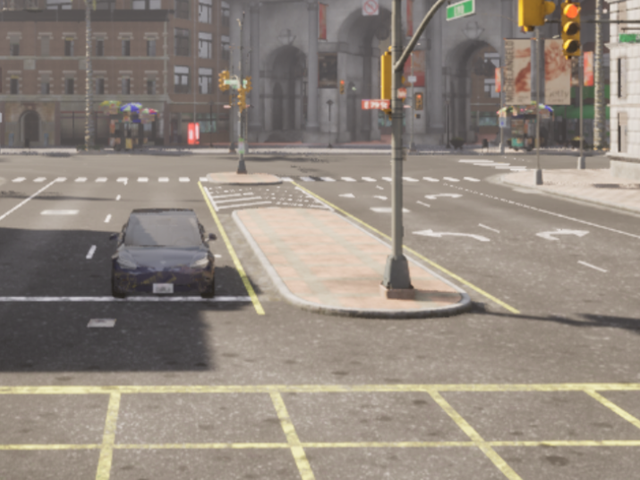} \\

    \bottomrule
    \end{tabular}}
\end{table*}

\subsubsection{Metrics}

To comprehensively evaluate the performance of vision-and-language navigation agents across different task settings and datasets, researchers have proposed a suite of metrics. These metrics assess navigation accuracy, path efficiency, trajectory similarity, and target-grounding capability. Their definitions are as follows.

(1) Navigation Error (NE)

NE measures the Euclidean distance between the agent's final position and the goal location, expressed in meters:
\begin{equation}
\mathrm{NE} = \frac{1}{N} \sum_{i=1}^{N} \lVert p_{i,T} - p_{i,g} \rVert_2,
\end{equation}
where $p_{i,T}$ is the final agent position and $p_{i,g}$ is the goal position. $N$ is the number of episodes.

(2) Trajectory Length (TL)

TL denotes the total distance traveled by the agent during a navigation episode:
\begin{equation}
\mathrm{TL} = \frac{1}{N} \sum_{i=1}^{N} \sum_{t=1}^T \lVert p_t - p_{t-1} \rVert_2,
\end{equation}
where $T$ is the number of steps and $p_t$ is the agent's position at step $t$. TL evaluates the absolute average path length.

(3) Success Rate (SR)

SR measures whether the agent stops within a predefined threshold distance $d_{\text{th}}$ of the goal:
\begin{equation}
\mathrm{SR} = \frac{1}{N}\sum_{i=1}^N \mathds{1}(\text{NE}_i \leq d_{\text{th}}),
\end{equation}
where $d_{\text{th}}$ is typically set to 3 meters.

(4) Success Rate weighted by Path Length (SPL)~\cite{anderson2018evaluation}

SPL extends SR by incorporating path efficiency:
\begin{equation}
\mathrm{SPL} = \frac{1}{N} \sum_{i=1}^N S_i \cdot \frac{SL_i}{\max(TL_i, SL_i)},
\end{equation}
where $S_i$ is the success indicator, $SL_i$ is the shortest path length, and $TL_i$ is the actual trajectory length. SPL is a key metric as it jointly evaluates success and efficiency.

(5) Oracle Success Rate (OSR)

An episode is considered successful if the agent enters the success radius at any time step:
\begin{equation}
\mathrm{OSR} = \frac{1}{N} \sum_{i=1}^N \mathds{1}\Big(\min_t \lVert p_t - p_g \rVert_2 \leq d_{\text{th}}\Big).
\end{equation}
OSR reflects whether the trajectory could have succeeded with an optimal stopping policy. A large gap between OSR and SR indicates difficulty in learning when to stop.

(6) Normalized Dynamic Time Warping (nDTW)~\cite{magalhaes2019general}

nDTW measures the spatial and temporal similarity between predicted and reference trajectories:
\begin{equation}
\mathrm{nDTW} = \frac{1}{N}\sum_{i=1}^N
\exp\left(- \frac{\mathrm{DTW}(R_i, Q_i)}{|R_i| \cdot d_{\text{th}}}\right),
\end{equation}
where $|R_i|$ is the length of the reference trajectory and $d_{\text{th}}$ is the success threshold. DTW computes the minimum cumulative alignment cost:
\begin{equation}
\mathrm{DTW}(R, Q) = \min_{M \in \mathcal{M}} \sum_{(i_k, j_k)\in M} \gamma(r_{i_k}, q_{j_k}),
\end{equation}
where $\gamma(\cdot,\cdot)$ is a distance function and $\mathcal{M}$ is the set of valid alignment paths.

(7) Success-weighted DTW (sDTW)

sDTW introduces a success constraint on top of nDTW, assigning non-zero scores only when the trajectory ends within the success threshold:
\begin{equation}
\mathrm{SDTW} = \frac{1}{N}\sum_{i=1}^N
S_i \cdot
\exp\left(- \frac{\mathrm{DTW}(R_i, Q_i)}{|R_i| \cdot d_{th}}\right),
\end{equation}
providing a stricter evaluation that considers both trajectory fidelity and task success.

(8) Remote Grounding Success Rate (RGS)

In target-object-oriented tasks (\textit{e.g.}, REVERIE~\cite{qi2020reverie}), the agent must not only navigate to the vicinity of the target location but also correctly identify the specified object. RGS measures success under both conditions:
\begin{equation}
\mathrm{RGS} = \frac{1}{N}\sum_{i=1}^N \mathds{1}\big(\text{NE}_i \leq d_{\text{th}} \wedge \hat{o}_i = o_i \big),
\end{equation}
where $\hat{o}_i$ is the predicted target and $o_i$ is the ground truth.

(9) Remote Grounding Success Rate weighted by Path Length (RGSPL)

To jointly assess grounding success and path efficiency, RGSPL integrates RGS with SPL-style path weighting:
\begin{equation}
\mathrm{RGSPL} = \frac{1}{N}\sum_{i=1}^N \mathds{1}\big(\mathrm{NE}_i \leq d_{\text{th}} \wedge \hat{o}_i = o_i \big) \cdot \frac{SL_i}{\max(TL_i, SL_i)},
\end{equation}
RGSPL rewards agents that achieve both accurate target grounding and efficient navigation.

\subsection{Action Paradigms: Waypoint-based vs. Action-based}
\begin{figure}[t]
    \centering
    \includegraphics[width=\linewidth]{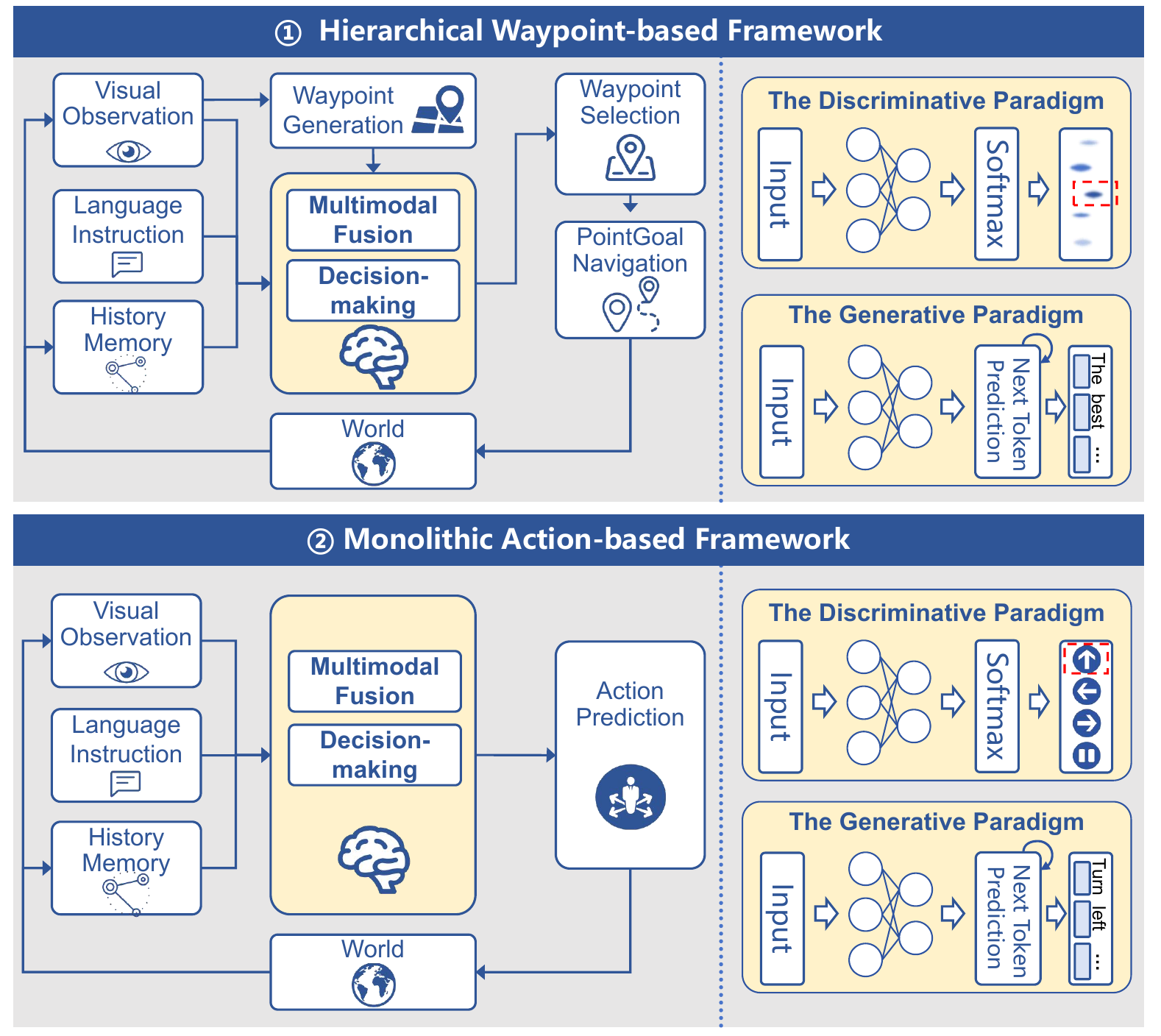}
    \caption{Comparison of different VLN action paradigms.}
    \label{fig_framework_comparison}
\end{figure}

After years of development, two dominant action paradigms have emerged in the VLN literature, as illustrated in Fig.~\ref{fig_framework_comparison}. The first paradigm introduces candidate waypoints as intermediate decision targets, decomposing navigation into high-level decision making and low-level execution, typically handled by separate models or policies. The second paradigm directly defines the action space as low-level continuous control signals, such as translational distance and rotation angle, enabling end-to-end navigation using a single policy without explicit hierarchical decomposition.

\subsubsection{The Hierarchical Waypoint-based Framework}
The hierarchical waypoint-based framework formulates navigation as a two-stage decision process. At each time step, the agent first selects a high-level waypoint or subgoal from a finite candidate set, and then executes a sequence of low-level actions to reach the selected waypoint. Let $o_t$ denote the agent’s observation, $\mathcal{I}$ the instruction, and $\mathcal{Q}_t = \{q_t^1, \dots, q_t^K\}$ a set of candidate waypoints derived from the environment graph or panoramic viewpoints. The high-level policy is defined as:
\begin{equation}
\pi_{\text{high}}(q_t \mid o_t, h_t, \mathcal{I}), \quad q_t \in \mathcal{Q}_t,
\end{equation}
where $h_t$ represents the agent's internal state or memory.

Once a waypoint $q_t$ is selected, a low-level controller executes a sequence of primitive actions $\{a_{t,1}, \dots, a_{t,M}\}$ to reach the target:
\begin{equation}
\pi_{\text{low}}(a_{t,m} \mid s_{t,m}, q_t),
\end{equation}
where $s_{t,m}$ denotes the low-level state during execution.

This hierarchical decomposition reduces the complexity of long-horizon planning by constraining high-level decisions to a discrete and structured space, while delegating motion execution to specialized controllers. As a result, waypoint-based frameworks are widely adopted in classical VLN benchmarks and graph-based environments, offering stable training and strong performance under well-defined navigation graphs. However, their reliance on predefined waypoint generation and environment-specific graph structures may limit flexibility in continuous or highly dynamic settings.

\subsubsection{The Monolithic Action-based Framework}
In contrast, the monolithic action-based framework removes explicit hierarchical structure and directly predicts low-level actions in a continuous or fine-grained action space. Typically, navigation is modeled as a sequential control problem, where the agent outputs an action $a_t$ at each time step based solely on its current observation and instruction:
\begin{equation}
\pi_\theta(a_t \mid o_t, h_t, \mathcal{I}), \quad a_t \in \mathcal{A}_{\text{low}},
\end{equation}
where $\mathcal{A}_{\text{low}}$ denotes a predefined set of primitive actions or control commands.

The policy is trained end-to-end using imitation learning or reinforcement learning objectives, for example by minimizing the cross-entropy loss between the predicted action distribution and the expert actions:
\begin{equation}
\mathcal{L}_{\text{IL}} = - \sum_{t} \log \pi_\theta(a_t^{*} \mid o_t, h_t, \mathcal{I}),
\end{equation}
or by maximizing expected cumulative reward:
\begin{equation}
\max_{\theta} \; \mathbb{E}_{\pi_\theta} \left[ \sum_t reward_t \right].
\end{equation}

More recently, a line of work~\cite{wang2025rethinking,internvla-n1,xue2025omninav,hu2025astranav,wei2025ground} has explored conditional diffusion policies~\cite{sridhar2024nomad,chi2023diffusion} as the action head, enabling the generation of dense, multi-step trajectories rather than single-step actions:
\begin{equation}
\mathcal{U} = \{x_i, y_i, \sin \theta_i, \cos \theta_i, c_i\}_{i=1}^{N},
\end{equation}
where $\mathcal{U}$ denotes the predicted waypoint trajectory, $(x_i, y_i)$ represent the planar coordinates of the $i$-th waypoint, $\theta_i$ is the corresponding orientation encoded via sine and cosine components, and $c_i \in \{0,1\}$ is a binary completion flag indicating the termination (``stop'') action. Here, $N$ specifies the prediction horizon, with $N=5$, for example, indicating that the policy outputs waypoints for the next five future steps.

By directly operating in the continuous control space, monolithic frameworks eliminate the need for additional waypoint generation and hierarchical coordination, enabling smoother trajectories and greater adaptability to unseen or unstructured environments. Nevertheless, the lack of explicit high-level abstraction often increases the difficulty of long-horizon reasoning and may lead to compounding errors during extended navigation tasks.

\subsection{Model Paradigms: From Specialists to Generalists}
Early VLN agents primarily adopt discriminative decision policies, selecting actions from a closed action space conditioned on observations and instructions, whereas recent approaches increasingly rely on generative agents that perform open-ended reasoning and action generation. This evolution reflects a broader paradigm shift from task-specific, specialist agents toward more generalist embodied agents with stronger reasoning, abstraction, and transfer capabilities.

\subsubsection{The Discriminative Agent}
Discriminative VLN agents formulate navigation as a direct decision prediction problem, in which the agent predicts the next action, waypoint, or motion parameter conditioned on the current observation. \newcontent{The prediction may be implemented as classification or ranking over a finite candidate set, or as regression in a continuous action space. Formally, such agents learn a policy $\pi_\theta(a_t \mid o_t, h_t, \mathcal{I})$, where $a_t$ may denote a discrete action, a selected waypoint, or continuous motion parameters. For discrete action spaces, the predicted logits are typically normalized via a softmax function to produce a probability distribution, while continuous-action agents may directly regress waypoint coordinates or motion offsets.} A wide range of classical VLN approaches, spanning both hierarchical pipelines and monolithic policies, adopt this formulation. While effective in structured environments with well-defined action spaces, discriminative agents are often tightly coupled to specific tasks, sensor modalities, and training distributions, which constrains their generalization capability and adaptability to more open-ended or dynamic settings.

\subsubsection{The Generative Agent}
\newcontent{Generative VLN agents formulate navigation as a conditional generation problem, where actions, trajectories, subgoals, plans, or reasoning steps are generated through autoregressive decoding, diffusion-based sampling, or other generative processes.} Formally, a generative agent models the conditional distribution
\begin{equation}
p_\phi(\mathbf{y} \mid \mathbf{o}, \mathcal{I}),
\end{equation}
where $\mathbf{y} = (y_1, \ldots, y_L)$ denotes the generated token sequence, $\phi$ represents the parameters of the generative agent model, $\mathbf{o} = (o_1, \ldots, o_T)$ is the temporal sequence of observations. Using an autoregressive factorization, the generation process is defined as:
\begin{equation}
p_\phi(\mathbf{y} \mid \mathbf{o}, \mathcal{I}) = \prod_{m=1}^{L} p_\phi(y_m \mid y_{<m}, \mathbf{o}, \mathcal{I}),
\end{equation}
where $y_{<m} = (y_1, \ldots, y_{m-1})$ denotes the previously generated tokens.
The generated outputs may include high-level plans, textual rationales, or action tokens, which are subsequently grounded into executable actions in the environment. This formulation naturally supports open-vocabulary reasoning and flexible action representations, while also enabling the integration of external knowledge or skill-based execution. \newcontent{We clarify that the generative/discriminative distinction is defined according to the form of the navigation decision-making process, rather than merely by whether an LLM or VLM is employed.}

Training is typically performed via maximum likelihood estimation over reference sequences:
\begin{equation}
\mathcal{L}{\text{gen}} = - \sum_{m=1}^{L} \log p_\phi(y_m^{*} \mid y_{<m}^{*}, \mathbf{o}, \mathcal{I}),
\end{equation}
where $y_m^{*}$ denotes the reference token at the $m$-th index. 
% This supervised objective may be further augmented with reinforcement learning objectives or auxiliary supervision signals to improve task execution and grounding fidelity.
By modeling navigation as a generative process, these agents naturally support zero-shot generalization, instruction following across diverse tasks, and adaptation to previously unseen environments. As a result, generative agents are widely regarded as a key step toward generalist embodied intelligence.

\section{The Hierarchical Framework}
\label{sec_hierarchical_framework}

\subsection{Overview}

\begin{figure}[htbp]
    \centering
    \includegraphics[width=\linewidth]{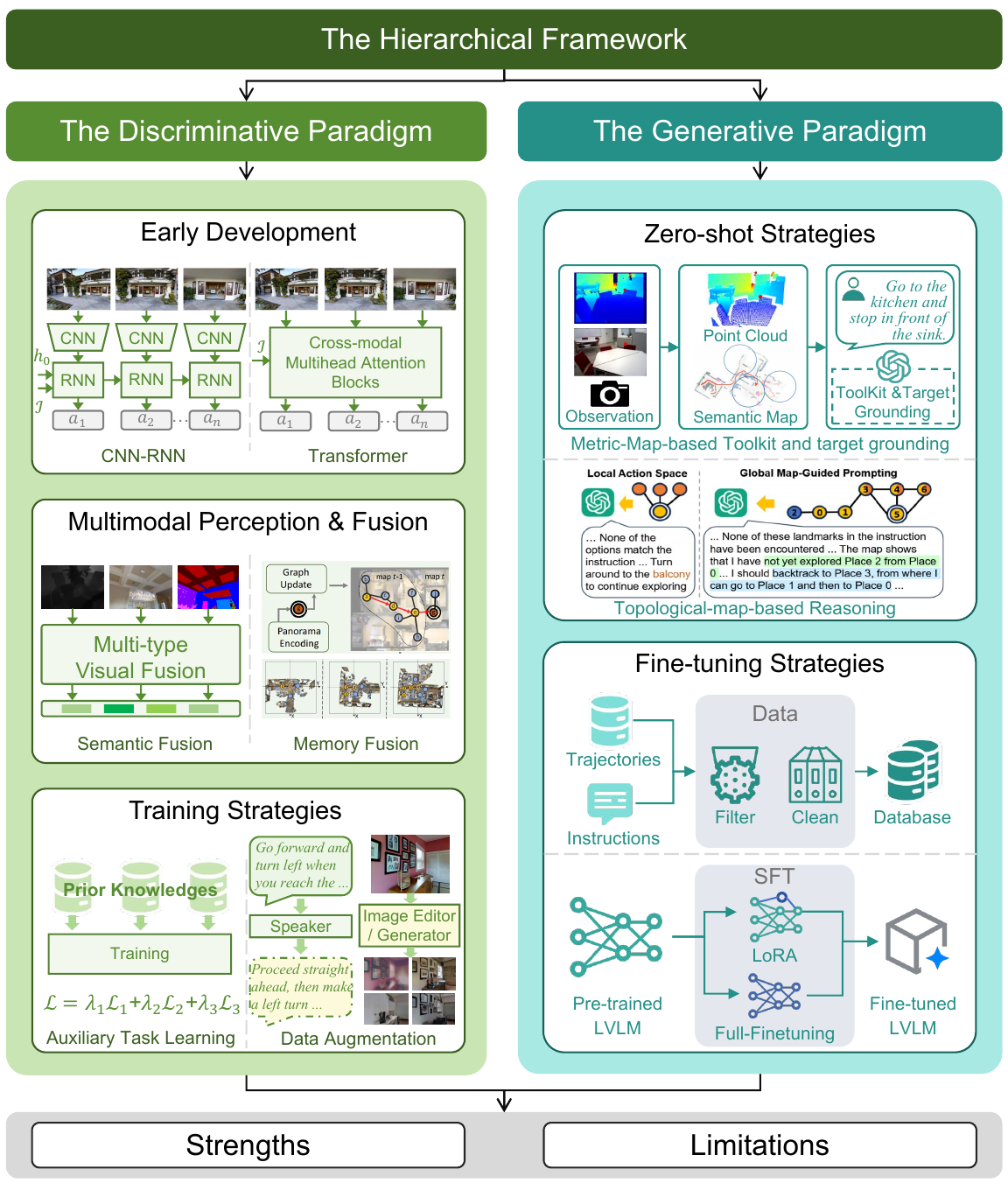}
    \caption{The structure of the hierarchical framework section.}
    \label{fig_hierarchical_framework_structure}
\end{figure}

\begin{figure*}[t]
    \centering
    \includegraphics[width=\linewidth]{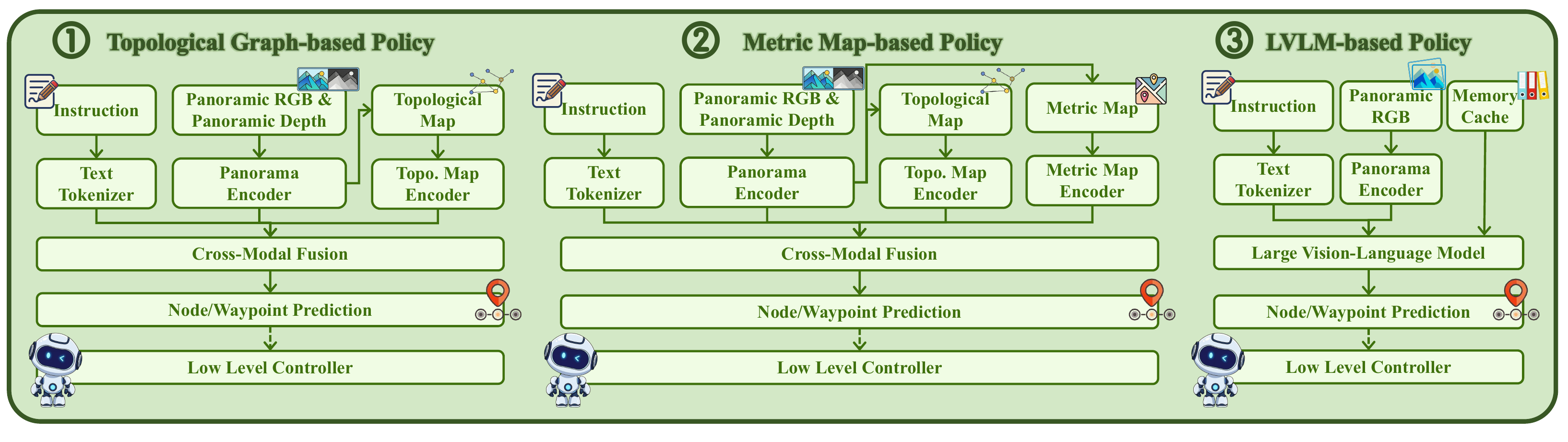}
    \caption{Representative VLN methods under the waypoint-based hierarchical framework. (1) Topological graph-based policies incrementally construct an online topological graph during navigation~\cite{chen2022think,wang2024causal,an2024etpnav}. (2) Metric map-based policies further densify the topological representation into a metric map to guide navigation with finer geometric detail~\cite{liu2023bird,an2022bevbert,wang2023gridmm}. (3) LVLM-based policies employ large vision-language models to encode instructions and visual observations, maintain structured historical memory, and generate navigation plans~\cite{zhou2024navgpt,sathyamoorthy2024convoi,chen2024mapgpt}.}
    \label{fig_method_hierarchical}
\end{figure*}

Contemporary studies typically decompose hierarchical VLN frameworks into two core components: a high-level \textit{planner}, which performs global route reasoning and subgoal selection, and a low-level \textit{controller}, which executes waypoints while handling local motion control and obstacle avoidance. Within this hierarchical taxonomy, two distinct methodological paradigms have emerged: the \textit{discriminative} paradigm, which frames navigation as a candidate selection and ranking problem (usually the task-specific small models), and the \textit{generative} paradigm, which leverages the emergent reasoning capabilities of large general foundation models to synthesize navigation plans and actions. The survey structure of the hierarchical waypoint-based framework is shown in Fig.~\ref{fig_hierarchical_framework_structure}. Additionally, the representative VLN structures under the waypoint-based hierarchical framework are shown in Fig.~\ref{fig_method_hierarchical}.

\subsection{The Discriminative Paradigm}

The discriminative paradigm encompasses methods in which the navigation agent is designed and trained to select the optimal action or waypoint from a predefined set of candidates. As illustrated in Fig.~\ref{fig_timeline}, our statistical analysis shows that the hierarchical discriminative paradigm accounts for approximately 54\% of existing VLN methods, making it the most extensively studied and widely adopted paradigm in the field. Broadly speaking, these methods focus on three core aspects: (i) enhancing semantic perception and cross-modal alignment between vision and language, (ii) strengthening historical memory modeling to support long-horizon, sequential decision-making, (iii) learning strategies to enhance the model's ability to learn from data. In this section, we first review the early evolution of model paradigms and then summarize the core technical components.

\subsubsection{Early Development}
The early evolution of hierarchical architectures in VLN can be characterized by two fundamental progressions: (i) the architectural transition from early CNN-RNN-based approaches to Transformer-based frameworks, and (ii) the environmental generalization from discrete navigation graphs to continuous spatial domains. 

\paragraph{CNN-RNN-based Methods}
The year 2018 marked the formal emergence of VLN as a distinct research task. Anderson \textit{et al.}~\cite{anderson2018vision} formally defined the VLN task and proposed an CNN-RNN-based sequence-to-sequence model with ResNet-152 visual encoding~\cite{he2016deep}, LSTM-based instruction encoding~\cite{hochreiter1997long}, and an attention-equipped recurrent decoder for action prediction. This early baseline achieved only about 20\% SR on the R2R val-unseen split~\cite{anderson2018vision}, underscoring the difficulty of the task. To overcome the limited observability, Fried \textit{et al.}~\cite{fried2018speaker} introduced panoramic visual representations, discretizing the panorama into 36 view angles and employing a soft-attention mechanism over memory to predict waypoint selections at each step. It shows that with the panoramic observation, the SR on val-unseen split can be improved from 20\% to 35\%. Tan \textit{et al.}~\cite{tan2019learning} proposed in 2019 to encode both visual observations and language instructions using bidirectional LSTMs (BiLSTMs). Huang \textit{et al.}~\cite{huang2019multi} introduced a dual-tower multimodal discriminator to align and enhance the input representations. Hu \textit{et al.}~\cite{hu2019you} investigated the relative contributions of visual and linguistic modalities in VLN models and observed that visual inputs play a comparatively limited role in existing approaches.

\paragraph{Evolution brought by Transformers}
With the rise of Transformers~\cite{vaswani2017attention} in natural language processing and computer vision, the dominant VLN architectures have gradually shifted from CNN-RNN pipelines to Transformer-based frameworks~\cite{vaswani2017attention}. Compared to RNNs, Transformers are capable of modeling longer-range and more comprehensive sequence dependencies, offering stronger encoding and decoding capacities. In 2020, Hao \textit{et al.}~\cite{hao2020towards} built a Transformer encoder upon the vision-language pre-trained model LXMERT~\cite{tan2019lxmert}, facilitating more effective multimodal representation learning. However, the fully parallel Transformer decoder introduces substantial computational overhead for long-horizon action prediction. To address this issue, Hong \textit{et al.}~\cite{hong2021vln} proposed the VLN$\circlearrowright$BERT model, which explicitly propagates memory across time steps to enhance historical dependency modeling. Magassouba \textit{et al.}~\cite{magassouba2021crossmap} further introduced a cross-map Transformer with dual back-translation, improving generalization across unseen environments.

\paragraph{Bridging Discrete and Continuous Environments}
VLN was initially formulated under an idealized discrete waypoint setting~\cite{anderson2018vision}. As the field has progressed, increasing attention has been directed toward navigation in continuous environments, where agents must directly operate with fine-grained control signals~\cite{krantz_beyond_2020}. To bridge these two settings, Krantz \textit{et al.}~\cite{krantz2021waypoint} in 2021 introduced a language-conditioned waypoint prediction framework that links discrete high-level decisions with continuous waypoint execution. This work established the foundation of hierarchical discriminative VLN in continuous environments, achieving 36\% SR and 30\% SPL on the VLN-CE val-unseen set.
Hong \textit{et al.}~\cite{hong2022bridging} further proposed a waypoint predictor that generates candidate waypoints online, allowing discrete high-level navigation policies to be transferred to and trained within continuous environments. By translating VLN$\circlearrowright$BERT~\cite{hong2021vln} into the VLN-CE setting, their approach improved the SR from 36\% to 44\% on the val-unseen set.
ETPNav~\cite{an2024etpnav} advanced this line of research by performing online topological mapping through self-organized predicted waypoints. Leveraging a transformer-based cross-modal planner~\cite{chen2022think}, ETPNav generates navigation plans conditioned on both global topological maps and language instructions, achieving 57\% SR on the R2R-CE val-unseen benchmark. 
As VLN has evolved from discrete to continuous environments, an increasing number of recent methods have focused on designing models specifically tailored to these more challenging continuous settings in order to achieve improved performance.

\subsubsection{Cross-modal Perception and Fusion}
Cross-modal Perception primarily concerns the semantic alignment and fusion of both visual and linguistic modalities. 

\paragraph{Semantic Representations}
Beyond architectural advances, enhancing semantic representations is crucial for improving navigation performance, as VLN agents fundamentally rely on effective understanding of visual observations and natural language instructions.
Some methods have demonstrated the advantages of semantic features for vision-based navigation~\cite{dang2022unbiased,dang2023search}. ORIST~\cite{qi2021road} and SOAT~\cite{moudgil2021soat} proposed to concatenate object-level features with the scene-level features and learn them through the transformer encoder in a parallel way. BiasVLN~\cite{zhang2021diagnosing} observed that the low-level image features result in environmental bias. SEvol~\cite{chen2022reinforced} utilized a graph-based method to construct relationships of objects. OAAM~\cite{qi2020object} and NvEM~\cite{an2021neighbor} applied independent soft attention to text embeddings to learn the object and action representations of instructions, and the latter also considers the neighboring objects of candidates. ADAPT~\cite{lin2022adapt} suggested using the CLIP~\cite{radford2021learning} with an action prompt to improve the action-level modality alignment. DSRG~\cite{wang2023dual} proposed a dual semantic-aware network to boost the significant semantic features from both visual and linguistic modalities. 
Fang \textit{et al.}~\cite{fang2025hierarchical} proposed a hierarchical semantic-augmented navigation (HSAN) framework, which constructed a dynamic hierarchical semantic scene graph to capture object-region-zone representations. 
For enhancing instruction-level semantics, FGR2R~\cite{hong2020sub} proposed an attention-based sub-instruction segmentation strategy and introduced a fine-grained progress annotation dataset. GELA~\cite{cui2023grounded} enhanced vision-language alignment through grounded entity-landmark adaptive pre-training.
With the success of contrastive vision-language pre-training models such as CLIP~\cite{radford2021learning}, several works have explored CLIP-based VLN methods. Sheng \textit{et al.}~\cite{shen2021much} investigated the impact of CLIP representations on VLN performance, CLIP-Nav~\cite{dorbala2022clip} achieved zero-shot navigation using CLIP, and CITL~\cite{liang2022contrastive} proposed a contrastive instruction-trajectory learning framework to improve generalization.

Beyond conventional RGB inputs, depth and other auxiliary visual cues have been shown to substantially enhance environmental perception. SEAT~\cite{wang2024enhanced} and MV-Typo~\cite{liu2024multiple} jointly integrates panoramic RGB, depth, and semantic features to represent the environment, while GeoVLN~\cite{huo2023geovln} further incorporates geometric information for more structured scene modeling. ETPNav~\cite{an2024etpnav} empirically demonstrated that augmenting RGB observations with depth significantly improves reasoning performance. DG-AdaIN~\cite{sun2021depth} proposed a depth-guided adaptive instance normalization module that modulates RGB features conditioned on depth information.

While explicitly modeling semantic information can enrich perceptual representations, it may also introduce biases stemming from spurious correlations. To mitigate such confounding effects, Parvaneh \textit{et al.}~\cite{parvaneh2020counterfactual} proposed learning from both real observations and generated counterfactual environments to eliminate spurious features that may bias the agent's decisions. GOAT~\cite{wang2024causal} represented the first efforts to incorporate causal inference into VLN by introducing end-to-end backdoor and front-door adjustment modules, coupled with a cross-modal feature pooling pre-training objective. DICCR~\cite{zhou2025diccr} further proposed a dual-gated intervention and confounder causal reasoning framework to improve the generalization.

Although most hierarchical methods assume panoramic visual inputs, some works investigate how to obtain broader visual observations under limited field-of-view settings. g3d-LF~\cite{wang2025g3d} and monoVLN~\cite{lu2025monovln} addressed the constraints of restricted viewpoints by leveraging NeRF~\cite{mildenhall2020nerf} or 3D Gaussian Splatting-based (3DGS)~\cite{kerbl2023gaussians} rendering techniques to augment monocular RGB-D perception for VLN.

\paragraph{Historical Memory Representations} 
Spatial memory is particularly critical for embodied agents, especially when dealing with long, complex tasks. Within hierarchical VLN frameworks, advanced memory encoding strategies can be broadly categorized into two types: \textit{topological graph memory} and \textit{metric map memory}.

In the early age, some efforts utilized recurrent neural networks (\textit{e.g.,} LSTM) to update states~\cite{wang2020vision,an2021neighbor,dang2022unbiased,tan2024self}. VLN$\circlearrowright$BERT~\cite{hong2021vln} integrated the recurrent units on an encoder-only Transformer structure. 
HAMT~\cite{chen2021history} and HOP~\cite{qiao2022hop,qiao2023hop_plus} proposed treating historical memories as token sequences, engaging in the self-attention operations of memory modules. 
As a seminal work on topological graph memory, DUET~\cite{chen2022think} encoded visited nodes into a structured graph representation and significantly enhances spatial memory and reasoning through both local and global cross-modal attention mechanisms. This framework has since become a common baseline for subsequent studies and has influenced the continued development of hierarchical architectures in both discrete and continuous VLN settings. 
TraceNav~\cite{zhu2025history} introduced a history-traceable framework with multi-granularity instruction-trajectory alignment. MG-VLN~\cite{zhang2024mg} demonstrated that explicitly maintaining long-term memory maps significantly benefits multi-goal and extended navigation tasks. To further improve semantic reasoning within memory structures, MFRA~\cite{yue2025think} proposed a hierarchical multimodal fusion architecture that aggregates visual and linguistic cues across multiple abstraction levels, and VLN-KHVR~\cite{kong2025vln} incorporated knowledge- and history-aware visual representations by retrieving relevant external knowledge while filtering instruction-irrelevant observations. Wen \textit{et al.}~\cite{wen2024vision} proposed a cross-modal feature fusion approach based on a history-aware attention mechanism. Building on structured memory representations, VLN-EventKG~\cite{zhao2024towards} presented a prompt-based framework for extracting event-centric knowledge graphs and integrated a dynamic history-aware correction mechanism.

Besides topological graphs, several methods have explored incorporating dense metric maps, typically represented as bird's-eye-view (BEV) scene graphs, to model global spatial memory. CM$^2$~\cite{georgakis2022cross} and WS-MGMap~\cite{chen2022weakly} enhanced map representations with a multi-granularity semantic map. With weakly supervised auxiliary tasks, its map representations better capture object-level information, allowing the agent to more effectively localize instruction-relevant objects.
BSG~\cite{liu2023bird} and BEVbert~\cite{an2022bevbert} introduced a hybrid top-metric map, where a topological map is used for long-term planning and a metric map for short-term reasoning. GridMM~\cite{wang2023gridmm} built a top-down egocentric and dynamically growing grid memory map to structure the visited environment. g3d-LF~\cite{wang2025g3d} and Dynam3D~\cite{wang2025dynam3d} introduced a generalizable 3D-Language feature fields for allowing generations of BEV maps from any position in the 3D scene, and querying targets using multigranularity language. 
VER~\cite{liu2024volumetric} proposed a volumetric environment representation that voxelizes the physical world into structured 3D cells, enabling the agent to jointly predict 3D occupancy, room layout, and bounding boxes.
OVL-MAP~\cite{wen2025ovl} introduced an online visual-language mapping framework in which an incrementally built global map serves both as environmental memory and as an explicit decision component for waypoint selection during navigation.

\subsubsection{Training Strategies}

Data-driven deep learning approaches form the foundation of VLN systems. A wide range of training strategies have been explored to improve learning efficiency and generalization.

\paragraph{Auxiliary Task Learning}
Auxiliary tasks designed based on expert knowledge and commonsense can provide additional supervisory signals, thereby improving learning efficiency and robustness~\cite{hernandez2019agent,lin2019adaptive}. Ma \textit{et al.}~\cite{ma2019self} introduced a self-monitoring navigation framework with auxiliary progress estimation to track instruction execution. Building on this, Ma \textit{et al.}~\cite{ma2019regretful} further proposed regretful navigation, replacing impractical beam search with a learnable exploration strategy.
Hong \textit{et al.}~\cite{hong2020sub} proposed sub-instruction segmentation to provide fine-grained supervision during training. Zhu \textit{et al.}~\cite{zhu2020vision} introduced AuxRN, an auxiliary reasoning navigation framework that optimizes additional reasoning objectives: explaining past actions, predicting navigation progress, forecasting the next action, and evaluating trajectory consistency.
Wang \textit{et al.}~\cite{wang2020active} explored active exploration strategies, focusing on when and where to explore, what information to collect during exploration, and how to adjust navigation behavior afterward. Zhou \textit{et al.}~\cite{zhou2021rethinking} reformulated VLN as a node classification problem on navigation graphs and demonstrated that incorporating spatial priors from path structures significantly improves performance. Kou \textit{et al.}~\cite{kuo2023structure} proposed a series of structured encoding auxiliary tasks (SEA) to enhance visual feature representations. Zhao \textit{et al.}~\cite{zhao2022target} introduced target location prediction as an auxiliary task, enabling agents to navigate more effectively toward the goal. HSPR~\cite{xu2024hierarchical} introduced a scene understanding auxiliary task to help the agent build knowledge base of hierarchical spatial proximity. 
Tan \textit{et al.}~\cite{tan2025source} proposed an elastic adaptation model (EAM) that uses online testing samples to adapt the auxiliary decision model to new environments.
To improve both effectiveness and efficiency, MAGIC~\cite{wang2024magic} proposed a meta-ability knowledge distillation strategy combined with an interactive chain-of-distillation mechanism, achieving enhanced performance while requiring only 5\% of the parameters to be trainable.

% ATENA~\cite{ko2025active} introduced a test-time active learning framework that leverages episodic human feedback to resolve uncertainty in navigation decisions, enabling an agent to evaluate its navigation outcomes based on confident predictions. 

\paragraph{The multitask learning framework}

In recent years, diverse navigation tasks have emerged, ranging from high-level goals~\cite{qi2020reverie,thomason2020vision,Zhu_2021_SOON} to fine-grained directives~\cite{ku2020room,jain2019stay,gao2025openfly}. Most are studied in isolation, with task-specific methods that generalize poorly. This limitation motivates multitask learning approaches that train unified models across multiple VLN tasks. 
MT-RCM~\cite{wang2020environment} proposed a multitask navigation model with a environment-agnostic representations that are invariant among the environments. 
Liu \textit{et al.}~\cite{liu2024volumetric} found that multi-task learning yields a substantial performance gain for navigation.
SAME~\cite{zhou2025same} proposed a state-adaptive mixture of experts model that effectively enables an agent to infer decisions based on different-granularity language, achiving improved performance among 6 different VLN tasks.

\paragraph{Data Augmentation Strategies}

A defining characteristic of VLN, compared with traditional navigation tasks, is its strong reliance on data-driven learning. This motivates extensive research on data augmentation to improve generalization. Overall, existing effective strategies broadly fall into \textit{follower-speaker architectures} and \textit{image generation-based approaches}.

The follower-speaker paradigm was first introduced by Fried \textit{et al.}~\cite{fried2018speaker}, where a Speaker model generates navigation instructions and a Follower model executes navigation based on these instructions. The two models are jointly optimized through mutual evaluation and interaction. Tan \textit{et al.}~\cite{tan2019learning} further proposed an environmental dropout strategy that, when combined with the Speaker, enables the online generation of diverse instructions. A series of works have focused on improving Speaker models, exploring better cross-modal alignment and instruction diversity~\cite{tan2019learning,magassouba2021crossmap,dou2022foam,liang2022contrastive,wang2023pasts,wang2023res,wang2023lana,gopinathan2024spatially}.

Building upon this paradigm, PREVALENT~\cite{hao2020towards} and Marky~\cite{wang2022less,kamath2023new} leveraged the Speaker models to randomly sample trajectories in Matterport3D~\cite{chang2017matterport3d}, enabling large-scale pretraining.
\newcontent{FCA-NIG~\cite{cui2025generating} introduced a fine-grained cross-modal alignment framework for generating VLN instructions, which combines R2R-like instruction synthesis, CLIP-based entity selection, and sub-instruction-trajectory pair generation.}
ScaleVLN~\cite{wang2023scaling} further extended Speaker-based data generation to 1,291 additional indoor scenes~\cite{ramakrishnan2021habitat,xia2018gibson}, producing a large corpus of high-quality pretraining data and boosting the SR on the R2R val-unseen set to 81\%. This dataset has since become a foundational resource for training large VLN models. More recently, Wang \textit{et al.}~\cite{wangbootstrapping} further proposed SRDF, a data flywheel framework that employs the Mantis-8B-SigLIP model~\cite{jiangmantis} as the Speaker and uses a navigation model to evaluate the quality of generated data. Through iterative generation, evaluation, and retraining, this approach enabled the DUET baseline~\cite{chen2022think} to achieve performance on the R2R Test-Unseen set (85\% SR) that is highly comparable to human performance, underscoring the critical role of data augmentation in improving model generalization.

Beyond instruction synthesis, image generation-based augmentation has emerged as another important direction, which is very similar to the advanced world model paradigm~\cite{bar2025navigation}. To mitigate overfitting caused by limited training environments, researchers employ image generation and editing techniques to diversify visual observations. Representative approaches include using generative adversarial networks (GANs)~\cite{li2022envedit} for style transfer and scene editing, leveraging world models~\cite{yao2025navmorph,koh2021pathdreamer,wang2023dreamwalker,wang2024lookahead,bar2025navigation,zhu2023vision} to predict plausible future observations, and adopting diffusion models~\cite{li2023panogen} to synthesize realistic panoramic views. 
VLN-MP~\cite{hong2024only} integrates both natural language and images into navigation instructions, demonstrating consistent performance improvements across a range of baseline methods through visual prompting.
In addition, several works expand VLN training data by mining large-scale Internet videos, such as indoor walkthroughs, to construct diverse navigation environments beyond curated simulators~\cite{majumdar2020improving,chen2022learning,jia2021scaling,cheng2024navila}.

\subsection{The Generative Paradigm}
Since the release of ChatGPT~\cite{openai2022chatgpt} in 2022, the strong generative and reasoning capabilities of large-scale foundation models have demonstrated substantial promise for VLN. The field has progressively shifted from discriminative architectures built upon BERT-style encoders~\cite{devlin2019bert} with task-specific classification heads toward fully generative, GPT-style frameworks~\cite{ouyang2022instructgpt}. Broadly, existing approaches that incorporate large models into VLN can be categorized into two paradigms: \textit{zero-shot} inference and \textit{fine-tuning}.

\subsubsection{Zero-shot Strategies}

Owing to the strong world knowledge and commonsense reasoning capabilities inherent in large models, many studies focus on designing more effective system prompts to better elicit their zero-shot potential.

\paragraph{Occupancy-map-based Inference}
Some methods integrate LLMs with explicit spatial representations, such as occupancy maps, to address the VLN problem.
\newcontent{VLMaps~\cite{huang23vlmaps} used LLMs to translate natural-language commands into sequences of open-vocabulary navigation goals that are directly localized in the map for sequential execution, while leveraging the code-writing capability of LLMs to generate executable Python code for robot skill composition and task execution.}
Similarly, LLM-Grounder~\cite{yang2024llm} uses LLMs to generate deliberate plans and interact with tools, such as target and landmark finders, to collect task-relevant information.
CoNVOI~\cite{sathyamoorthy2024convoi} introduced a multimodal visual marking approach that annotates numbered regions within the visual frame and queries a large vision-language model to construct a reference navigation path based on the selected annotations.
InstructNav~\cite{long2024instructnav} introduces the multi-sourced value maps to model key elements to help the dynamic chain-of-navigation for planning under the zero-shot setting. 

\paragraph{Topological-map-based Inference}
Deguchi \textit{et al.}~\cite{deguchi2024language} proposed a topological map style that can be created from only natural language, and used the LLM to estimate the navigation path based on the instruction.
NavGPT~\cite{zhou2024navgpt} performed a zero-shot sequential action prediction for VLN. The visual observation, navigation history, and future explorable directions are working as inputs to help the LLM to make the decision for the target place. 
MapGPT~\cite{chen2024mapgpt} proposed to build a topological map for storing the structured memory for VLM in boosting VLN decision-making. Zhang \textit{et al.}~\cite{zhang-etal-2025-vision-language} further improved the LLM's contextual understanding by incorporating textual descriptions from multiple perspectives that facilitate analogical reasoning across images.
OVER-NAV~\cite{zhao2024over} proposed to incorporate LLMs and open-vocabulary detectors to distill key information and establish correspondence between multi-modal signals for improving the iterative VLN performance.
OpenNav~\cite{qiao2025open} explored open-source LLMs for zero-shot VLN-CE. It combines different LLM for different tasks to collaboratively perform perception and reasoning.
AO-Planner~\cite{chen2025affordances} proposed an affordances-oriented planning framework, by employing a visual affordances prompting, where visible ground is segmented using SAM to provide navigational affordances.
SmartWay~\cite{shi2025smartway} addressed zero-shot VLN by enhancing waypoint prediction with history-aware reasoning and backtracking mechanisms, allowing agents to recover from navigation errors and improve path robustness in unseen environments. 
CityNavAgent~\cite{zhang2025citynavagent} proposed a LLM-empowered agent and designed a hierarchical semantic planning module (HSPM) that decomposes the long-horizon task into sub-goals with different semantic levels.
Chen \textit{et al.}~\cite{chen2025constraint} proposed CA-Nav, a zero-shot VLN-CE framework that decomposes navigation into sequential constraint-aware sub-instruction completion, using a sub-instruction manager and a value mapper to track and satisfy constraints.
CLASH~\cite{wang2025clash} introduced a hybrid collaborative framework with an uncertainty-aware collaboration module that dynamically fuses task-specific small models and general-purpose large models.
\newcontent{SFCo-Nav~\cite{xiong2026sfco} introduced a zero-shot VLN framework with slow-fast collaboration, where a slow LLM planner constructs subgoal-level imagined object graphs, a fast navigator executes subgoals online, and an asynchronous confidence-based bridge invokes replanning only when needed.}

\subsubsection{Fine-tuning Strategies}

With sufficient training resources, recent works pursue deeper navigation-specific adaptation of large-scale models. 
NavHint~\cite{zhang2024navhint} proposed a hint generator that produces detailed visual descriptions to support decision making toward the target location.
Building upon NavGPT~\cite{zhou2024navgpt}, NavGPT-2~\cite{zhou2024navgpt2} further incorporates a large vision-language model and a learnable navigation policy network. After fine-tuning, the SR and SPL on the R2R val-unseen set rise from 34\% to 72\%, and from 29\% to 61\%, respectively. 
FLAME~\cite{xu2025flame} proposed a multimodal LLM-based agent, including single perception tuning for route summarization, and end-to-end training on urban VLN datasets. 
NaviLLM~\cite{zheng2024towards} adapted LLMs to embodied navigation by introducing schema-based instruction, enabling to integrate diverse data sources from various datasets during training and inference.
NavCoT~\cite{lin2025navcot} let the LLM act as a world model to imagine the next observation according to the instruction, and select the candidate observation that best aligns with the imagination.
InternVLA-N1~\cite{internvla-n1} and DualVLN~\cite{wei2025ground} proposed a dual-system design that combines diffusion-based point-goal control with high-level pixel-wise waypoint prediction. System~2 performs VLM-based pixel-goal grounding, while System~1 leverages the latent goal representation produced by System~2 to generate trajectory waypoints.
By coupling large-scale pretraining with navigation-specific inductive biases, these methods achieve substantial performance improvements over zero-shot VLN inference.
\newcontent{AgentVLN~\cite{xin2026agentvln} proposed a deployable VLM-as-Brain framework with a plug-and-play skill library and self-correction mechanisms for robust long-horizon navigation.}

\subsection{Strengths and Limitations}

\paragraph{Strengths}
Hierarchical waypoint-based frameworks decompose VLN into high-level decision making and low-level execution, offering several notable advantages.

First, by explicitly separating global planning from local control, these methods allow the high-level policy to focus on instruction understanding, global reasoning, and waypoint selection, while delegating local path following and obstacle avoidance to a dedicated controller. This decomposition significantly reduces learning complexity and improves training stability, especially for long-horizon navigation tasks.

Second, the high-level decision module typically operates over a discrete and structured action space (\textit{i.e.}, candidate waypoints), enabling effective learning with relatively modest training data and computational resources. As a result, such frameworks are more accessible for research teams with limited resources and are easier to train and deploy compared to fully end-to-end continuous control models.

Third, the modular design enhances interpretability and debuggability. Errors can be localized to either the high-level planner or the low-level controller, facilitating targeted diagnosis and improvement. This property is particularly valuable for real-world robotic deployment, where system transparency and reliability are critical.

Finally, hierarchical waypoint-based frameworks naturally align with graph-based environments and panoramic observations, making them well suited for standard VLN benchmarks and enabling strong empirical performance.

\paragraph{Limitations}
Despite their advantages, hierarchical waypoint-based frameworks also exhibit inherent limitations.

First, modularization introduces the risk of error accumulation and suboptimal coordination between modules. Inaccuracies in high-level waypoint prediction may propagate to the low-level controller, and errors introduced during execution are not always recoverable by subsequent planning steps, hindering end-to-end optimization.

Second, overall navigation performance is highly dependent on the quality of waypoint generation and the robustness of the low-level execution module. In cluttered or dynamic environments, failures in obstacle avoidance or motion control can negate the benefits of accurate high-level planning.

Third, many hierarchical approaches rely on panoramic or multi-view visual inputs to generate reliable waypoint candidates. This assumption limits their applicability in real-world settings where only monocular or narrow field-of-view sensors are available.

Finally, the reliance on predefined waypoint representations or environment graphs can restrict flexibility and generalization. Transferring such frameworks to continuous, unstructured, or previously unseen environments often requires non-trivial adaptation of the waypoint generation mechanism.

\section{The Monolithic Framework}
\label{sec_monolithic_framework}

\subsection{Overview}
\begin{figure}[h]
    \centering
    \includegraphics[width=\linewidth]{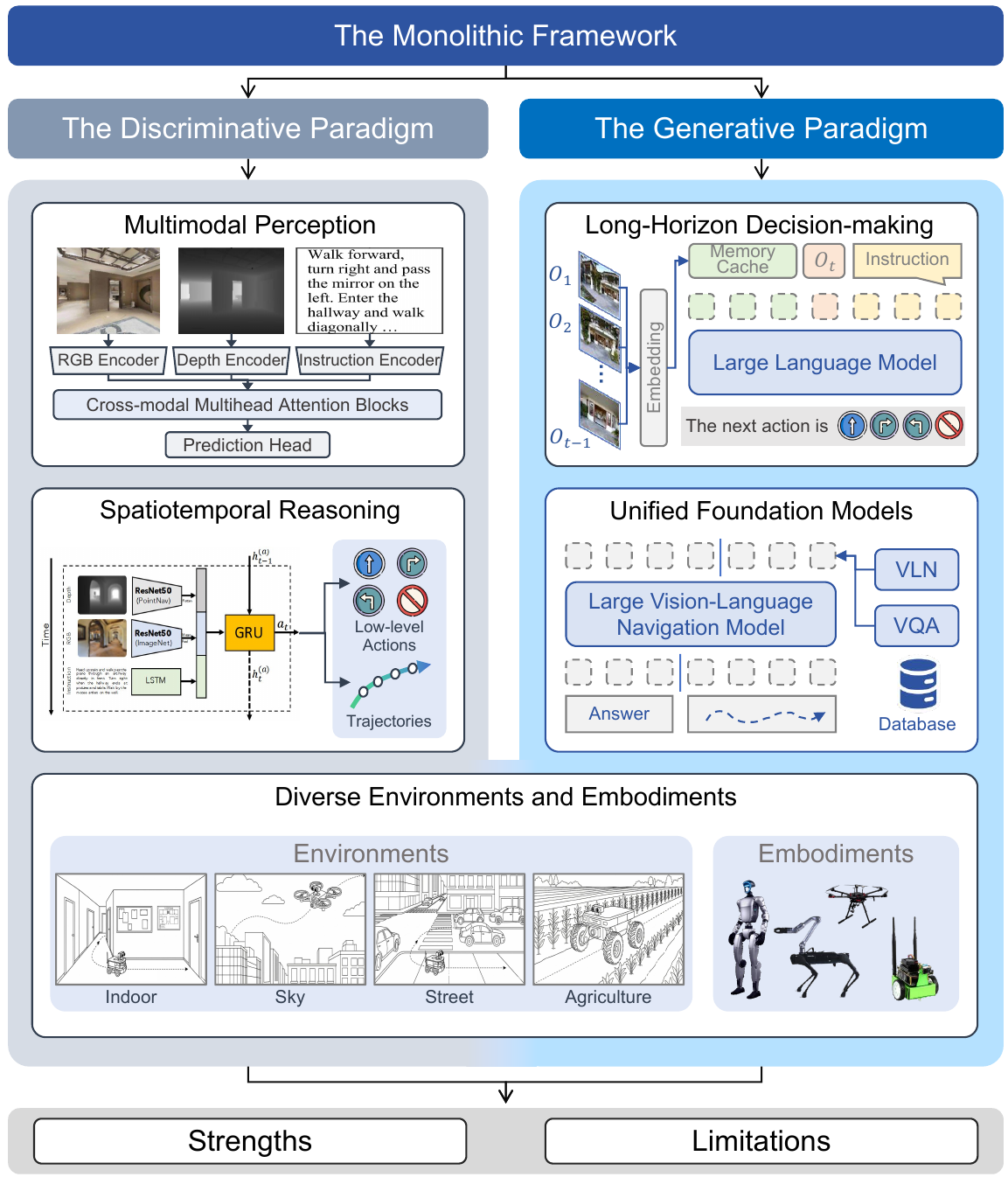}
    \caption{The structure of the monolithic framework section.}
    \label{fig_monolithic_framework_structure}
\end{figure}

\begin{figure*}[t]
    \centering  
    \includegraphics[width=0.9\linewidth]{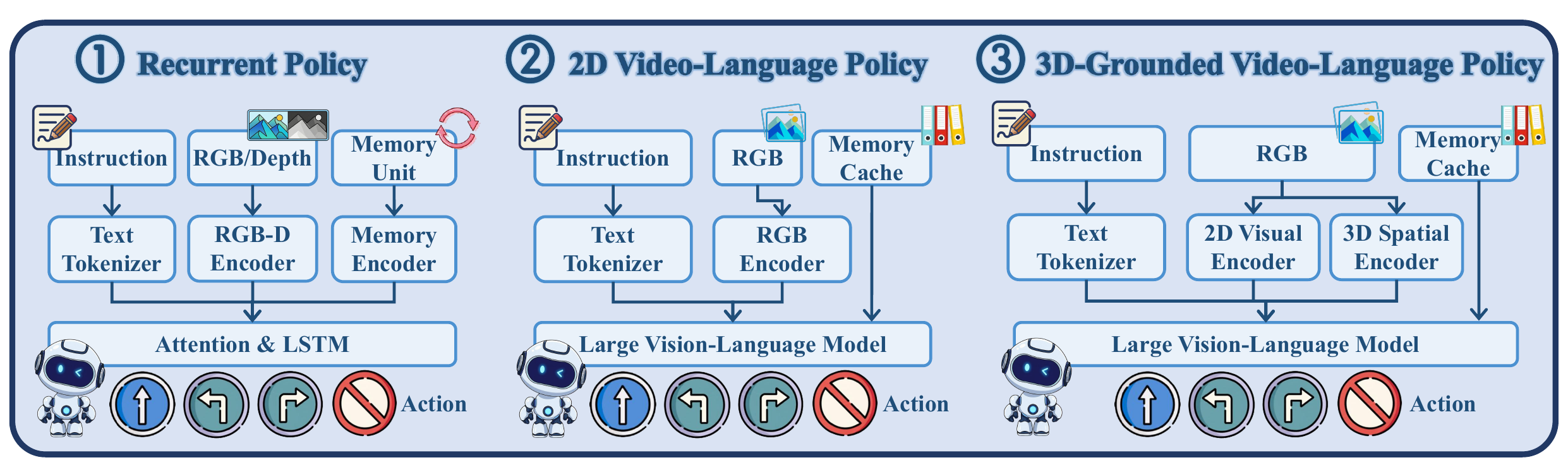}
    \caption{Representative VLN methods under the monolithic action-based framework. (1) Recurrent policies employ recurrent neural networks to encode historical states and maintain temporal memory~\cite{krantz_beyond_2020,he2023mlanet,he2024mee}. (2) 2D video-language policies treat sequential visual observations as video streams, leveraging large vision-language models for observation encoding and KV-cache mechanisms to store history and accelerate inference~\cite{zhang2024navid,zhang2024uni,cheng2024navila}. (3) 3D-grounded video-language policies further incorporate explicit 3D encoding networks to enhance spatial perception and long-term geometric memory~\cite{zeng2025janusvln,navid4d,xue2025omninav}.}
    \label{fig_method_monolithic}
\end{figure*}

The monolithic VLN framework refers to models that bypass intermediate decision-making, instead directly outputting control signals for robotic motion.
Under this definition, a VLN model acts as a sophisticated function mapping visual and linguistic observations onto a specific action space.
To implement this, early methods employed classifier-based agents, referred to here as the \textit{discriminative} paradigm.
In this paradigm, the action space is restricted to a discrete, limited set.
While this approach simplifies task definition, data collection, and model design, it suffers from drawbacks such as limited execution flexibility and generalization.
With the advances of LLMs and LVLMs, a new \textit{generative} paradigm has emerged.
In this paradigm, a language model is utilized to generate navigation actions for the robot; that is, action prediction is treated as a text generation task.
The advantages are significant: LLMs possess massive parameter counts and are pre-trained on diverse datasets, granting them rich common-sense knowledge and reasoning capabilities.
Furthermore, their open-ended output format enables more flexible action spaces.
The survey structure of the monolithic framework is shown in Fig.~\ref{fig_monolithic_framework_structure}. Additionally, the representative VLN methods under the monolithic framework are shown in Fig.~\ref{fig_method_monolithic}.

\subsection{The Discriminative Paradigm}
Framed as a classification task over a predefined action space, models within the discriminative paradigm are typically compact and specialized.
Current efforts are mainly engaged in two enduring research trajectories: the enhancement of multimodal perception and the integration of spatiotemporal reasoning into decision-making.

\subsubsection{Multimodal Perception}

CMA~\cite{krantz_beyond_2020} represented one of the pioneering efforts to address VLN in continuous environments.
It employs cross-modal attention modules to align visual and linguistic observations, enabling the agent to ground textual instructions in the surrounding environment.
Building upon CMA, LAW~\cite{raychaudhuri2021language} refined the training supervision paradigm.
It argues that standard shortest-path supervision often deviates from the provided instructions and thus introduces a language-aligned supervision scheme to facilitate more coherent action prediction.
MLANet~\cite{he2023mlanet} also emphasized fine-grained alignment but shifts the focus toward structural granularity.
By leveraging manageable sub-instructions and employing a multi-level attention mechanism, it establishes a tighter correspondence between dynamic visual observations and linguistic guidance, thereby enhancing the agent's fidelity in following instructions.
VISITRON~\cite{visitron} introduced a multimodal Transformer that aligns object-level visual semantics with dialogue context, enabling interactive VLN agents to ground language in visual perception while handling dynamically evolving dialog history.
MEE~\cite{he2024mee} proposed a unified encoder for four modalities and adopted an evolutionary three-stage pre-training strategy to bridge the gap between complex instructions and continuous environments.
This multi-stage adaptation facilitates robust cross-modal alignment and enhances the agent's generalization to unseen scenes.
While these methods enhance perception, predicting low-level actions in continuous environments remains challenging.
As pointed out by Subedi \textit{et al.}~\cite{subedi2025can}, pretrained multimodal embeddings alone are insufficient to fully support navigation tasks, indicating the necessity of improving state-aware navigation components.
Consequently, researchers have also explored leveraging long-horizon information to improve decision-making.

\subsubsection{Spatiotemporal Reasoning}

As a long-horizon task, VLN requires an agent to remember previously traversed rooms, observed objects, and completed sub-goals, as well as to recall past experiences to accelerate exploration and correct navigation errors.
Compared to the relatively simple recurrent modeling (\textit{e.g.}, GRU-based memory~\cite{cho2014learning}) adopted in early CMA~\cite{krantz_beyond_2020}, a series of approaches have been proposed to explicitly enhance spatiotemporal reasoning.
For instance, SASRA~\cite{irshad2021sasra} employed a hybrid transformer-recurrence model that combines classical semantic mapping with cross-modal attention, enabling the agent to reason over spatiotemporal cues.
DifNav \cite{shi2025dagger} unified waypoint prediction and planning via a conditional diffusion model, using DAgger~\cite{ross2011reduction} for online training to improve long-horizon spatial reasoning.
CVLN-Think~\cite{liu2025cvln} introduced causal inference into continuous VLN by explicitly reasoning along the observation-action chain.
It generates counterfactual visual observations to learn invariant spatial structures and applies causal intervention to debias action decisions.
% A\textsuperscript{2}Nav~\cite{chen20232_a2nav} uses LLMs to decompose complex instructions into sub-tasks and select skill-specific policies for low-level action prediction, representing an early LLM-based VLN approach further discussed in Section~\ref{sec:monolithic:generative}.
NavFoM~\cite{zhang2025embodied} introduces a temporal-viewpoint indicator (TVI) to encode viewpoint awareness, temporal awareness, and viewpoint separability, enabling flexible adaptation to arbitrary camera configurations. Qwen-RobotNav~\cite{zhang2026qwen} develops an agentic navigation system built upon Qwen3-VL, demonstrating strong cross-task and cross-embodiment generalization. Since both methods predict the next navigation action with an MLP-based head, we categorize them as discriminative rather than generative approaches.

\subsubsection{Diverse Environments and Embodiments}
Beyond improving foundational VLN capabilities, recent research has explored VLN in more diverse environments and across multiple embodiments.
IVLN~\cite{krantz2023iterative} introduced a persistent environment paradigm in which agents maintain memory across sequences of R2R episodes, demonstrating that map-based agents can leverage environmental persistence to improve performance.
VLN-PE~\cite{wang2025rethinking} highlighted the challenges of physically embodied deployment by systematically evaluating humanoid, quadruped, and wheeled robots, revealing performance degradation due to limited observation space, lighting variations, and locomotion constraints.
SINGER~\cite{adang2025singer} trains a lightweight, end-to-end visuomotor policy using onboard sensing and photorealistic simulation, achieving strong zero-shot sim-to-real transfer for open-vocabulary drone navigation.

In outdoor scenarios, ArraMon~\cite{kim-etal-2020-arramon} introduced a joint navigation-and-assembly task that extends VLN beyond goal reaching, requiring agents to sequentially follow natural language instructions to navigate, collect target objects, and subsequently assemble them into specified configurations in dynamic outdoor environments.
MTST~\cite{zhu2021multimodal} addressed the scarcity of human-annotated instructions by augmenting datasets with stylistically transferred instructions, thereby improving task completion rates.
Similarly, ORAR~\cite{schumann-riezler-2022-analyzing} demonstrated a reliance on environment-specific graph features, motivating the development of larger-scale and more geographically diverse benchmarks.
Loc4Plan~\cite{loc4plan} emphasized spatial localization prior to planning, demonstrating that explicit position grounding enhances outdoor navigation performance.

For aerial navigation, AerialVLN~\cite{liu2023aerialvln} and CityNav~\cite{Lee_2025_ICCV} introduced UAV-based VLN tasks and datasets, highlighting the importance of 3D spatial reasoning and semantic map representations for urban navigation.
Safety-critical control is partially addressed by ASMA~\cite{Sanyal2025ASMA}, which integrates scene-aware Control Barrier Functions for collision avoidance in drone navigation, while GRaD-Nav++~\cite{chen2025grad} demonstrated fully onboard, language-guided flight using Gaussian Radiance Fields and differentiable dynamics, achieving robust generalization across simulated and real-world environments.

Collectively, these studies demonstrated the growing emphasis on embodiment diversity, environmental complexity, and operational realism in VLN research.

\subsection{The Generative Paradigm}
\label{sec:monolithic:generative}
With the emergence of LLMs and multimodal foundation models, VLN has gradually shifted from discriminative policy learning toward a generative paradigm.
Instead of predicting actions via task-specific heads or handcrafted intermediate representations, recent approaches formulate VLN as an autoregressive decision-making process.
Navigation actions are generated in a language-like or tokenized form, minimizing reliance on explicit maps, waypoints, or auxiliary sensors.
This paradigm enables stronger generalization, flexible task formulation, and potential to unify different navigation tasks.

\subsubsection{Long-Horizon Decision-making}
As VLN inherently involves long-horizon decision-making in partially observed environments, recent generative approaches primarily focus on enhancing reasoning capabilities for navigation tasks.
RDP~\cite{wang2025rethinking} proposed a recurrent diffusion policy to predict dense future trajectories. 
VLN-R1~\cite{qi2025vln} and ActiveVLN~\cite{zhang2025activevln} enhanced long-horizon decision-making through reinforcement learning.
Specifically, VLN-R1 introduces reinforcement fine-tuning for LVLM-based VLN, enabling continuous action prediction from egocentric observations without relying on discrete topological graphs.
ActiveVLN~\cite{zhang2025activevln}, in contrast, proposes a multi-turn reinforcement learning framework that supports active environment interaction and self-collected trajectory generation, allowing generative agents to surpass the limitations of expert-demonstration-only training.

Supervised fine-tuning approaches also contribute to reasoning improvements.
CAST~\cite{glossop2025cast} enhanced instruction-following via counterfactual relabeling, augmenting existing datasets with diverse and fine-grained language–action pairs without additional data collection.
NaVid~\cite{zhang2024navid} proposed a video-based vision-language model that predicts actions solely from a monocular RGB video stream and instructions, without requiring maps, depth, or odometry.
By encoding long-term visual history as spatiotemporal context, NaVid demonstrates strong cross-dataset and sim-to-real generalization.
AdaNav~\cite{ding2025adanav} introduced an uncertainty-aware adaptive reasoning framework that dynamically triggers explicit reasoning, enabling difficulty-aware perception–action alignment with reduced computational overhead.
Aux-Think~\cite{wang2025think} systematically investigated reasoning strategies in VLN and found that inference-time chain-of-thought reasoning can degrade performance; instead, it uses reasoning traces as auxiliary supervision during training while preserving direct action prediction at test time, offering insights into effective reasoning integration.
CorrectNav~\cite{yu2025correctnav} emphasized error correction through a self-correction flywheel that iteratively mines erroneous trajectories and generates corrective supervision, improving recovery from deviations and long-instruction following.

Beyond temporal reasoning, effective long-horizon navigation also hinges on robust spatial understanding, which has recently attracted significant attention in LLM-based VLN approaches.
StreamVLN~\cite{wei2025streamvln} proposed a slow-fast context modeling strategy that decouples responsive action generation from long-term visual memory compression, enabling efficient navigation over long video streams with bounded inference cost.
NaVid-4D~\cite{navid4d} employed a 3D-aware vision encoder to process egocentric RGB-D video, enabling explicit spatio-temporal reasoning for precise instruction-following actions.
JanusVLN~\cite{zeng2025janusvln} introduced a dual implicit memory paradigm that encodes visual semantics and spatial geometry as compact neural representations, avoiding explicit structure reconstruction while enhancing spatial reasoning from RGB input alone.
MapNav~\cite{mapnav} introduced an annotated semantic map-based spatial representation, maintaining a top-down, continuously updated map that provides structured spatial context.
GC-VLN~\cite{yin2025gc} presented a training-free framework that interprets instructions as graph constraints over a map-level representation, enabling zero-shot navigation in continuous environments without learned policies.
VLN-Zero~\cite{bhatt2025vln} constructed symbolic scene graphs through rapid exploration and performed navigation via graph-based reasoning with cache-enabled execution within a neurosymbolic planning framework.

As K{\aa}sene \textit{et al.}~\cite{kaasene2025following} revealed, monolithic LLM policies that directly map observations to low-level actions struggled to complete VLN tasks, compared to approaches using hierarchical waypoint-based spaces.
Even in a relatively simple setting in NaviTrace~\cite{windecker2025navitrace}, where VLMs generate navigation traces on static images under specific constraints, performance remains far below human levels (\textit{i.e.,} 34 Score \textit{v.s.} 75 Score).
This limitation explains why most competitive LLM-based VLN systems require substantial domain-specific fine-tuning.
Although several zero-shot methods have been proposed, such as Fast-SmartWay~\cite{shi2025fastsmartWay} and CL-CoTNav~\cite{cai2025cl}, their performance remained modest in comparison.
Therefore, bridging the gap between zero-shot generalization and effective domain adaptation, especially for robust real-world deployment, remains a critical open challenge.

\subsubsection{Unified Foundation Models}
Besides approaches that focus solely on VLN tasks, another research direction explores unified generative models that are suited for different tasks, embodiments, and environments.
Uni-NaVid~\cite{zhang2024uni} introduced a vision-language-action model that unifies multiple embodied navigation tasks, including VLN, object navigation, embodied question answering, and human following, within a single framework.
By harmonizing input-output formats and leveraging large-scale multi-task data, Uni-NaVid demonstrates that task unification can improve both performance and real-world efficiency.
NaVILA~\cite{cheng2024navila} adopted a generative framework that bridges high-level language understanding and low-level robot control for legged navigation.
It generates spatially grounded language actions (\textit{e.g.}, \textit{``move forward 75 cm''}) that can be executed by a vision-based locomotion policy, enabling unified deployment across different embodiments and demonstrating generalization to related embodied tasks such as ScanQA.
AstraNav-World~\cite{hu2025astranav} proposed a novel architecture that integrates two key components: a spatio-temporal variational autoencoder and a diffusion transformer network. The diffusion policy head adopted to generate a sequence of waypoint directly without the need of an independent waypoint generator model.
OmniVLA~\cite{hirose2025omnivla} introduced an omni-modal vision-language-action framework for robot navigation that enables a single policy to handle diverse goal modalities, including language instructions, 2D poses, and egocentric images, while demonstrating robust generalization to unseen environments and flexible adaptation to new tasks.
Abot-N0~\cite{chu2026abot} proposed a unified vision-language-action foundation model consisting of a universal multimodal encoder, a cognitive brain, and an action expert that uses flow matching to generate trajectories, demonstrating strong cross-task and cross-embodiment generalization across navigation tasks.
NavFoM~\cite{zhang2025embodied} and Qwen-RobotNav~\cite{zhang2026qwen} improve cross-task navigation performance through unified large-model-based architectures.

Overall, the generative paradigm marks a fundamental shift in VLN from task-specific pipelines toward unified, data-driven, and autoregressive decision-making.
While these approaches demonstrate strong generalization and flexibility, they also raise new challenges related to training cost, inference efficiency, and effective long-horizon memory modeling.
Addressing these challenges is likely to be a key focus of future research on LLM-based embodied navigation.

\subsubsection{Diverse Environments and Embodiments}

Recent advances in LVLMs and LLMs have enabled embodied agents to tackle navigation tasks across a wide range of environments and robotic embodiments.
Outdoor VLN has benefited from generative approaches that leverage diverse visual data and high-level semantic reasoning.
VLN-Video~\cite{li2024vlnvideo} used driving videos from multiple cities to pretrain agents with automatically generated instruction-action pairs, improving generalization in outdoor VLN.
Similarly, NavAgent~\cite{liu2024navagent} employed a large VLM to fuse multi-scale urban scene representations, including topological maps, panoramas, and fine-grained landmarks, enabling UAV agents to navigate complex city environments through semantic reasoning.
VLM-GroNav~\cite{elnoor2024robot} incorporated proprioceptive sensing to update terrain traversability in real time, enhancing navigation performance on deformable and slippery outdoor surfaces.
In parallel, SAGE-3D~\cite{miao2025towards} extended 3D Gaussian Splatting environments with object-centric semantics and physics-aware execution, providing a realistic and physically executable framework for LLM-driven agents.
NavComposer~\cite{he2025navcomposercomposinglanguageinstructions} proposed a modular framework that enables large-scale generation of navigation instructions by decomposing trajectories into action, scene, and object modules.
Generative approaches also improve generalization by modeling diverse trajectory patterns.
Baghaei \textit{et al.}~\cite{baghaei-etal-2025-follow} demonstrated that action sequences and trajectory shapes constitute an overlooked source of information that LLM-based agents can leverage to better generalize to unseen routes.
This insight motivates data augmentation strategies that expose agents to richer patterns of navigation behavior.

Aerial VLN presents unique challenges due to complex spatial relationships and large-scale environments, making it a suitable testbed for methods integrating semantic reasoning and vision-language models.
UAV-ON~\cite{uavon} introduced a benchmark for open-world object-goal navigation, where aerial agents must reason over high-level semantic goals in diverse urban and natural environments, highlighting the difficulty of grounding instructions without step-by-step guidance.
Methods such as STMR~\cite{gao2024aerial} and Xu \textit{et al.}~\cite{xu2025aerial} leveraged LLMs to perform action prediction over semantic-topo-metric maps or via prompt-guided trajectory reasoning, enabling long-horizon, goal-directed flight in unseen environments.
To improve training and deployment, curriculum learning and safety-aware strategies have been proposed: SA-GCS~\cite{cai2025sa} systematically adjusted the difficulty of training samples to accelerate learning and generalization, while OpenFly~\cite{gao2025openfly} provided a comprehensive aerial VLN platform that integrated multi-engine simulation, large-scale data collection, and a keyframe-aware agent to enhance real-time adaptability.

In specialized domains such as agriculture and underwater robotics, experimental generative VLN approaches have also emerged in constrained environments.
AgriVLN~\cite{zhao2025agrivln} focused on agricultural field navigation using quadruped robots and adopted VLM-based instruction prompting to ground high-level language commands in crop-level visual observations.
Building on this setting, T-araVLN~\cite{zhao2025t} enhanced instruction decomposition to better align complex agricultural tasks with sequential navigation actions, while SUM-AgriVLN~\cite{zhao2025sum} incorporated spatial memory mechanisms to address long-horizon navigation and revisitation in large-scale farmland.
Furthermore, MDE-AgriVLN~\cite{zhao2025mde} integrated monocular depth estimation to compensate for limited sensing and perceptual ambiguity in agricultural scenes, improving spatial reasoning under constrained visual input.
UnderwaterVLA~\cite{wang2025underwatervla} combined dual-brain LLM reasoning with vision-language-action models and hydrodynamics-informed control to enable robust navigation for autonomous underwater vehicles, highlighting the adaptability of generative VLN beyond terrestrial and aerial domains.

Collectively, these studies reflect ongoing efforts to apply generative navigation methods across diverse environments, while accounting for varying spatial scales, temporal horizons, and robotic embodiment constraints.

\subsection{Strengths and Limitations}
\subsubsection{Strengths}
The monolithic framework establishes a direct mapping from visual-language inputs to navigation actions, requiring minimal reliance on auxiliary modules.
As a result, one of its most prominent strengths lies in its conceptual clarity and system simplicity.
By avoiding multi-stage pipelines, monolithic approaches reduce the risk of error propagation across components, thereby simplifying both implementation and deployment.
From a modeling perspective, the monolithic formulation aligns naturally with the design of modern generative models.
The sequential image input closely resembles video data, enabling LVLMs with video understanding capabilities to be applied to end-to-end VLN with minimal architectural modification.
Moreover, training data in this format can be readily sourced from large-scale internet video corpora, significantly expanding the diversity of visual-language experiences available to the model.
In addition, the input-output interface of monolithic VLN is broadly compatible with a wide range of robotic platforms: monocular RGB observations can be obtained using low-cost cameras, while the generated navigation actions require only lightweight control policies for execution.
These properties make monolithic approaches attractive for real-world deployment and rapid prototyping across diverse robotic embodiments.

\subsubsection{Limitations}
Despite these advantages, monolithic frameworks also exhibit notable limitations.
Relying primarily on monocular perception and implicit spatial memory, such models may fail to accurately perceive nearby obstacles, increasing the risk of collisions.
The absence of explicit global localization and dense motion planning further limits the precision and smoothness of robot locomotion compared to map-based or hierarchical approaches.
End-to-end monolithic models often operate as black boxes: their decision-making processes are difficult to interpret, analyze, or debug, resulting in weaker explainability than modular systems.
Additionally, recent high-performing generative methods typically depend on extensive fine-tuning of foundation models.
The substantial computational cost and data requirements associated with such training may hinder scalability and slow broader adoption.

While monolithic VLN frameworks offer an elegant and unified formulation, balancing their simplicity and generality with safety, interpretability, and training efficiency remains an open challenge for future research.

\begin{table*}[t]
    \centering
    \caption{Comparison of recent monolithic foundation-model-based VLN methods.}
    \label{tab:foundation_vln_methods}
    \resizebox{\textwidth}{!}{
    \begin{tabular}{p{1.7cm} p{1.8cm} p{1.8cm} p{3.0cm} p{3.2cm} p{3.0cm} p{3.0cm}}
    \toprule
    \textbf{Method} & \textbf{Visual Input} & \textbf{Output} & \textbf{Foundation Model} & \textbf{Training Data} & \textbf{Evaluation} & \textbf{GPU Resources} \\
    \midrule
    
    NaVid~\cite{zhang2024navid} &
    Mono. RGB video &
    Low-level textual action &
    EVA-CLIP  + Vicuna-7B &
    10,819 R2R-CE train split &
    R2R-CE, RxR-CE &
    24 A100, 28h; $\sim$672 GPU hours \\ \hline
    
    Uni-NaVid~\cite{zhang2024uni} &
    Mono. RGB video &
    4-step action chunk &
    EVA-CLIP + Vicuna-7B &
    3.6M navigation samples + 2.3M VQA/captioning samples &
    VLN-CE, RxR-CE, OGNav, EQA, human following &
    40 H800, 35h; $\sim$1400 GPU hours \\ \hline
    
    NaVILA~\cite{cheng2024navila} &
    Mono. RGB video &
    Mid-level textual action &
    VILA-based VLM + visual locomotion policy &
    2K YouTube tour videos + R2R-CE + RxR-CE + EnvDrop + ScanQA + VQA &
    VLN-CE, RxR-CE, ScanQA, VLN-CE-Isaac &
    PT: 16 A100, 34h ($\sim$544 GPU hours); FT: 4 A100, 18h ($\sim$72 GPU hours) \\ \hline
    
    StreamVLN~\cite{wei2025streamvln} &
    Mono. RGB video &
    4-step action chunk &
    LLaVA-Video-7B &
    450K nav clips + 300K ScaleVLN + 240K DAgger + 478K general VL data &
    R2R-CE, RxR-CE, ScanQA &
    $\sim$1500 A100 GPU hours \\ \hline
    
    JanusVLN~\cite{zeng2025janusvln} &
    Mono. RGB video &
    Discrete action &
    Qwen2.5-VL-7B + VGGT &
    R2R-CE + RxR-CE + 155K ScaleVLN + 14K DAgger + General VQA &
    R2R-CE, RxR-CE &
    Not reported\\ \hline
    
    CorrectNav~\cite{yu2025correctnav} &
    Mono. RGB video &
    4-step action chunk &
    LLaVA-Video-7B &
    R2R-CE + RxR-CE + 30K generated instructions + 178K LLaVA-Video + 240K VQA &
    R2R-CE, RxR-CE &
    8 A100; 80h fine-tuning + 20h per self-correction iteration \\ \hline
    
    NavFoM~\cite{zhang2025embodied} &
    multi-view RGB video. &
    Trajectory / text &
    DINOv2 + SigLIP + Qwen2-7B &
    8.02M navigation samples + 3.15M image-QA + 1.61M video-QA &
    R2R-CE, RxR-CE, Open-UAV, HM3D-OVON, EVT-Bench, nuScenes, NAVSIM &
    56 H100, 72h; $\sim$4,032 GPU hours \\ \hline
    
    Aux-Think~\cite{wang2025think} &
    Mono. RGB video &
    Low-level textual action &
    NVILA-lite 8B &
    R2R-CoT-320K + R2R-CE + 600K RxR + 500K DAgger + 500K web data &
    R2R-CE, RxR-CE &
    8 H20, 60h; $\sim$480 GPU hours \\ \hline

    Qwen-RobotNav~\cite{zhang2026qwen} &
    Mono. RGB / Pano &
    Trajectory / text &
    Qwen3-VL-4B/8B &
    R2R-CE + RxR-CE + 1.8M PointNav + 2.0M ObjNav + 1.5M Track + 3.2M Autonomous-driving &
    R2R-CE, RxR-CE, R2R-PE &
    2,816 H100 GPU hours \\
    
    \bottomrule
    \end{tabular}
    }
    \end{table*}

\section{Real-world Evaluation}
\label{sec_real_world}

Recent advances in VLN have increasingly focused on deployment and application on physical robotic platforms. Building upon a systematic theoretical taxonomy and review, this work further provides empirical evaluations and analyses aimed at offering valuable insights for future research.

\subsection{Experimental Design}

\subsubsection{Hardware Platform}
\begin{figure}[b]
    \centering
    \includegraphics[width=0.9\linewidth]{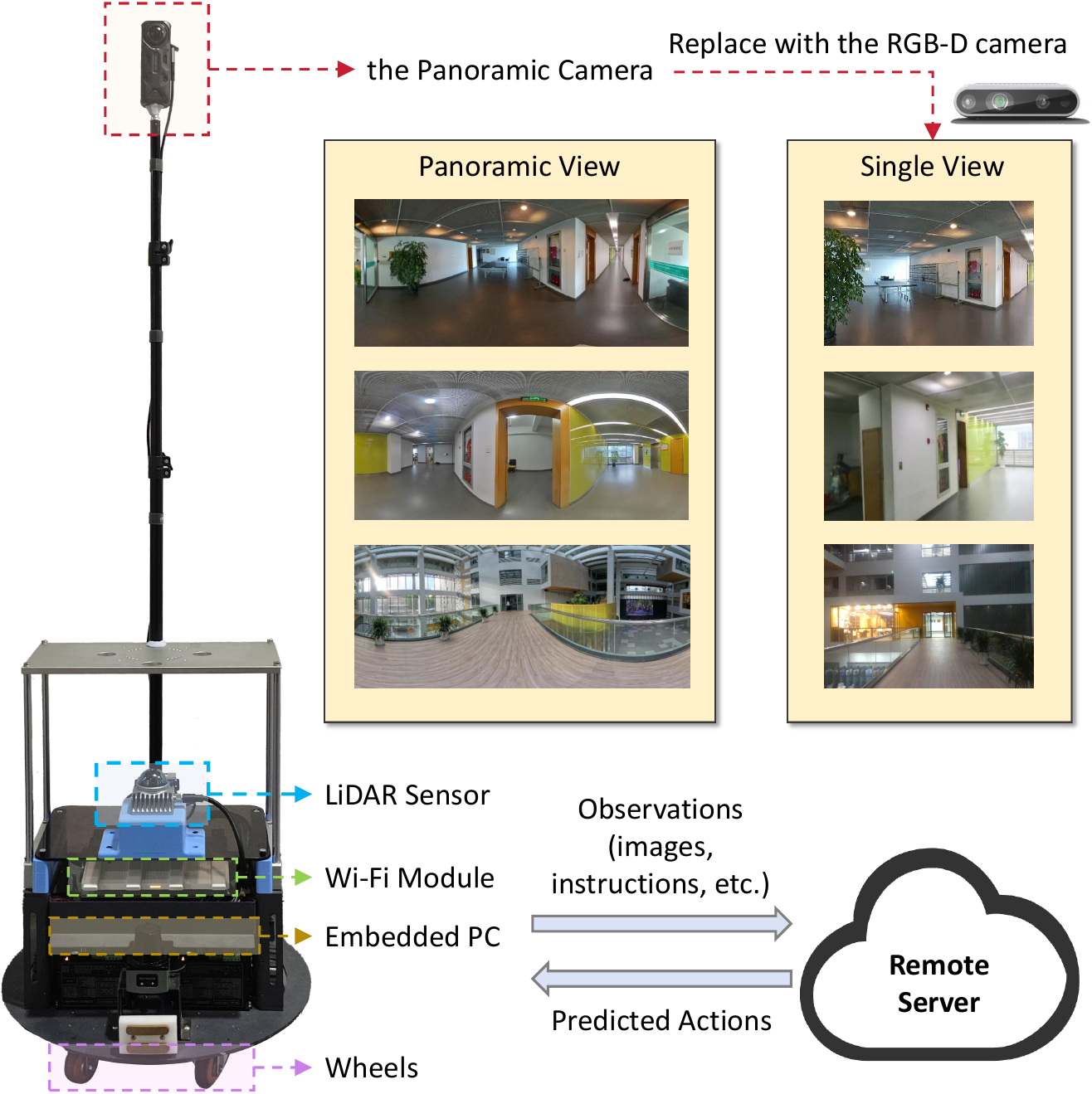}
    \caption{The used wheeled robot for the real-world evaluation.}
    \label{fig_robot_display}
\end{figure}
As illustrated in Fig. \ref{fig_robot_display}, we built a wheeled mobile robot platform for VLN real-world deployment. The platform is equipped with a differential-drive chassis enabling translational and rotational motion, an embedded computing unit (NVIDIA Jetson Orin Nano Developer Kit), and a 3D LiDAR (Mid-360) installed for geometric perception.
To acquire visual observations, a camera is mounted on a telescopic tripod at a default height of 1.5 m. Depending on the experimental configuration, the visual sensor can be either a 360° panoramic camera (Insta360 X4) or an RGB-D camera (RealSense D435i).
The onboard processor handles SLAM, low-level motion control, and local image preprocessing, while high-level VLN inference is offloaded to a remote server via Wi-Fi. Upon receiving a navigation task, the server initializes the model state and predicts action sequences according to the chosen navigation paradigm. The action space is defined either in a waypoint-based formulation or as a set of step-wise discrete actions (\textit{e.g.}, turning left/right by 15$\degree$, moving forward by 25 cm, or stop).
The inferred actions are transmitted back to the onboard processor for execution, enabling real-world interaction and continuous sensor feedback, which is streamed back to the remote server to form a closed-loop system. For waypoint-based navigation, SLAM is used to provide online localization and mapping for point-goal navigation. For step-wise actions, the agent directly executes predefined commands using a PID feedback controller to ensure accurate and stable low-level control. All experiments are conducted using a remote server equipped with an NVIDIA L40 GPU.

\subsubsection{Selection of VLN Baselines}

As discussed earlier, existing VLN approaches can be categorized into two paradigms: \emph{Hierarchical Frameworks} and \emph{Monolithic Frameworks}. For the hierarchical paradigm, we select \texttt{CLASH}~\cite{wang2025clash} as the representative baseline due to its SoTA VLN-CE performance, with a 66\% SR on the test-unseen split, and its practical waypoint generation strategy for real-world navigation. Panoramic depth is not used as input because pixel-aligned depth sensing is unavailable in the real-world evaluation.
For the monolithic paradigm, we adopt \texttt{JanusVLN}~\cite{zeng2025janusvln} as the representative baseline with a SR of 61\%, achiving advanced performance compared to other monolithic approaches. Specifically, the hierarchical waypoint-based method uses panoramic images captured by an Insta360 X4 as visual input. Candidate waypoints are generated automatically during inference~\cite{wang2025clash}. The monolithic action-based method relies on single-view RGB images captured by a RealSense D435i. For fair comparison, all episodes share the same starting positions and language instructions.

\subsubsection{Evaluation Scenes and Instructions}

\begin{figure*}[t]
    \centering
    \includegraphics[width=0.9\linewidth]{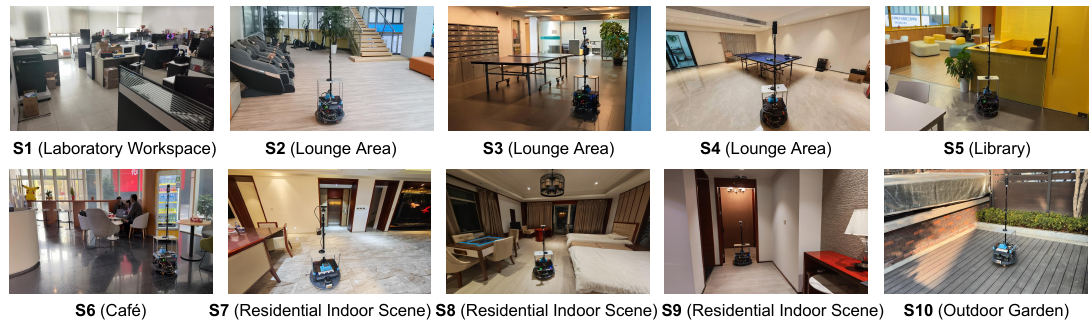}
    \caption{Real-world scenes for evaluation of vision-and-language navigation, including 10 different environments.}
    \label{fig_real_world_scenes}
\end{figure*}

\begin{figure}[htbp]
    \centering
    \includegraphics[width=0.8\linewidth]{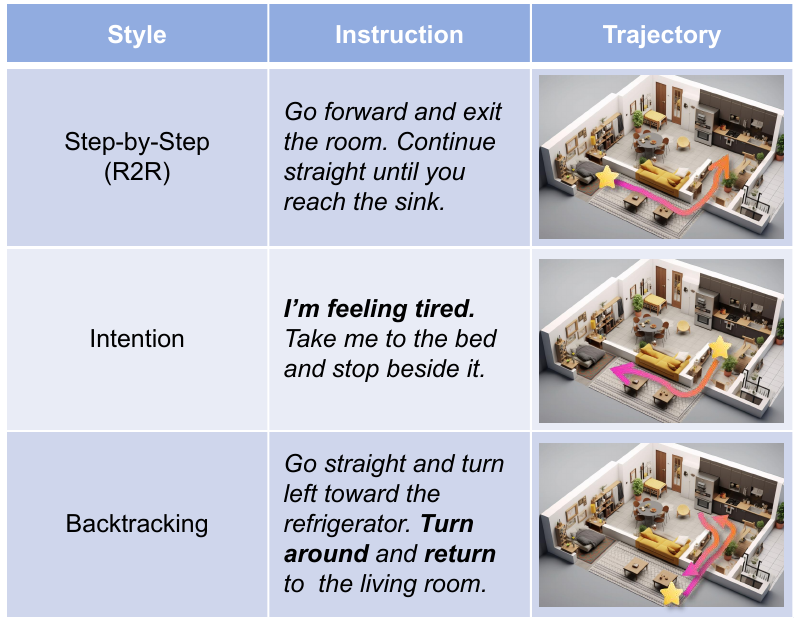}
    \caption{Examples of instuctions and trajectories for real-world evaluation.}
    \label{fig_instruction_style}
\end{figure}

\begin{figure}[htbp]
    \centering
    \includegraphics[width=0.8\linewidth]{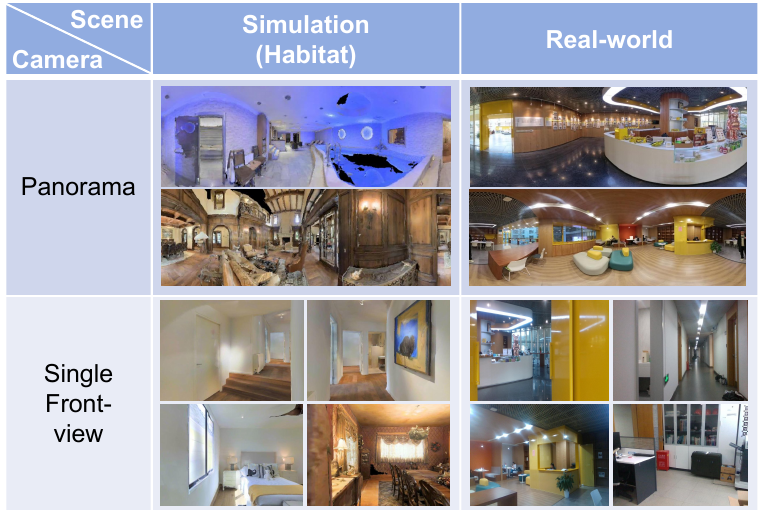}
    \caption{Comparison of the first-person visual observations between simulation (Habitat) and real-world.}
    \label{fig_compare_habitat_realworld}
\end{figure}

To comprehensively evaluate the performance of current VLN methods in real-world settings, we conduct experiments across 10 diverse environments, as illustrated in Fig.~\ref{fig_real_world_scenes}. These environments include laboratory spaces, lounge areas, libraries, cafés, residential indoor scenes, and an outdoor garden. \newcontent{Fig.~\ref{fig_compare_habitat_realworld} presents a qualitative comparison of first-person visual observations between simulation (Habitat) and real-world environments. As illustrated, real-world scenes exhibit more pronounced motion blur, lens flare, and greater scene diversity, whereas simulated environments tend to be visually cleaner and primarily focused on indoor residential settings. These perceptual discrepancies contribute to the sim-to-real gap observed in practice.}
We adopt a strict unseen evaluation assumption: the robot has no prior knowledge of the scene before each task. For each environment, 10 navigation tasks are designed based on its spatial layout and functional characteristics. \newcontent{As shown in Fig.~\ref{fig_instruction_style}, 70\% of the instructions follow the standard R2R-style step-by-step format, while the remaining 30\% consist of more challenging instruction types, including intent-ambiguous commands (20\%) and instructions involving multi-stage backtracking (10\%). In total, the evaluation set comprises 200 navigation episodes, with 100 episodes for each baseline. During evaluation, each episode was tested once with a fixed random seed of 42.}
%=====
% Examples of instuctions and trajectories for real-world evaluation.
%=====
% =====
% Compare habitat with real-world
% =====

\subsubsection{Additional Evaluation Metrics}
Commonly used metrics such as SR rely on coarse success criteria, deeming an episode successful if the agent stops within a fixed distance (\textit{e.g.}, 3 m) of the target. 
However, this coarse criterion may misclassify cases where the robot stops near the target but fails to reach the semantically correct location specified by the instruction. Therefore, we introduce a stricter metric, the \emph{Strict Success Rate (SSR)}, which counts an episode as successful only when two conditions are simultaneously satisfied: the robot stops within 3 m of the target, and its final location semantically satisfies the instruction. \newcontent{Formally, SSR is defined as}
\begin{equation}
\newcontent{\mathrm{SSR} = \frac{1}{N}\sum_{i=1}^{N} 
\mathbb{I}\left[d(\mathbf{p}_i, \mathbf{g}_i) \leq d_{th} \right]
\cdot
\mathbb{I}\left[sem_i = 1\right],}
\end{equation}
\newcontent{where $d(\mathbf{p}_i, \mathbf{g}_i)$ is the Euclidean distance between the final agent position $\mathbf{p}_i$ and the goal position $\mathbf{g}_i$, and $d_{th}$ is the success threshold.
$sem_i \in \{0,1\}$ denotes the semantic correctness annotation for the final stopping location. For example, under the instruction \textit{``stop once you are out the door,''} stopping before exiting the door yields $sem_i=0$ and thus an SSR failure, even if the final position is within 3 m of the target and is counted as successful by SR.}

In addition, obstacle avoidance is evaluated using the \emph{Collision Rate (CR)}, which measures the proportion of episodes in which the robot collides with environmental obstacles. \newcontent{It is defined as}
\begin{equation}
\newcontent{\mathrm{CR} = \frac{1}{N}\sum_{i=1}^{N} \mathbb{I}\left[col_i = 1\right],}
\end{equation}
\newcontent{where $col_i \in \{0,1\}$ indicates whether a collision occurs in the $i$-th episode. During evaluation, an episode is terminated immediately once the robot encounters a collision.}

\subsection{Quantitative Analysis}

Through extensive real-world embodied experiments, this study systematically investigates the following questions:

(1) How do representative VLN system configurations differ in navigation performance across diverse real-world environments?

(2) How do these systems compare in their obstacle avoidance capabilities under identical evaluation conditions?

(3) How does navigation performance vary when agents are evaluated with instruction types beyond classical step-by-step commands, including ambiguous intent and backtracking?

Based on this experimental protocol, we identify key limitations of existing approaches and derive actionable insights to guide future research in real-world VLN.

\subsubsection{Average Performance Comparison}

\begin{figure}[t]
    \centering
    \includegraphics[width=0.9\linewidth]{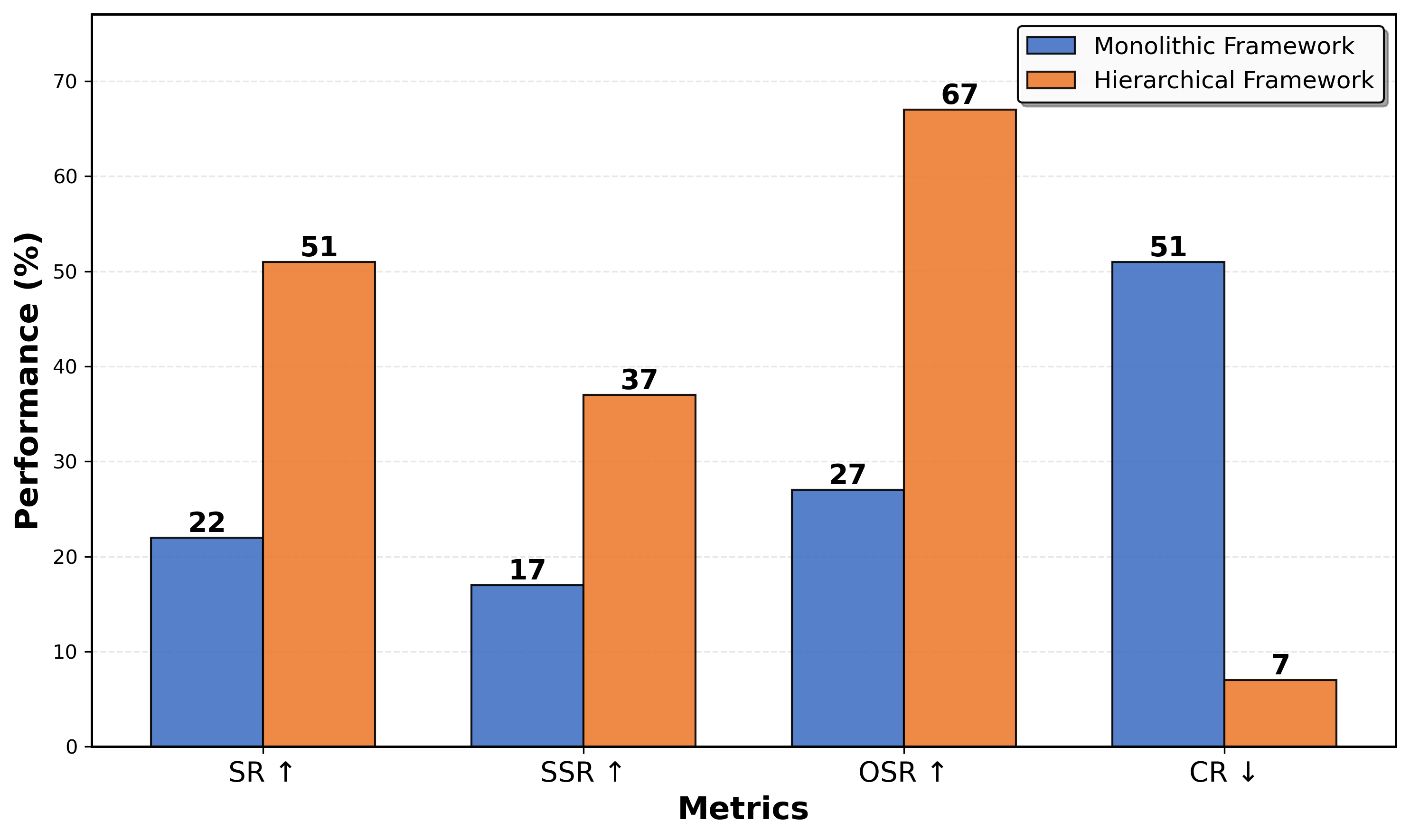}
    \caption{\newcontent{System-level comparison of the average performance of hierarchical and monolithic VLN frameworks in real-world evaluations under the tested configurations.}}
    \label{fig_real_world_exp_comparison}
\end{figure}

Fig.~\ref{fig_real_world_exp_comparison} summarizes the average performance of representative hierarchical and monolithic VLN system configurations over 200 real-world navigation episodes conducted across 10 distinct scenes. Overall, the monolithic configuration achieves an SR of 22\%, an SSR of 17\%, and an OSR of 27\%, whereas the hierarchical configuration attains higher performance, with an SR of 51\%, an SSR of 37\%, and an OSR of 67\%. \newcontent{These results provide empirical evidence that, under our tested configurations, the hierarchical system transfers more effectively to diverse real-world environments than the monolithic RGB-only system.}

Beyond navigation accuracy, safety emerges as a pronounced limitation in the tested monolithic configuration. In real-world deployment, the monolithic VLN system exhibits a high CR of 51\%, reflecting brittle obstacle avoidance behavior. In contrast, the SLAM-based hierarchical system demonstrates a lower CR of 7\%, \newcontent{suggesting that explicit spatial modeling and geometry-aware planning can play an important role in reliable real-world navigation.}

\newcontent{The observed performance gap is likely influenced by multiple factors beyond the high-level architectural paradigm.} 
\textit{First}, the use of panoramic visual sensing enables the hierarchical system to capture richer and more complete environmental observations, mitigating perceptual limitations caused by restricted fields of view. 
\textit{Second}, the topological graph memory used in the hierarchical system explicitly supports long-range memory retrieval, allowing the agent to detect navigation errors, revise its plan, and revisit previously explored locations. This global-grounding capability is largely absent in the tested monolithic configuration.
\textit{Third}, point-goal navigation supported by LiDAR and SLAM substantially reduces collision risk, whereas the tested monolithic RGB-only configuration lacks access to explicit geometric sensing and struggles to operate safely and reliably under physical constraints.
\newcontent{Therefore, while our results reveal a performance gap between the tested hierarchical and monolithic configurations, they should not be interpreted as fully isolating the effect of architecture from sensing, mapping, and control-stack differences.}

A closer examination reveals that when strict semantic stopping conditions are enforced, the success rate of the hierarchical configuration drops from 51\% (SR) to 37\% (SSR). This phenomenon has rarely been discussed in prior simulator-based studies~\cite{anderson2018vision,puig2023habitat3,wang2025rethinking}, primarily because it is difficult to precisely define semantic correctness of stopping behavior in simulation. In real-world settings, however, humans can reliably judge whether an instruction has been followed exactly as specified. 
This semantic misalignment in stopping behavior is expected to become increasingly problematic when VLN is integrated with embodied manipulation tasks~\cite{ahn2022saycan,brohan2023rt2}, where accurate spatial grounding directly affects task feasibility. Consequently, precise semantic termination remains a critical challenge for future VLN research.

\subsubsection{Per-Scene Performance Comparison}

\begin{figure}[t]
    \centering
    \includegraphics[width=\linewidth]{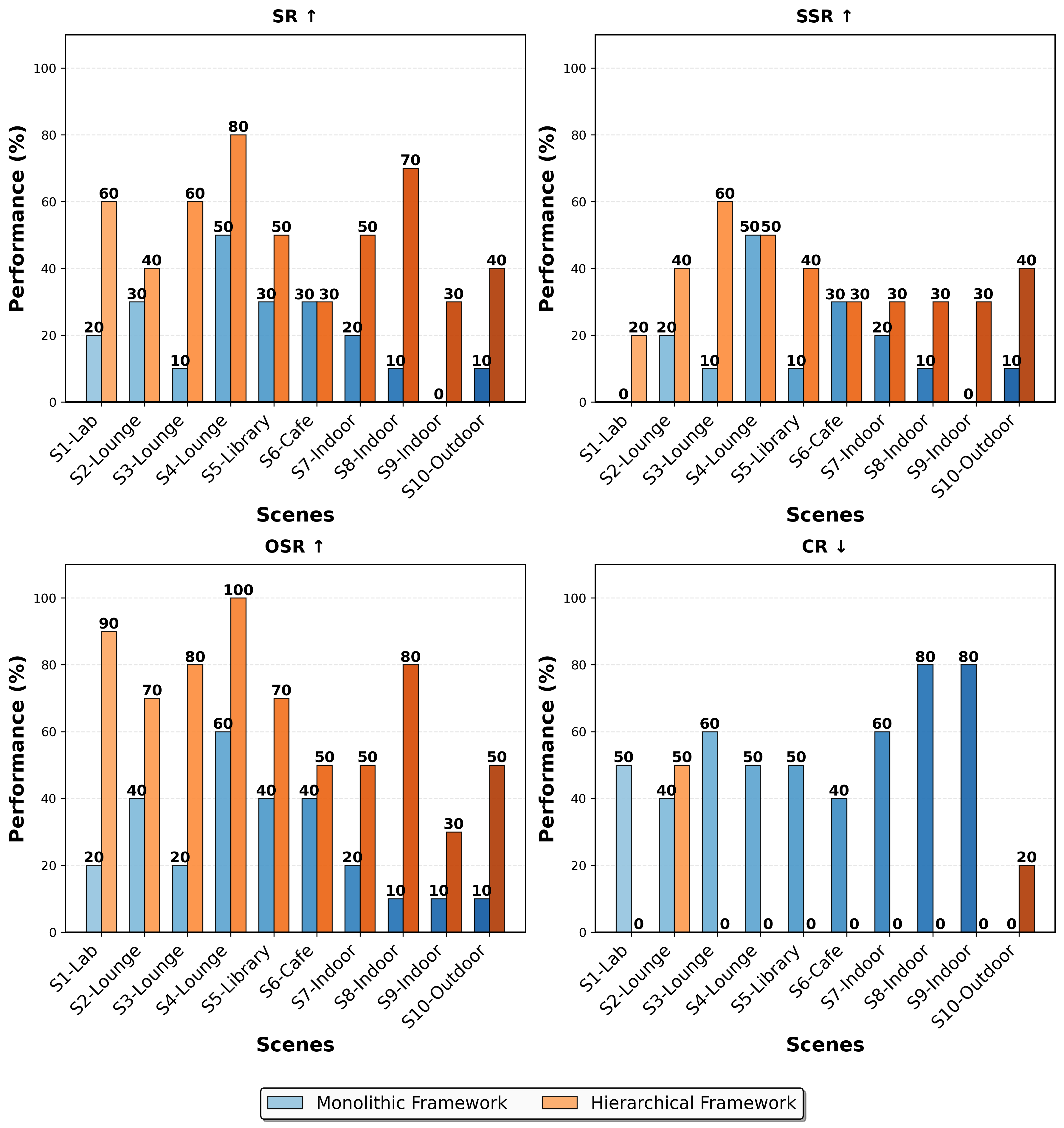}
    \caption{Comparison of the performance of the hierarchical and monolithic frameworks in the real-world evaluation on every scene.}
    \label{fig_real_world_every_scene_exp}
\end{figure}

Fig.~\ref{fig_real_world_every_scene_exp} further presents per-scene performance comparisons between hierarchical and monolithic frameworks. Across most scenes, the hierarchical method achieves higher SR and lower CR. Notably, collisions are entirely avoided in 8 scenes, while the remaining 2 scenes involve collisions caused by stair structures lower than the LiDAR scanning plane.

Additionally, although the evaluation environments cover a wide range of real-world settings, we do not observe a pronounced navigation performance advantage in residential indoor scenes, even though such environments dominate standard VLN simulators. We hypothesize that this is primarily due to the limited generalization capability of current VLN models, which have yet to demonstrate robust adaptation even to environment categories that dominate existing training data. This observation further underscores the gap between simulation-based benchmarks and real-world VLN deployment.

\subsubsection{Intention and Backtracking Performance Comparison}

\begin{figure}[t]
    \centering
    \includegraphics[width=\linewidth]{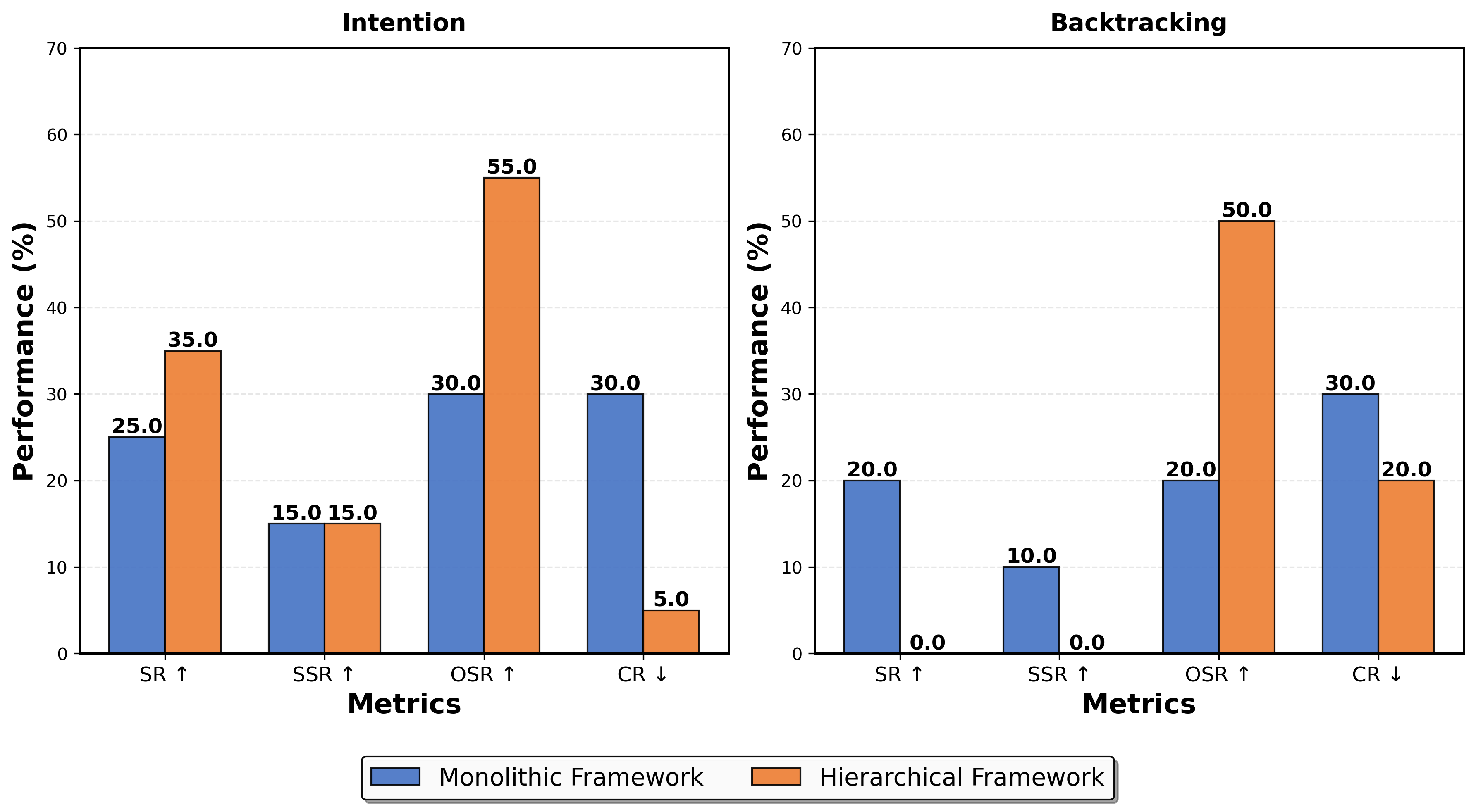}
    \caption{Comparison of the performance of the hierarchical and monolithic frameworks in the real-world evaluation on intention and backtracking tasks.}
    \label{fig_intent_backtracking_exp}
\end{figure}

Fig.~\ref{fig_intent_backtracking_exp} presents a comparative evaluation of hierarchical and monolithic VLN frameworks on tasks involving intention inference and backtracking. Compared with classical step-by-step instructions, these instruction types are more representative of real-world human commands. Specifically, we evaluate 20 intention instructions and 10 backtracking instructions across 10 different scenes.
Specifically, intention instructions require the agent to exhibit spatial reasoning and exploratory behavior in order to infer underspecified goals, whereas backtracking instructions demand long-horizon memory and the ability to retrace previously visited paths. Experimental results show that the hierarchical framework attains marginally higher performance on intention tasks (SR 35\% \textit{vs.} 25\%), but performs worse on backtracking tasks (SR 0\% \textit{vs.} 20\%).

A key factor underlying this observation is the absence of explicit stopping cues in many intention and backtracking instructions. Without clear termination signals, both types of models struggle to determine when to issue a \emph{stop} action based solely on instruction semantics. In practice, agents often pass through the ground-truth target region but continue navigating until the maximum step limit is reached, resulting in task failure despite semantically reasonable trajectories.

These findings highlight a fundamental limitation of current VLN systems in handling more naturalistic and challenging instruction formulations. Improving semantic stopping decisions, long-term intention tracking, and memory-aware control for such realistic navigation tasks remains an open and important direction for future VLN research.

\subsubsection{Comparison of Computation Time and Inference Step Length}

\begin{figure}[htbp]
    \centering
    \includegraphics[width=\linewidth]{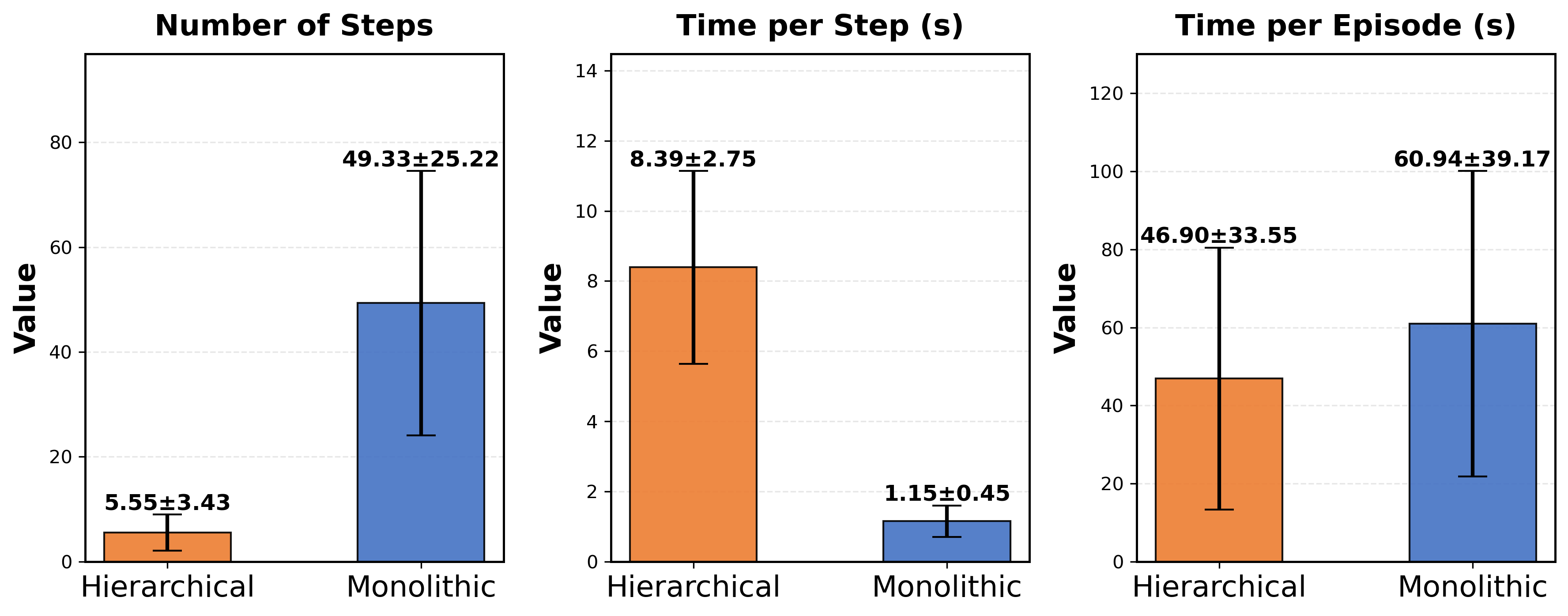}
    \caption{Comparison of the model inference time and the average step length of the hierarchical and monolithic frameworks in the real-world evaluation.}
    \label{fig_time_and_step_exp}
\end{figure}

Fig.~\ref{fig_time_and_step_exp} compares the average model-side inference time and inference steps of hierarchical and monolithic VLN frameworks in real-world deployments, excluding low-level actuation and cloud-edge communication overhead.
It shows that the hierarchical framework requires significantly fewer inference steps than the monolithic framework. However, each inference step in the hierarchical framework incurs a longer computation time, primarily due to the invocation of large-scale language models (Qwen2.5-VL-72B~\cite{Qwen2.5-VL} via a remote API). Despite the higher per-step latency, the overall inference time of the hierarchical framework remains lower, as the reduction in the number of inference steps outweighs the increased cost per step.

During real-world experiments, we further observe that network latency can substantially affect the execution efficiency of the robot, particularly when inference is offloaded to remote servers. Given that future robotic systems are expected to operate with greater autonomy and reduced reliance on cloud infrastructure, these observations highlight the importance of model lightweighting and real-time deployment on edge devices. Addressing this challenge will likely require advances not only in system engineering but also in algorithmic design, making it a critical direction for future VLN research.

\subsection{Qualitative Analysis}
\subsubsection{Real-world Evaluation Examples}
\label{sec:real_world_evaluation_examples}

\begin{figure*}[htbp]
    \centering
    \includegraphics[width=0.9\linewidth]{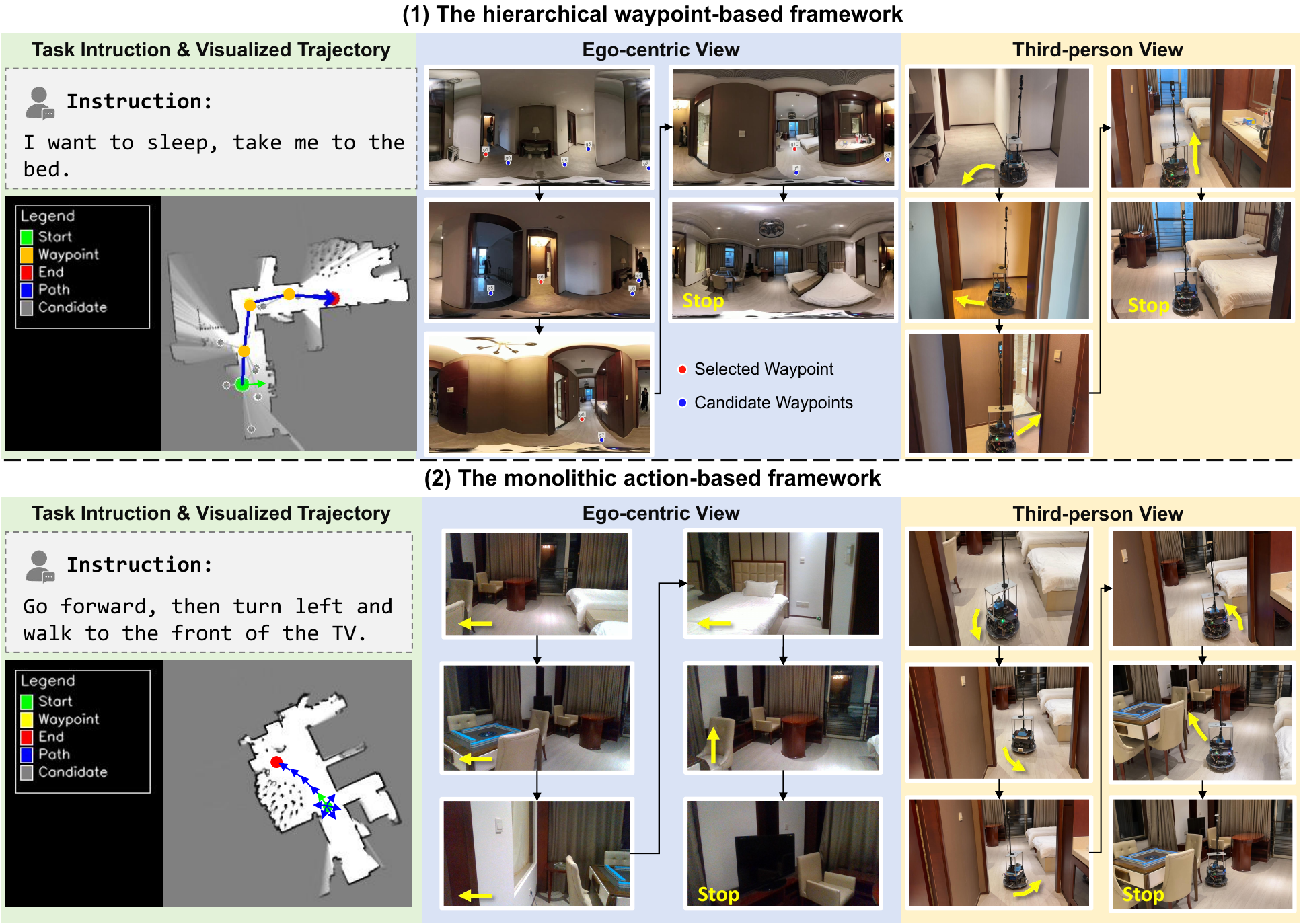}
    \caption{Examples of real-world navigation trajectories produced by hierarchical and monolithic VLN frameworks \newcontent{under the tested configurations.}}
    \label{fig_real_world_success_cases}
\end{figure*}

Fig.~\ref{fig_real_world_success_cases} presents representative navigation examples of both hierarchical and monolithic VLN frameworks in real-world environments. For visualization purposes, we leverage the mapping and localization capabilities of the SLAM module to record the robot's positions and orientations. 
During the experiments, we observe that the monolithic framework tends to perform an initial 360-degree in-place rotation before proceeding with instruction-following navigation, whereas this behavior is absent in the hierarchical framework. This difference arises because the hierarchical framework can directly access panoramic visual observations, eliminating the need for explicit rotational exploration to acquire a complete environmental view. From a principled perspective, panoramic sensing substantially increases the amount of environmental information available at each decision step, while LiDAR measurements provide accurate geometric structure of the surroundings. The complementary use of these sensing modalities contributes to the higher navigation success rates and safer execution observed for the hierarchical framework in real-world deployment.

Looking forward, the choice of sensing modalities and the effective fusion of heterogeneous sensory inputs remain important and open research questions. Systematic investigation into sensor selection and multimodal fusion strategies is likely to play a critical role in improving the robustness and generalizability of future VLN systems.

\subsubsection{Real-world Failure Cases}
\label{sec:failure_analysis}

In real-world deployments, VLN failures mainly stem from two factors. The first is incorrect action prediction, which causes the agent to navigate in an erroneous direction. 
This issue largely stems from limited model generalization and robustness, as idealized simulation data cannot adequately capture the ambiguity of real-world observations, the diversity of environmental appearances, or the variability of natural language instructions.

\begin{figure}[t]
    \centering
    \includegraphics[width=0.7\linewidth]{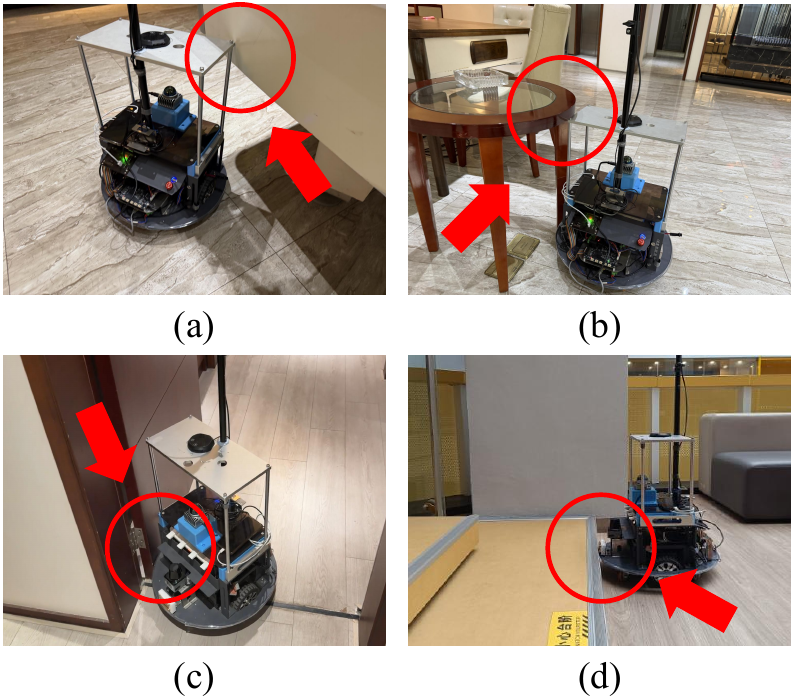}
    \caption{Examples of collision failure cases in the real-world evaluation.}
    \label{fig_real_world_failure_cases}
\end{figure}

The second, and often more critical, factor is collision, where the robot fails to accurately perceive environmental obstacles and executes unsafe motions.
Fig.~\ref{fig_real_world_failure_cases} illustrates representative collision failure cases encountered during our experiments. Such failures are particularly pronounced in monolithic frameworks that rely solely on RGB-D observations. As shown in Fig.~\ref{fig_real_world_failure_cases}(a)--(c), the absence of explicit geometric sensing, coupled with a limited field of view, prevents the model from reliably perceiving obstacles during inference, ultimately resulting in collisions.
Fig.~\ref{fig_real_world_failure_cases}(d) presents a collision failure case for the hierarchical framework. In this scenario, the root cause lies in the presence of obstacles located below the LiDAR scanning plane. These obstacles are incorrectly classified as traversable regions during SLAM-based mapping, and the waypoint selection module lacks mechanisms to explicitly reject potentially unreachable or unsafe target points. As a result, the robot fails to avoid the obstacle during execution.

Notably, collision-related failure modes of this severity have received limited attention in prior VLN literature. This is largely because most existing studies are conducted in simulation environments, where such issues are either abstracted away or mitigated through simulator-specific mechanisms (\textit{e.g.}, sliding along object boundaries to escape collision-induced deadlocks). In contrast, collisions in real-world environments are difficult to avoid, often catastrophic, and directly impact both task success and operational safety. The primary causes of such collisions in current VLN models stem from limitations in their visual perception and memory modules: (1) the visual modality has limited capability to accurately resolve obstacles and often lacks explicit supervision during training; and (2) the memory module fails to retain long-term representations of potential obstacle regions, particularly when direct observations are lost due to restricted local views, which can lead the model to produce unsafe action predictions.

\section{Limitations and Future Directions}
\label{sec_limitations}

\begin{figure}[htbp]
    \centering
    \includegraphics[width=\linewidth]{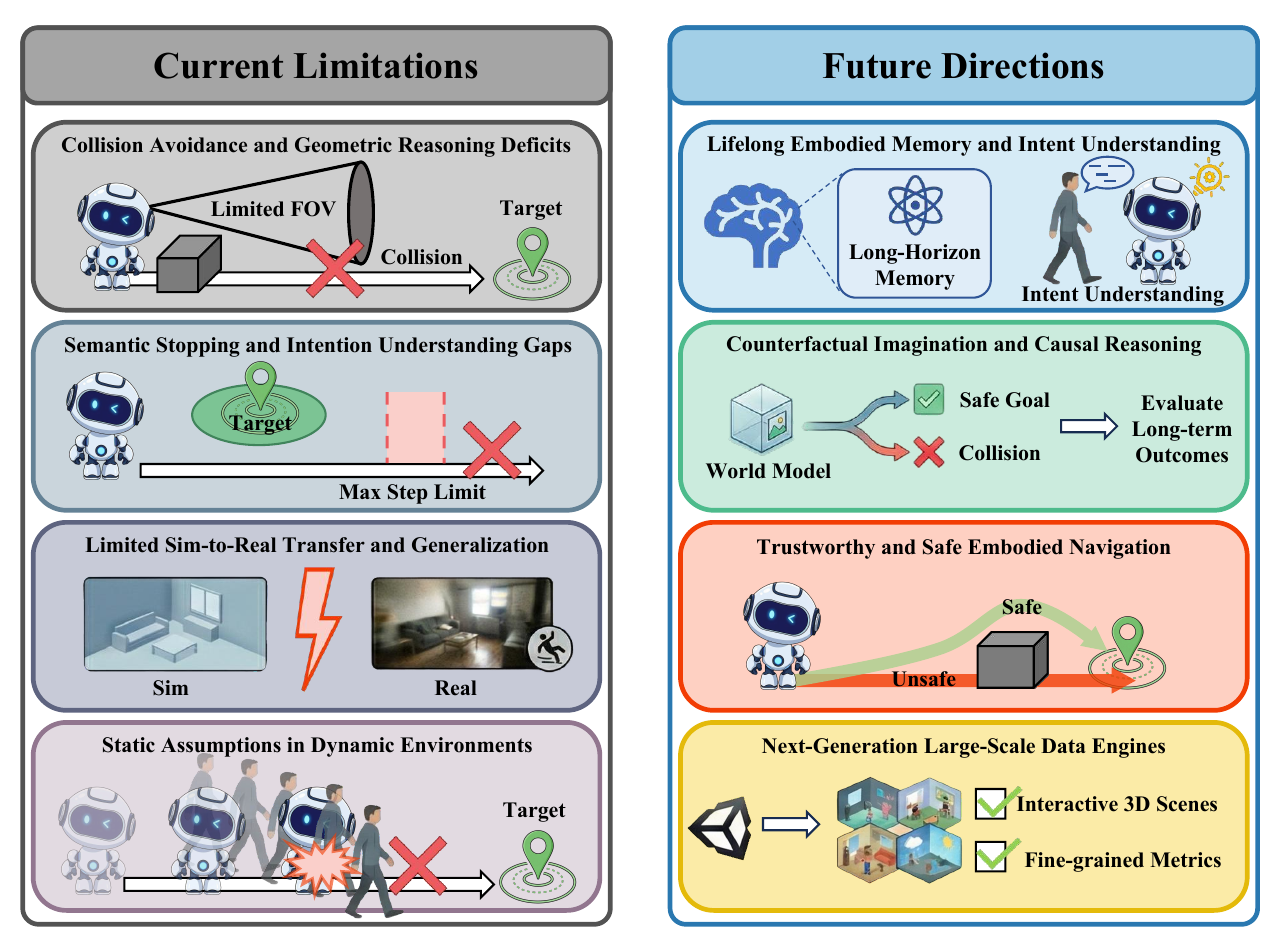}
    \caption{Limitations and future directions for VLN research.}
    \label{fig_limitations_future_directions}
\end{figure}

As shown in Fig.~\ref{fig_limitations_future_directions}, this section summarizes the limitations and outlines directions for future VLN research.

\subsection{Limitations}

\subsubsection{Collision Avoidance and Geometric Reasoning Deficits}

Real-world experiments reveal that collisions are a dominant failure mode limiting robust VLN deployment. \textit{First}, most simulator-generated datasets assume a fixed robot footprint and rely on shortest-path planners, producing trajectories that closely follow obstacles and provide limited supervision on safety margins and dynamic clearance. \textit{Second}, existing monolithic VLN agents lack persistent obstacle-aware memory: obstacles observed at a distance may disappear from the local field of view as the robot approaches, causing the policy to incorrectly assume free space ahead and execute unsafe actions. \textit{Third}, hierarchical frameworks often select waypoints without considering robot-specific kinematic and safety constraints, such as avoiding stairs for wheeled platforms while permitting them for legged robots. These limitations point to the need for enhanced geometric reasoning, obstacle-aware memory, and platform-aware safety constraints in real-world VLN. \newcontent{From a deployment-risk perspective, these limitations also highlight the need for safety mechanisms when VLN agents operate in shared human environments, where collisions, unexpected pedestrian motion, and socially inappropriate navigation behaviors may cause practical risks. Therefore, current VLN systems should be deployed only with reliable low-level collision checking, emergency stopping, conservative speed control, and human-aware local planning. Real-world VLN evaluations should thus be interpreted as controlled deployment studies rather than evidence of readiness for unrestricted operation in open shared spaces.}

\subsubsection{Semantic Stopping and Intention Understanding Gaps}

Current VLN algorithms demonstrate substantial difficulty in determining when to issue a \emph{stop} action based on semantic correctness rather than coarse distance-based criteria. This semantic misalignment becomes even more pronounced in intention and backtracking instructions, where agents often pass through semantically correct target regions but continue navigating until reaching maximum step limits due to the absence of explicit termination cues. Furthermore, intention tasks requiring spatial reasoning to infer underspecified goals remain highly challenging, with both paradigms demonstrating limited capability in grounding naturalistic human commands. As VLN systems are increasingly integrated with downstream embodied tasks such as vision-language manipulation~\cite{ahn2022saycan,brohan2023rt2}, where precise spatial grounding directly affects task feasibility, improving semantic termination decisions and intention tracking represents a critical yet underexplored challenge.

\subsubsection{Limited Sim-to-Real Transfer and Generalization}

Despite advances in large-scale pre-training and vision-language foundation models, our real-world evaluation \newcontent{shows that, under the tested system configurations, current VLN agents still face significant challenges in transferring from simulation to real-world environments, including environment categories that are well represented in existing training datasets.} The substantial sim-to-real performance gap \newcontent{suggests important transferability limitations, while also reflecting the combined influence of architecture, sensing, mapping, and control-stack differences in practical deployment.} Contributing factors include: (i) \textit{visual domain shift}, as models trained on photorealistic but static datasets struggle with real-world camera height change, motion blur, dynamic lighting, and sensor noise; (ii) \textit{physics mismatch}, where simulators assume perfect odometry and instantaneous actuation, whereas real robots experience drift, wheel slippage, and momentum-induced control lag; and (iii) \textit{distributional mismatch} in instruction complexity and environmental diversity.

\newcontent{These findings indicate that improving sim-to-real transfer requires not only stronger data-driven generalization, but also better integration of physical priors, robust perception, reliable mapping, and closed-loop control. Since exhaustively replicating real-world variability within simulators is inherently difficult, a promising direction is to incorporate world models or physics-informed formulations that capture fundamental physical constraints and reduce the sensitivity of VLN agents to irrelevant perceptual noise. In parallel, enriching training data with greater diversity in scenes, instructions, sensing conditions, and physical constraints remains essential for improving real-world generalization.}

\subsubsection{Static Assumptions in Dynamic Environments}

Most existing VLN datasets and benchmarks adopt a ``frozen world” assumption, in which environmental states remain static throughout navigation episodes. Agents trained under such assumptions struggle to cope with dynamic obstacles (\textit{e.g.,} humans or pets) and environmental changes (\textit{e.g.,} rearranged furniture), as their memory and spatial reasoning modules implicitly presume temporal invariance of the observed world.
In real-world deployments, particularly in crowded environments, dynamic obstacles frequently cause target occlusion, trajectory deviation, and loss of localization. Some experiments reveal that when goal-related objects become temporarily occluded or displaced, agents may unpredictably alter their routes and execute incorrect actions. These failure modes are largely absent from current simulators and datasets. Together, these observations highlight the need for more principled world modeling and explicit temporal reasoning to support robust VLN in dynamic, real-world environments.

\subsection{Future Directions}

\subsubsection{Lifelong Embodied Memory and Intent Understanding}

\newcontent{Lifelong VLN is a key direction for moving from benchmark-oriented navigation to real-world embodied deployment~\cite{krantz2023iterative,Jeong_2025_BMVC,wang2026allday,li2024vision,jiang2026m3e}. Unlike conventional VLN tasks that reset the agent after each episode, lifelong VLN requires agents to execute multiple instructions in the same or related environments, reuse accumulated knowledge, and maintain persistent spatial awareness. A central challenge is to build persistent yet adaptive embodied memory, which should organize historical observations and trajectories into structured spatial-semantic representations that can be queried by future instructions. Future methods may combine topological maps, semantic scene graphs, episodic memory, and foundation-model-based retrieval to support long-term spatial recall and context-aware decision making.}

Equally important is intent understanding beyond surface-level instruction parsing. Natural language instructions often underspecify actions, omit intermediate goals, or rely on implicit human conventions. Robust VLN agents must should have the capability to infer latent navigational intent by jointly reasoning over linguistic cues, environmental context, and interaction history. 

\subsubsection{Counterfactual Imagination and Causal Reasoning}

Human navigation is inherently anticipatory, relying on counterfactual reasoning to evaluate alternative actions and foresee their consequences before execution. In contrast, most existing VLN agents remain largely reactive, selecting actions based on immediate observations without explicitly reasoning about how different decisions might lead to divergent outcomes. This limitation constrains their ability to recover from errors, adapt to dynamic changes, and make safety-critical decisions.

World models offer a principled framework for enabling counterfactual imagination and causal reasoning in VLN. By learning predictive models of environment dynamics and agent-environment interactions, agents can internally simulate multiple hypothetical futures under alternative action sequences, allowing them to evaluate long-term outcomes and avoid failure modes such as collisions or dead ends. Integrating counterfactual simulation with causal inference mechanisms represents a promising direction toward more adaptive, resilient, and trustworthy VLN systems.

\subsubsection{Trustworthy and Safe Embodied Navigation}

VLN agents require navigation policies that are not only effective but also safe, reliable, and physically feasible. However, many existing approaches rely on idealized assumptions about collision-free execution, neglecting the physical constraints and safety requirements intrinsic to embodied systems. These methods often exhibit brittle behaviors, such as unsafe trajectories and collisions, hindering real-world deployment.

Future research should place greater emphasis on trustworthy navigation under physical constraints by explicitly incorporating kinematics, dynamics, and safety considerations into the decision-making process. This includes integrating motion feasibility constraints, uncertainty-aware collision avoidance, and risk-sensitive planning objectives into both high-level reasoning and low-level control. Safety evaluation benchmarks should place greater emphasis on metrics such as collision rate, entrapment rate, and compliance with social norms.

\subsubsection{Next-Generation Large-Scale Data Engines}

Progress in practical VLN is increasingly limited by the scale, diversity, and realism of existing datasets and simulators. Future research should move beyond static 3D scans toward generative data engines capable of producing large-scale, physically interactive environments. In particular, world models based on 3DGS and its variants provide an efficient foundation for photorealistic scene generation with fast rendering and low hardware requirements, especially when integrated with simulation platforms such as Unity, Isaac Sim, and Unreal Engine.

Equally critical is the advancement of instruction annotation and evaluation protocols. Next-generation data engines should support more accurate and diverse instruction generation, reflecting varied intents and linguistic styles, while adopting finer-grained evaluation metrics that go beyond coarse measures such as SR. Metrics targeting safety, physical feasibility, semantic grounding, and decision robustness are essential for meaningful comparison and sustained community progress.

\begin{figure}[htbp]
    \centering
    \includegraphics[width=\linewidth]{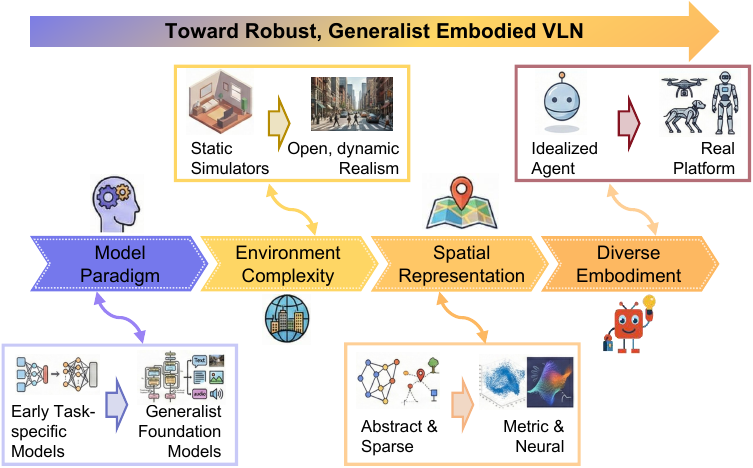}
    \caption{The long-term evolution trends of VLN research.}
    \label{fig_long_term_evolution_trends}
\end{figure}
Finally, Fig.~\ref{fig_long_term_evolution_trends} illustrates the long-term evolution trends of VLN research, highlighting a paradigm shift from specialized indoor navigation models toward generalist, large-scale embodied agents operating in open-world environments.

\section{Conclusion}
\label{sec_conclusion}

This survey provides a comprehensive and systematic analysis of embodied VLN tasks, organizing existing methods into hierarchical and monolithic paradigms, reviewing representative datasets, simulators, and SoTA algorithms, and critically analyzing their architectural principles and comparative characteristics. Crucially, \newcontent{we present a systematic real-world VLN evaluation on a physical robot platform, conducting experiments across 10 diverse real-world scenes to examine how representative VLN system configurations transfer from simulation to real-world deployment.} Our results \newcontent{reveal a substantial sim-to-real performance gap under the tested configurations.}

Building on these findings, we outline several promising directions for future research, including lifelong embodied memory and intent understanding, world-model-based counterfactual imagination and causal reasoning, trustworthy and safe navigation under physical constraints, and next-generation large-scale data engines. We hope this survey provides a timely and comprehensive synthesis of the VLN literature, while offering a valuable reference for future large-scale deployment, optimization, and evaluation in complex real-world environments, ultimately advancing the ability of embodied agents to reason and act robustly under environmental uncertainty.

\section*{Note to Practitioners}
For practitioners building mobile robots that navigate using natural language commands, this survey provides practical guidance on selecting, designing, and deploying VLN systems. Our empirical evaluations in real-world environments indicate that architectural choices have a substantial impact on both navigation performance and operational safety. 
At present, hierarchical navigation systems that combine panoramic visual perception with explicit 3D sensing and decompose navigation into high-level planning and low-level control achieve more reliable task success and lower collision frequencies.
At the same time, several practical limitations remain. Current VLN systems often struggle to determine when to stop based on the semantic intent of an instruction rather than simple distance-based heuristics. They also show limited robustness when faced with ambiguous or underspecified instructions that commonly arise in complex human-robot interactions. 
Continued advances in intent understanding, safety-aware reasoning, and real-world learning are expected to gradually enable more flexible and robust deployment in less structured settings.

% =============
% References
% =============
\bibliographystyle{IEEEtran}
\bibliography{main}

@String(CVPR= {IEEE Conf. Comput. Vis. Pattern Recog.})

@String(ICCV= {Int. Conf. Comput. Vis.})

@String(ECCV= {Eur. Conf. Comput. Vis.})

@String(ICPR = {Int. Conf. Pattern Recog.})

@String(BMVC= {Brit. Mach. Vis. Conf.})

@String(ICME = {Int. Conf. Multimedia and Expo})

@String(ICLR = {Int. Conf. Learn. Represent.})

@String(IJCAI = {IJCAI})

@String(AAAI = {AAAI})

@String(CVPR  = {CVPR})

@String(ICCV  = {ICCV})

@String(ECCV  = {ECCV})

@String(ICPR  = {ICPR})

@String(BMVC  =	{BMVC})

@String(ICME  =	{ICME})

@String(ICLR  = {ICLR})

@inproceedings{an2021neighbor,
  title={Neighbor-view enhanced model for vision and language navigation},
  author={An, Dong and Qi, Yuankai and Huang, Yan and Wu, Qi and Wang, Liang and Tan, Tieniu},
  booktitle={Proceedings of the 29th ACM International Conference on Multimedia},
  pages={5101--5109},
  year={2021}
}

@inproceedings{sun2021depth,
  title={Depth-Guided AdaIN and Shift Attention Network for Vision-And-Language Navigation},
  author={Sun, Qiang and Zhuang, Yifeng and Chen, Zhengqing and Fu, Yanwei and Xue, Xiangyang},
  booktitle={2021 IEEE International Conference on Multimedia and Expo (ICME)},
  pages={1--6},
  year={2021},
  organization={IEEE}
}

@article{zhou2021rethinking,
  title={Rethinking the Spatial Route Prior in Vision-and-Language Navigation},
  author={Zhou, Xinzhe and Liu, Wei and Mu, Yadong},
  journal={arXiv preprint arXiv:2110.05728},
  year={2021}
}

@inproceedings{zhu2020vision,
  title={Vision-language navigation with self-supervised auxiliary reasoning tasks},
  author={Zhu, Fengda and Zhu, Yi and Chang, Xiaojun and Liang, Xiaodan},
  booktitle={Proceedings of the IEEE/CVF Conference on Computer Vision and Pattern Recognition},
  pages={10012--10022},
  year={2020}
}

@inproceedings{ma2019self,
  title={Self-monitoring navigation agent via auxiliary progress estimation},
  author={Ma, Chih-Yao and Lu, Jiasen and Wu, Zuxuan and AlRegib, Ghassan and Kira, Zsolt and Socher, Richard and Xiong, Caiming},
  booktitle = {Proceedings of the International Conference on Learning Representations (ICLR)},
  year={2019}
}

@inproceedings{ma2019regretful,
  title={The regretful agent: Heuristic-aided navigation through progress estimation},
  author={Ma, Chih-Yao and Wu, Zuxuan and AlRegib, Ghassan and Xiong, Caiming and Kira, Zsolt},
  booktitle={Proceedings of the IEEE/CVF Conference on Computer Vision and Pattern Recognition},
  pages={6732--6740},
  year={2019}
}

@article{fried2018speaker,
  title={Speaker-follower models for vision-and-language navigation},
  author={Fried, Daniel and Hu, Ronghang and Cirik, Volkan and Rohrbach, Anna and Andreas, Jacob and Morency, Louis-Philippe and Berg-Kirkpatrick, Taylor and Saenko, Kate and Klein, Dan and Darrell, Trevor},
  journal={Advances in Neural Information Processing Systems},
  volume={31},
  year={2018}
}

@article{magassouba2021crossmap,
  title={CrossMap transformer: A crossmodal masked path transformer using double back-translation for vision-and-language navigation},
  author={Magassouba, Aly and Sugiura, Komei and Kawai, Hisashi},
  journal={IEEE Robotics and Automation Letters},
  volume={6},
  number={4},
  pages={6258--6265},
  year={2021},
  publisher={IEEE}
}

@inproceedings{tan2019learning,
  title={Learning to Navigate Unseen Environments: Back Translation with Environmental Dropout},
  author={Tan, Hao and Yu, Licheng and Bansal, Mohit},
  booktitle={Proceedings of the 2019 Conference of the North American Chapter of the Association for Computational Linguistics: Human Language Technologies, Volume 1 (Long and Short Papers)},
  pages={2610--2621},
  year={2019}
}

@inproceedings{hong2021vln,
  title={VLN BERT: A Recurrent Vision-and-Language BERT for Navigation},
  author={Hong, Yicong and Wu, Qi and Qi, Yuankai and Rodriguez-Opazo, Cristian and Gould, Stephen},
  booktitle={Proceedings of the IEEE/CVF Conference on Computer Vision and Pattern Recognition},
  pages={1643--1653},
  year={2021}
}

@inproceedings{majumdar2020improving,
  title={Improving vision-and-language navigation with image-text pairs from the web},
  author={Majumdar, Arjun and Shrivastava, Ayush and Lee, Stefan and Anderson, Peter and Parikh, Devi and Batra, Dhruv},
  booktitle={European Conference on Computer Vision},
  pages={259--274},
  year={2020},
  organization={Springer}
}

@inproceedings{hao2020towards,
  title={Towards learning a generic agent for vision-and-language navigation via pre-training},
  author={Hao, Weituo and Li, Chunyuan and Li, Xiujun and Carin, Lawrence and Gao, Jianfeng},
  booktitle={Proceedings of the IEEE/CVF Conference on Computer Vision and Pattern Recognition},
  pages={13137--13146},
  year={2020}
}

@article{chen2021history,
  title={History Aware Multimodal Transformer for Vision-and-Language Navigation},
  author={Chen, Shizhe and Guhur, Pierre-Louis and Schmid, Cordelia and Laptev, Ivan},
  journal={Advances in Neural Information Processing Systems},
  volume={34},
  year={2021}
}

@inproceedings{irshad2021hierarchical,
  title={Hierarchical cross-modal agent for robotics vision-and-language navigation},
  author={Irshad, Muhammad Zubair and Ma, Chih-Yao and Kira, Zsolt},
  booktitle={2021 IEEE international conference on robotics and automation (ICRA)},
  pages={13238--13246},
  year={2021},
  organization={IEEE}
}

@inproceedings{irshad2021sasra,
  title={Semantically-aware spatio-temporal reasoning agent for vision-and-language navigation in continuous environments},
  author={Irshad, Muhammad Zubair and Mithun, Niluthpol Chowdhury and Seymour, Zachary and Chiu, Han-Pang and Samarasekera, Supun and Kumar, Rakesh},
  booktitle={2022 26th International conference on pattern recognition (ICPR)},
  pages={4065--4071},
  year={2022},
  organization={IEEE}
}

@inproceedings{krantz2021waypoint,
  title={Waypoint Models for Instruction-guided Navigation in Continuous Environments},
  author={Krantz, Jacob and Gokaslan, Aaron and Batra, Dhruv and Lee, Stefan and Maksymets, Oleksandr},
  booktitle={Proceedings of the IEEE/CVF International Conference on Computer Vision},
  pages={15162--15171},
  year={2021}
}

@inproceedings{raychaudhuri2021language,
  title={Language-aligned waypoint (law) supervision for vision-and-language navigation in continuous environments},
  author={Raychaudhuri, Sonia and Wani, Saim and Patel, Shivansh and Jain, Unnat and Chang, Angel},
  booktitle={Proceedings of the 2021 conference on empirical methods in natural language processing},
  pages={4018--4028},
  year={2021}
}

@article{chang2017matterport3d,
  title={{Matterport3D}: Learning from {RGB-D} Data in Indoor Environments},
  author={Chang, Angel and Dai, Angela and Funkhouser, Thomas and Halber, Maciej and Niessner, Matthias and Savva, Manolis and Song, Shuran and Zeng, Andy and Zhang, Yinda},
  journal={International Conference on 3D Vision (3DV)},
  year={2017}
}

@inproceedings{hong2020sub,
  title={Sub-Instruction Aware Vision-and-Language Navigation},
  author={Hong, Yicong and Rodriguez, Cristian and Wu, Qi and Gould, Stephen},
  booktitle={Proceedings of the 2020 Conference on Empirical Methods in Natural Language Processing (EMNLP)},
  pages={3360--3376},
  year={2020}
}

@inproceedings{ku2020room,
  title={Room-Across-Room: Multilingual Vision-and-Language Navigation with Dense Spatiotemporal Grounding},
  author={Ku, Alexander and Anderson, Peter and Patel, Roma and Ie, Eugene and Baldridge, Jason},
  booktitle={Proceedings of the 2020 Conference on Empirical Methods in Natural Language Processing (EMNLP)},
  pages={4392--4412},
  year={2020}
}

@inproceedings{qi2020reverie,
  title={Reverie: Remote embodied visual referring expression in real indoor environments},
  author={Qi, Yuankai and Wu, Qi and Anderson, Peter and Wang, Xin and Wang, William Yang and Shen, Chunhua and Hengel, Anton van den},
  booktitle={Proceedings of the IEEE/CVF Conference on Computer Vision and Pattern Recognition},
  pages={9982--9991},
  year={2020}
}

@inproceedings{anderson2018vision,
  title={Vision-and-language navigation: Interpreting visually-grounded navigation instructions in real environments},
  author={Anderson, Peter and Wu, Qi and Teney, Damien and Bruce, Jake and Johnson, Mark and S{\"u}nderhauf, Niko and Reid, Ian and Gould, Stephen and Van Den Hengel, Anton},
  booktitle={Proceedings of the IEEE Conference on Computer Vision and Pattern Recognition},
  pages={3674--3683},
  year={2018}
}

@inproceedings{he2016deep,
  title={Deep residual learning for image recognition},
  author={He, Kaiming and Zhang, Xiangyu and Ren, Shaoqing and Sun, Jian},
  booktitle={Proceedings of the IEEE conference on computer vision and pattern recognition},
  pages={770--778},
  year={2016}
}

@inproceedings{vaswani2017attention,
  title={Attention is all you need},
  author={Vaswani, Ashish and Shazeer, Noam and Parmar, Niki and Uszkoreit, Jakob and Jones, Llion and Gomez, Aidan N and Kaiser, {\L}ukasz and Polosukhin, Illia},
  booktitle={Advances in neural information processing systems},
  pages={5998--6008},
  year={2017}
}

@inproceedings{liang2022contrastive,
  title={Contrastive instruction-trajectory learning for vision-language navigation},
  author={Liang, Xiwen and Zhu, Fengda and Zhu, Yi and Lin, Bingqian and Wang, Bing and Liang, Xiaodan},
  booktitle={Proceedings of the AAAI Conference on Artificial Intelligence},
  volume={36},
  number={2},
  pages={1592--1600},
  year={2022}
}

@article{wang2020vision,
  title={Vision-language navigation policy learning and adaptation},
  author={Wang, Xin and Huang, Qiuyuan and Celikyilmaz, Asli and Gao, Jianfeng and Shen, Dinghan and Wang, Yuan-Fang and Wang, William and Zhang, Lei},
  journal={IEEE transactions on pattern analysis and machine intelligence},
  year={2020},
  publisher={IEEE}
}

@inproceedings{jain2019stay,
  title={Stay on the Path: Instruction Fidelity in Vision-and-Language Navigation},
  author={Jain, Vihan and Magalhaes, Gabriel and Ku, Alexander and Vaswani, Ashish and Ie, Eugene and Baldridge, Jason},
  booktitle={Proceedings of the 57th Annual Meeting of the Association for Computational Linguistics},
  pages={1862--1872},
  year={2019}
}

@inproceedings{chen2022think,
  title={Think Global, Act Local: Dual-scale Graph Transformer for Vision-and-Language Navigation},
  author={Chen, Shizhe and Guhur, Pierre-Louis and Tapaswi, Makarand and Schmid, Cordelia and Laptev, Ivan},
  booktitle={Proceedings of the IEEE/CVF Conference on Computer Vision and Pattern Recognition},
  pages={16537--16547},
  year={2022}
}

@inproceedings{li2022envedit,
  title={EnvEdit: Environment Editing for Vision-and-Language Navigation},
  author={Li, Jialu and Tan, Hao and Bansal, Mohit},
  booktitle={Proceedings of the IEEE/CVF Conference on Computer Vision and Pattern Recognition},
  pages={15407--15417},
  year={2022}
}

@inproceedings{qiao2022hop,
  title={HOP: History-and-Order Aware Pre-training for Vision-and-Language Navigation},
  author={Qiao, Yanyuan and Qi, Yuankai and Hong, Yicong and Yu, Zheng and Wang, Peng and Wu, Qi},
  booktitle={Proceedings of the IEEE/CVF Conference on Computer Vision and Pattern Recognition},
  pages={15418--15427},
  year={2022}
}

@inproceedings{shen2021much,
  title={How Much Can CLIP Benefit Vision-and-Language Tasks?},
  author={Shen, Sheng and Li, Liunian Harold and Tan, Hao and Bansal, Mohit and Rohrbach, Anna and Chang, Kai-Wei and Yao, Zhewei and Keutzer, Kurt},
  booktitle={International Conference on Learning Representations}
}

@inproceedings{tan2019lxmert,
  title={LXMERT: Learning Cross-Modality Encoder Representations from Transformers},
  author={Tan, Hao and Bansal, Mohit},
  booktitle={Proceedings of the 2019 Conference on Empirical Methods in Natural Language Processing and the 9th International Joint Conference on Natural Language Processing (EMNLP-IJCNLP)},
  pages={5100--5111},
  year={2019}
}

@InProceedings{Zhu_2021_SOON,
    author    = {Zhu, Fengda and Liang, Xiwen and Zhu, Yi and Yu, Qizhi and Chang, Xiaojun and Liang, Xiaodan},
    title     = {SOON: Scenario Oriented Object Navigation With Graph-Based Exploration},
    booktitle = {Proceedings of the IEEE/CVF Conference on Computer Vision and Pattern Recognition (CVPR)},
    month     = {June},
    year      = {2021},
    pages     = {12689-12699}
}

@inproceedings{krantz_beyond_2020,
	location = {Cham},
	title = {Beyond the Nav-Graph: Vision-and-Language Navigation in Continuous Environments},
	isbn = {978-3-030-58604-1},
	pages = {104--120},
	booktitle = {Computer Vision – {ECCV} 2020},
	publisher = {Springer International Publishing},
	author = {Krantz, Jacob and Wijmans, Erik and Majumdar, Arjun and Batra, Dhruv and Lee, Stefan},
	editor = {Vedaldi, Andrea and Bischof, Horst and Brox, Thomas and Frahm, Jan-Michael},
	date = {2020},
}

@inproceedings{zhao2022target,
  title={Target-driven structured transformer planner for vision-language navigation},
  author={Zhao, Yusheng and Chen, Jinyu and Gao, Chen and Wang, Wenguan and Yang, Lirong and Ren, Haibing and Xia, Huaxia and Liu, Si},
  booktitle={Proceedings of the 30th ACM International Conference on Multimedia},
  pages={4194--4203},
  year={2022}
}

@inproceedings{wang2023dual,
  title={A Dual Semantic-Aware Recurrent Global-Adaptive Network For Vision-and-Language Navigation},
  author={Wang, Liuyi and He, Zongtao and Tang, jiagui and Dang, Ronghao and Wang, naijia and Liu, Chengju and Chen, Qijun},
  booktitle={International Joint Conferences on Artificial Intelligence (IJCAI)},
  year={2023}
}

@inproceedings{huang2019multi,
  title={Multi-modal Discriminative Model for Vision-and-Language Navigation},
  author={Huang, Haoshuo and Jain, Vihan and Mehta, Harsh and Baldridge, Jason and Ie, Eugene},
  booktitle={Proceedings of the Combined Workshop on Spatial Language Understanding (SpLU) and Grounded Communication for Robotics (RoboNLP)},
  pages={40--49},
  year={2019}
}

@inproceedings{dang2022unbiased,
author = {Dang, Ronghao and Shi, Zhuofan and Wang, Liuyi and He, Zongtao and Liu, Chengju and Chen, Qijun},
title = {Unbiased Directed Object Attention Graph for Object Navigation},
year = {2022},
publisher = {Association for Computing Machinery},
address = {New York, NY, USA},
booktitle = {Proceedings of the 30th ACM International Conference on Multimedia},
pages = {3617–3627},
numpages = {11},
location = {Lisboa, Portugal},
series = {MM '22}
}

@inproceedings{dang2023search,
  title={Search for or navigate to? dual adaptive thinking for object navigation},
  author={Dang, Ronghao and Wang, Liuyi and He, Zongtao and Su, Shuai and Tang, Jiagui and Liu, Chengju and Chen, Qijun},
  booktitle={Proceedings of the IEEE/CVF International Conference on Computer Vision},
  pages={8250--8259},
  year={2023}
}

@inproceedings{qi2021road,
  title={The road to know-where: An object-and-room informed sequential bert for indoor vision-language navigation},
  author={Qi, Yuankai and Pan, Zizheng and Hong, Yicong and Yang, Ming-Hsuan and van den Hengel, Anton and Wu, Qi},
  booktitle={Proceedings of the IEEE/CVF International Conference on Computer Vision},
  pages={1655--1664},
  year={2021}
}

@article{moudgil2021soat,
  title={Soat: A scene-and object-aware transformer for vision-and-language navigation},
  author={Moudgil, Abhinav and Majumdar, Arjun and Agrawal, Harsh and Lee, Stefan and Batra, Dhruv},
  journal={Advances in Neural Information Processing Systems},
  volume={34},
  pages={7357--7367},
  year={2021}
}

@inproceedings{chen2022reinforced,
  title={Reinforced Structured State-Evolution for Vision-Language Navigation},
  author={Chen, Jinyu and Gao, Chen and Meng, Erli and Zhang, Qiong and Liu, Si},
  booktitle={Proceedings of the IEEE/CVF Conference on Computer Vision and Pattern Recognition},
  pages={15450--15459},
  year={2022}
}

@inproceedings{qi2020object,
  title={Object-and-action aware model for visual language navigation},
  author={Qi, Yuankai and Pan, Zizheng and Zhang, Shengping and van den Hengel, Anton and Wu, Qi},
  booktitle={Computer Vision--ECCV 2020: 16th European Conference, Glasgow, UK, August 23--28, 2020, Proceedings, Part X 16},
  pages={303--317},
  year={2020},
  organization={Springer}
}

@article{qiao2023hop_plus,
  title={HOP+: History-enhanced and Order-aware Pre-training for Vision-and-Language Navigation},
  author={Qiao, Yanyuan and Qi, Yuankai and Hong, Yicong and Yu, Zheng and Wang, Peng and Wu, Qi},
  journal={IEEE Transactions on Pattern Analysis and Machine Intelligence},
  year={2023},
  publisher={IEEE}
}

@article{wang2023res,
  title={RES-StS: Referring Expression Speaker via Self-training with Scorer for Goal-Oriented Vision-Language Navigation},
  author={Wang, Liuyi and He, Zongtao and Dang, Ronghao and Chen, Huiyi and Liu, Chengju and Chen, Qijun},
  journal={IEEE Transactions on Circuits and Systems for Video Technology},
  year={2023},
  publisher={IEEE}
}

@inproceedings{dou2022foam,
  title={FOAM: A Follower-aware Speaker Model For Vision-and-Language Navigation},
  author={Dou, Zi-Yi and Peng, Nanyun},
  booktitle={Proceedings of the 2022 Conference of the North American Chapter of the Association for Computational Linguistics: Human Language Technologies},
  pages={4332--4340},
  year={2022}
}

@inproceedings{magalhaes2019general,
  title={General evaluation for instruction conditioned navigation using dynamic time warping},
  author={Magalhaes, Gabriel Ilharco and Jain, Vihan and Ku, Alexander and Ie, Eugene and Baldridge, Jason},
  booktitle={NeurIPS Visually Grounded Interaction and Language (ViGIL) Workshop},
  volume={1},
  year={2019}
}

@inproceedings{zhang2021diagnosing,
author = {Zhang, Yubo and Tan, Hao and Bansal, Mohit},
title = {Diagnosing the Environment Bias in Vision-and-Language Navigation},
year = {2021},
isbn = {9780999241165},
booktitle = {Proceedings of the Twenty-Ninth International Joint Conference on Artificial Intelligence},
articleno = {124},
numpages = {8},
location = {Yokohama, Yokohama, Japan},
series = {IJCAI'20}}

@article{wang2023pasts,
  title={Pasts: Progress-aware spatio-temporal transformer speaker for vision-and-language navigation},
  author={Wang, Liuyi and Liu, Chengju and He, Zongtao and Li, Shu and Yan, Qingqing and Chen, Huiyi and Chen, Qijun},
  journal={Engineering Applications of Artificial Intelligence},
  volume={128},
  pages={107487},
  year={2024},
  publisher={Elsevier}
}

@inproceedings{kamath2023new,
  title={A New Path: Scaling Vision-and-Language Navigation with Synthetic Instructions and Imitation Learning},
  author={Kamath, Aishwarya and Anderson, Peter and Wang, Su and Koh, Jing Yu and Ku, Alexander and Waters, Austin and Yang, Yinfei and Baldridge, Jason and Parekh, Zarana},
  booktitle={Proceedings of the IEEE/CVF Conference on Computer Vision and Pattern Recognition},
  pages={10813--10823},
  year={2023}
}

@inproceedings{wang2022less,
  title={Less is More: Generating Grounded Navigation Instructions from Landmarks},
  author={Wang, Su and Montgomery, Ceslee and Orbay, Jordi and Birodkar, Vighnesh and Faust, Aleksandra and Gur, Izzeddin and Jaques, Natasha and Waters, Austin and Baldridge, Jason and Anderson, Peter},
  booktitle={Proceedings of the IEEE/CVF Conference on Computer Vision and Pattern Recognition},
  pages={15428--15438},
  year={2022}
}

@article{he2023mlanet,
  title = {A Multilevel Attention Network with Sub-Instructions for Continuous Vision-and-Language Navigation},
  author = {He, Zongtao and Wang, Liuyi and Li, Shu and Yan, Qingqing and Liu, Chengju and Chen, Qijun},
  year = {2025},
  month = apr,
  journal = {Applied Intelligence},
  volume = {55},
  number = {7},
  pages = {657},
  issn = {1573-7497},
  doi = {10.1007/s10489-025-06544-9},
}

@inproceedings{hu2019you,
  title={Are You Looking? Grounding to Multiple Modalities in Vision-and-Language Navigation},
  author={Hu, Ronghang and Fried, Daniel and Rohrbach, Anna and Klein, Dan and Darrell, Trevor and Saenko, Kate},
  booktitle={Proceedings of the 57th Annual Meeting of the Association for Computational Linguistics},
  pages={6551--6557},
  year={2019}
}

@inproceedings{cui2023grounded,
  title={Grounded Entity-Landmark Adaptive Pre-training for Vision-and-Language Navigation},
  author={Cui, Yibo and Xie, Liang and Zhang, Yakun and Zhang, Meishan and Yan, Ye and Yin, Erwei},
  booktitle={Proceedings of the IEEE/CVF International Conference on Computer Vision},
  pages={12043--12053},
  year={2023}
}

@inproceedings{liu2023bird,
  title={Bird's-Eye-View Scene Graph for Vision-Language Navigation},
  author={Liu, Rui and Wang, Xiaohan and Wang, Wenguan and Yang, Yi},
  booktitle={Proceedings of the IEEE/CVF International Conference on Computer Vision},
  pages={10968--10980},
  year={2023}
}

@article{an2022bevbert,
  title={BEVBert: Multimodal Map Pre-training for Language-guided Navigation},
  author={An, Dong and Qi, Yuankai and Li, Yangguang and Huang, Yan and Wang, Liang and Tan, Tieniu and Shao, Jing},
  journal={Proceedings of the IEEE/CVF International Conference on Computer Vision},
  year={2023}
}

@inproceedings{wang2023scaling,
  title={Scaling data generation in vision-and-language navigation},
  author={Wang, Zun and Li, Jialu and Hong, Yicong and Wang, Yi and Wu, Qi and Bansal, Mohit and Gould, Stephen and Tan, Hao and Qiao, Yu},
  booktitle={Proceedings of the IEEE/CVF International Conference on Computer Vision},
  pages={12009--12020},
  year={2023}
}

@inproceedings{huo2023geovln,
  title={GeoVLN: Learning Geometry-Enhanced Visual Representation with Slot Attention for Vision-and-Language Navigation},
  author={Huo, Jingyang and Sun, Qiang and Jiang, Boyan and Lin, Haitao and Fu, Yanwei},
  booktitle={Proceedings of the IEEE/CVF Conference on Computer Vision and Pattern Recognition},
  pages={23212--23221},
  year={2023}
}

@inproceedings{wang2023gridmm,
  title={GridMM: Grid Memory Map for Vision-and-Language Navigation},
  author={Wang, Zihan and Li, Xiangyang and Yang, Jiahao and Liu, Yeqi and Jiang, Shuqiang},
  booktitle={Proceedings of the IEEE/CVF International Conference on Computer Vision},
  pages={15625--15636},
  year={2023}
}

@inproceedings{ramakrishnan2021habitat,
 author = {Ramakrishnan, Santhosh K and Gokaslan, Aaron and Wijmans, Erik and Maksymets, Oleksandr and Clegg, Alex and Turner, John and Undersander, Eric and Galuba, Wojciech and Westbury, Andrew and Chang, Angel X and others},
 booktitle = {Proceedings of the Neural Information Processing Systems Track on Datasets and Benchmarks},
 pages = {},
 publisher = {Curran},
 title = {Habitat-Matterport 3D Dataset (HM3D): 1000 Large-scale 3D Environments for Embodied AI},
 volume = {1},
 year = {2021}
}

@inproceedings{chen2022learning,
  title={Learning from unlabeled 3d environments for vision-and-language navigation},
  author={Chen, Shizhe and Guhur, Pierre-Louis and Tapaswi, Makarand and Schmid, Cordelia and Laptev, Ivan},
  booktitle={European Conference on Computer Vision},
  pages={638--655},
  year={2022},
  organization={Springer}
}

@article{parvaneh2020counterfactual,
  title={Counterfactual vision-and-language navigation: Unravelling the unseen},
  author={Parvaneh, Amin and Abbasnejad, Ehsan and Teney, Damien and Shi, Javen Qinfeng and Van den Hengel, Anton},
  journal={Advances in Neural Information Processing Systems},
  volume={33},
  pages={5296--5307},
  year={2020}
}

@inproceedings{nguyen2019help,
  title={Help, Anna! Visual Navigation with Natural Multimodal Assistance via Retrospective Curiosity-Encouraging Imitation Learning},
  author={Nguyen, Khanh and Daum{\'e} III, Hal},
  booktitle={Proceedings of the 2019 Conference on Empirical Methods in Natural Language Processing and the 9th International Joint Conference on Natural Language Processing (EMNLP-IJCNLP)},
  pages={684--695},
  year={2019}
}

@inproceedings{thomason2020vision,
  title={Vision-and-dialog navigation},
  author={Thomason, Jesse and Murray, Michael and Cakmak, Maya and Zettlemoyer, Luke},
  booktitle={Conference on Robot Learning},
  pages={394--406},
  year={2020},
  organization={PMLR}
}

@article{he2023learning,
  title={Learning depth representation from RGB-D videos by time-aware contrastive pre-training},
  author={He, Zongtao and Wang, Liuyi and Dang, Ronghao and Li, Shu and Yan, Qingqing and Liu, Chengju and Chen, Qijun},
  journal={IEEE Transactions on Circuits and Systems for Video Technology},
  volume={34},
  number={6},
  pages={4143--4158},
  year={2023},
  publisher={IEEE}
}

@article{wang2024causal,
  title={Vision-and-Language Navigation via Causal Learning},
  author={Wang, Liuyi and Tang, Jiagui and He, Zongtao and Dang, Ronghao and Liu, Chengju and Chen, Qijun},
  journal={Conference on Computer Vision and Pattern Recognition (CVPR)},
  year={2024}
}

@inproceedings{zhou2024navgpt,
  title={Navgpt: Explicit reasoning in vision-and-language navigation with large language models},
  author={Zhou, Gengze and Hong, Yicong and Wu, Qi},
  booktitle={Proceedings of the AAAI Conference on Artificial Intelligence},
  volume={38},
  number={7},
  pages={7641--7649},
  year={2024}
}

@inproceedings{chen2024mapgpt,
  title={Mapgpt: Map-guided prompting with adaptive path planning for vision-and-language navigation},
  author={Chen, Jiaqi and Lin, Bingqian and Xu, Ran and Chai, Zhenhua and Liang, Xiaodan and Wong, Kwan-Yee},
  booktitle={Proceedings of the 62nd Annual Meeting of the Association for Computational Linguistics (Volume 1: Long Papers)},
  pages={9796--9810},
  year={2024}
}

@inproceedings{xia2018gibson,
  title={Gibson env: Real-world perception for embodied agents},
  author={Xia, Fei and Zamir, Amir R and He, Zhiyang and Sax, Alexander and Malik, Jitendra and Savarese, Silvio},
  booktitle={Proceedings of the IEEE conference on computer vision and pattern recognition},
  pages={9068--9079},
  year={2018}
}

@ARTICLE{tan2024self,
  author={Tan, Sinan and Sima, Kuankuan and Wang, Dunzheng and Ge, Mengmeng and Guo, Di and Liu, Huaping},
  journal={IEEE Transactions on Neural Networks and Learning Systems}, 
  title={Self-Supervised 3-D Semantic Representation Learning for Vision-and-Language Navigation}, 
  year={2024},
  volume={},
  number={},
  pages={1-14},
  doi={10.1109/TNNLS.2024.3395633}}

@inproceedings{long2024instructnav,
  title={InstructNav: Zero-shot System for Generic Instruction Navigation in Unexplored Environment},
  author={Long, Yuxing and Cai, Wenzhe and Wang, Hongcheng and Zhan, Guanqi and Dong, Hao},
  booktitle={8th Annual Conference on Robot Learning}
}

@article{zhang2024navid,
        title={NaVid: Video-based VLM Plans the Next Step for Vision-and-Language Navigation},
        author={Zhang, Jiazhao and Wang, Kunyu and Xu, Rongtao and Zhou, Gengze and Hong, Yicong and Fang, Xiaomeng and Wu, Qi and Zhang, Zhizheng and Wang, He},
        journal={Robotics: Science and Systems},
        year={2024}
      }

@article{an2024etpnav,
  title={ETPNav: Evolving Topological Planning for Vision-Language Navigation in Continuous Environments},
  author={An, Dong and Wang, Hanqing and Wang, Wenguan and Wang, Zun and Huang, Yan and He, Keji and Wang, Liang},
  journal={IEEE Transactions on Pattern Analysis and Machine Intelligence},
  year={2024}
}

@inproceedings{sridhar2024nomad,
  title={Nomad: Goal masked diffusion policies for navigation and exploration},
  author={Sridhar, Ajay and Shah, Dhruv and Glossop, Catherine and Levine, Sergey},
  booktitle={2024 IEEE International Conference on Robotics and Automation (ICRA)},
  pages={63--70},
  year={2024},
  organization={IEEE}
}

@article{chi2023diffusion,
  title={Diffusion policy: Visuomotor policy learning via action diffusion},
  author={Chi, Cheng and Xu, Zhenjia and Feng, Siyuan and Cousineau, Eric and Du, Yilun and Burchfiel, Benjamin and Tedrake, Russ and Song, Shuran},
  journal={The International Journal of Robotics Research},
  pages={02783649241273668},
  year={2023},
  publisher={SAGE Publications Sage UK: London, England}
}

@inproceedings{hong2022bridging,
  title={Bridging the gap between learning in discrete and continuous environments for vision-and-language navigation},
  author={Hong, Yicong and Wang, Zun and Wu, Qi and Gould, Stephen},
  booktitle={Proceedings of the IEEE/CVF conference on computer vision and pattern recognition},
  pages={15439--15449},
  year={2022}
}

@article{li2024behavior,
  title={BEHAVIOR-1K: A Human-Centered, Embodied AI Benchmark with 1, 000 Everyday Activities and Realistic Simulation},
  author={Li, Chengshu and Zhang, Ruohan and Wong, Josiah and Gokmen, Cem and Srivastava, Sanjana and Mart{\'\i}n-Mart{\'\i}n, Roberto and Wang, Chen and Levine, Gabrael and Ai, Wensi and Martinez, Benjamin and others},
  journal={CoRR},
  year={2024}
}

@inproceedings{srivastava2022behavior,
  title={Behavior: Benchmark for everyday household activities in virtual, interactive, and ecological environments},
  author={Srivastava, Sanjana and Li, Chengshu and Lingelbach, Michael and Mart{\'\i}n-Mart{\'\i}n, Roberto and Xia, Fei and Vainio, Kent Elliott and Lian, Zheng and Gokmen, Cem and Buch, Shyamal and Liu, Karen and others},
  booktitle={Conference on robot learning},
  pages={477--490},
  year={2022},
  organization={PMLR}
}

@inproceedings{huang23vlmaps,
               title={Visual Language Maps for Robot Navigation},
               author={Chenguang Huang and Oier Mees and Andy Zeng and Wolfram Burgard},
               booktitle = {Proceedings of the IEEE International Conference on Robotics and Automation (ICRA)},
               year={2023},
               address = {London, UK}
}

@inproceedings{savva2019habitat,
  title={Habitat: A platform for embodied ai research},
  author={Savva, Manolis and Kadian, Abhishek and Maksymets, Oleksandr and Zhao, Yili and Wijmans, Erik and Jain, Bhavana and Straub, Julian and Liu, Jia and Koltun, Vladlen and Malik, Jitendra and others},
  booktitle={Proceedings of the IEEE/CVF international conference on computer vision},
  pages={9339--9347},
  year={2019}
}

@inproceedings{shridhar2020alfred,
  title={Alfred: A benchmark for interpreting grounded instructions for everyday tasks},
  author={Shridhar, Mohit and Thomason, Jesse and Gordon, Daniel and Bisk, Yonatan and Han, Winson and Mottaghi, Roozbeh and Zettlemoyer, Luke and Fox, Dieter},
  booktitle={Proceedings of the IEEE/CVF conference on computer vision and pattern recognition},
  pages={10740--10749},
  year={2020}
}

@article{deitke2022️procthor,
  title={ProcTHOR: Large-Scale Embodied AI Using Procedural Generation},
  author={Deitke, Matt and VanderBilt, Eli and Herrasti, Alvaro and Weihs, Luca and Ehsani, Kiana and Salvador, Jordi and Han, Winson and Kolve, Eric and Kembhavi, Aniruddha and Mottaghi, Roozbeh},
  journal={Advances in Neural Information Processing Systems},
  volume={35},
  pages={5982--5994},
  year={2022}
}

@inproceedings{ehsani2021manipulathor,
  title={Manipulathor: A framework for visual object manipulation},
  author={Ehsani, Kiana and Han, Winson and Herrasti, Alvaro and VanderBilt, Eli and Weihs, Luca and Kolve, Eric and Kembhavi, Aniruddha and Mottaghi, Roozbeh},
  booktitle={Proceedings of the IEEE/CVF conference on computer vision and pattern recognition},
  pages={4497--4506},
  year={2021}
}

@inproceedings{deitke2020robothor,
  title={Robothor: An open simulation-to-real embodied ai platform},
  author={Deitke, Matt and Han, Winson and Herrasti, Alvaro and Kembhavi, Aniruddha and Kolve, Eric and Mottaghi, Roozbeh and Salvador, Jordi and Schwenk, Dustin and VanderBilt, Eli and Wallingford, Matthew and others},
  booktitle={Proceedings of the IEEE/CVF conference on computer vision and pattern recognition},
  pages={3164--3174},
  year={2020}
}

@article{szot2021habitat,
  title={Habitat 2.0: Training home assistants to rearrange their habitat},
  author={Szot, Andrew and Clegg, Alexander and Undersander, Eric and Wijmans, Erik and Zhao, Yili and Turner, John and Maestre, Noah and Mukadam, Mustafa and Chaplot, Devendra Singh and Maksymets, Oleksandr and others},
  journal={Advances in neural information processing systems},
  volume={34},
  pages={251--266},
  year={2021}
}

@article{puig2023habitat3,
  title={Habitat 3.0: A Co-Habitat for Humans},
  author={Puig, Xavi and Undersander, Eric and Szot, Andrew and Cote, Mikael Dallaire and Partsey, Ruslan and Yang, Jimmy and Desai, Ruta and Clegg, Alexander William and Hlavac, Michal and Min, Tiffany and others},
  journal={Avatars and Robots},
  volume={4},
  year={2023}
}

@inproceedings{cheng2024navila,
        title={Navila: Legged robot vision-language-action model for navigation},
        author={Cheng, An-Chieh and Ji, Yandong and Yang, Zhaojing and Gongye, Zaitian and Zou, Xueyan and Kautz, Jan and B{\i}y{\i}k, Erdem and Yin, Hongxu and Liu, Sifei and Wang, Xiaolong},
        booktitle={RSS},
        year={2025}
}

@inproceedings{bar2025navigation,
  title={Navigation world models},
  author={Bar, Amir and Zhou, Gaoyue and Tran, Danny and Darrell, Trevor and LeCun, Yann},
  booktitle={Proceedings of the Computer Vision and Pattern Recognition Conference},
  pages={15791--15801},
  year={2025}
}

@inproceedings{gao2025openfly,
  title={OpenFly: A Versatile Toolchain and Large-scale Benchmark for Aerial Vision-Language Navigation},
  author={Gao, Yunpeng and Li, Chenhui and You, Zhongrui and Liu, Junli and Li, Zhen and Chen, Pengan and Chen, Qizhi and Tang, Zhonghan and Wang, Liansheng and Yang, Penghui and others},
  booktitle = {Proceedings of the International Conference on Learning Representations (ICLR)},
  year={2026}
}

@ARTICLE{he2025navcomposercomposinglanguageinstructions,
  author={He, Zongtao and Wang, Liuyi and Chen, Lu and Liu, Chengju and Chen, Qijun},
  journal={IEEE Transactions on Circuits and Systems for Video Technology}, 
  title={NavComposer: Composing Language Instructions for Navigation Trajectories through Action-Scene-Object Modularization}, 
  year={2025},
  volume={},
  number={},
  pages={1-1},
  doi={10.1109/TCSVT.2025.3596386}}

@inproceedings{jia2021scaling,
  title={Scaling up visual and vision-language representation learning with noisy text supervision},
  author={Jia, Chao and Yang, Yinfei and Xia, Ye and Chen, Yi-Ting and Parekh, Zarana and Pham, Hieu and Le, Quoc and Sung, Yun-Hsuan and Li, Zhen and Duerig, Tom},
  booktitle={International conference on machine learning},
  pages={4904--4916},
  year={2021},
  organization={PMLR}
}

@inproceedings{chen2019touchdown,
  title={Touchdown: Natural language navigation and spatial reasoning in visual street environments},
  author={Chen, Howard and Suhr, Alane and Misra, Dipendra and Snavely, Noah and Artzi, Yoav},
  booktitle={Proceedings of the IEEE/CVF Conference on Computer Vision and Pattern Recognition},
  pages={12538--12547},
  year={2019}
}

@article{zhao2025agrivln,
  title={AgriVLN: Vision-and-Language Navigation for Agricultural Robots},
  author={Zhao, Xiaobei and Lyu, Xingqi and Li, Xiang},
  journal={arXiv preprint arXiv:2508.07406},
  year={2025}
}

@inproceedings{zhou2024navgpt2,
  title={Navgpt-2: Unleashing navigational reasoning capability for large vision-language models},
  author={Zhou, Gengze and Hong, Yicong and Wang, Zun and Wang, Xin Eric and Wu, Qi},
  booktitle={European Conference on Computer Vision},
  pages={260--278},
  year={2024},
  organization={Springer}
}

@article{Qwen2.5-VL,
  title={Qwen2.5-VL Technical Report},
  author={Bai, Shuai and Chen, Keqin and Liu, Xuejing and Wang, Jialin and Ge, Wenbin and Song, Sibo and Dang, Kai and Wang, Peng and Wang, Shijie and Tang, Jun and Zhong, Humen and Zhu, Yuanzhi and Yang, Mingkun and Li, Zhaohai and Wan, Jianqiang and Wang, Pengfei and Ding, Wei and Fu, Zheren and Xu, Yiheng and Ye, Jiabo and Zhang, Xi and Xie, Tianbao and Cheng, Zesen and Zhang, Hang and Yang, Zhibo and Xu, Haiyang and Lin, Junyang},
  journal={arXiv preprint arXiv:2502.13923},
  year={2025}
}

@inproceedings{chen2024internvl,
  title={Internvl: Scaling up vision foundation models and aligning for generic visual-linguistic tasks},
  author={Chen, Zhe and Wu, Jiannan and Wang, Wenhai and Su, Weijie and Chen, Guo and Xing, Sen and Zhong, Muyan and Zhang, Qinglong and Zhu, Xizhou and Lu, Lewei and others},
  booktitle={Proceedings of the IEEE/CVF Conference on Computer Vision and Pattern Recognition},
  pages={24185--24198},
  year={2024}
}

@article{zhang2024uni,
    title={Uni-NaVid: A Video-based Vision-Language-Action Model for Unifying Embodied Navigation Tasks},
    author={Zhang, Jiazhao and Wang, Kunyu and Wang, Shaoan and Li, Minghan and Liu, Haoran and Wei, Songlin and Wang, Zhongyuan and Zhang, Zhizheng and Wang, He},
    journal={Robotics: Science and Systems},
    year={2025}
}

@inproceedings{wijmansdd2019ddppo,
  title={DD-PPO: Learning Near-Perfect PointGoal Navigators from 2.5 Billion Frames},
  author={Wijmans, Erik and Kadian, Abhishek and Morcos, Ari and Lee, Stefan and Essa, Irfan and Parikh, Devi and Savva, Manolis and Batra, Dhruv},
  booktitle={International Conference on Learning Representations}
}

@article{li2022reve-ce,
  title={Reve-ce: Remote embodied visual referring expression in continuous environment},
  author={Li, Xinghang and Guo, Di and Liu, Huaping and Sun, Fuchun},
  journal={IEEE Robotics and Automation Letters},
  volume={7},
  number={2},
  pages={1494--1501},
  year={2022},
  publisher={IEEE}
}

@inproceedings{wang2025g3d,
  title={g3d-lf: Generalizable 3d-language feature fields for embodied tasks},
  author={Wang, Zihan and Lee, Gim Hee},
  booktitle={Proceedings of the Computer Vision and Pattern Recognition Conference},
  pages={14191--14202},
  year={2025}
}

@article{chen2022weakly,
  title={Weakly-supervised multi-granularity map learning for vision-and-language navigation},
  author={Chen, Peihao and Ji, Dongyu and Lin, Kunyang and Zeng, Runhao and Li, Thomas and Tan, Mingkui and Gan, Chuang},
  journal={Advances in Neural Information Processing Systems},
  volume={35},
  pages={38149--38161},
  year={2022}
}

@inproceedings{yu2025correctnav,
  title={Correctnav: Self-correction flywheel empowers vision-language-action navigation model},
  author={Yu, Zhuoyuan and Long, Yuxing and Yang, Zihan and Zeng, Chengyan and Fan, Hongwei and Zhang, Jiyao and Dong, Hao},
  booktitle={Proceedings of the AAAI Conference on Artificial Intelligence},
  volume={40},
  number={22},
  pages={18737--18745},
  year={2026}
}

@inproceedings{georgakis2022cross,
  title={Cross-modal map learning for vision and language navigation},
  author={Georgakis, Georgios and Schmeckpeper, Karl and Wanchoo, Karan and Dan, Soham and Miltsakaki, Eleni and Roth, Dan and Daniilidis, Kostas},
  booktitle={Proceedings of the IEEE/CVF conference on computer vision and pattern recognition},
  pages={15460--15470},
  year={2022}
}

@inproceedings{koh2021pathdreamer,
  title={Pathdreamer: A world model for indoor navigation},
  author={Koh, Jing Yu and Lee, Honglak and Yang, Yinfei and Baldridge, Jason and Anderson, Peter},
  booktitle={Proceedings of the IEEE/CVF International Conference on Computer Vision},
  pages={14738--14748},
  year={2021}
}

@inproceedings{wang2023dreamwalker,
  title={Dreamwalker: Mental planning for continuous vision-language navigation},
  author={Wang, Hanqing and Liang, Wei and Van Gool, Luc and Wang, Wenguan},
  booktitle={Proceedings of the IEEE/CVF international conference on computer vision},
  pages={10873--10883},
  year={2023}
}

@inproceedings{yao2025navmorph,
  title     = {NavMorph: A Self-Evolving World Model for Vision-and-Language Navigation in Continuous Environments},
  author    = {Yao, Xuan and Gao, Junyu and Xu, Changsheng},
  booktitle = {Proceedings of the IEEE/CVF International Conference on Computer Vision (ICCV)},
  year      = {2025}
}

@article{li2023panogen,
  title={Panogen: Text-conditioned panoramic environment generation for vision-and-language navigation},
  author={Li, Jialu and Bansal, Mohit},
  journal={Advances in neural information processing systems},
  volume={36},
  pages={21878--21894},
  year={2023}
}

@inproceedings{xia2020interactive,
  title={Interactive Gibson Benchmark: A benchmark for interactive navigation in cluttered environments},
  author={Xia, Fei and R. Zamir, Amir and Toshev, Alexander and Malik, Jitendra and Savarese, Silvio},
  booktitle={Proceedings of the IEEE International Conference on Robotics and Automation (ICRA)},
  year={2020}
}

@inproceedings{kolve2017ai2thor,
  title={AI2-THOR: An interactive 3D environment for visual AI},
  author={Kolve, Eric and Mottaghi, Roozbeh and Han, Winson and VanderBilt, Eli and Weihs, Luca and Herrmann, Charles and Gordon, Daniel and Zhu, Yuke and Gupta, Abhinav and Farhadi, Ali},
  booktitle={arXiv preprint arXiv:1712.05474},
  year={2017}
}

@inproceedings{dosovitskiy2017carla,
  title={CARLA: An open urban driving simulator},
  author={Dosovitskiy, Alexey and Ros, German and Codevilla, Felipe and Lopez, Antonio and Koltun, Vladlen},
  booktitle={Proceedings of the 1st Annual Conference on Robot Learning (CoRL)},
  pages={1--16},
  year={2017}
}

@inproceedings{chi2020just,
  title={Just ask: An interactive learning framework for vision and language navigation},
  author={Chi, Ta-Chung and Shen, Minmin and Eric, Mihail and Kim, Seokhwan and Hakkani-Tur, Dilek},
  booktitle={Proceedings of the AAAI conference on artificial intelligence},
  volume={34},
  number={03},
  pages={2459--2466},
  year={2020}
}

@article{de2018talk,
  title={Talk the walk: Navigating new york city through grounded dialogue},
  author={De Vries, Harm and Shuster, Kurt and Batra, Dhruv and Parikh, Devi and Weston, Jason and Kiela, Douwe},
  journal={arXiv preprint arXiv:1807.03367},
  year={2018}
}

@inproceedings{banerjee2021robotslang,
  title={The RobotSlang benchmark: Dialog-guided robot localization and navigation},
  author={Banerjee, Shurjo and Thomason, Jesse and Corso, Jason},
  booktitle={Conference on Robot Learning},
  pages={1384--1393},
  year={2021},
  organization={PMLR}
}

@inproceedings{fan2023aerial,
  title={Aerial Vision-and-Dialog Navigation},
  author={Fan, Yue and Chen, Winson and Jiang, Tongzhou and Zhou, Chun and Zhang, Yi and Wang, Xin},
  booktitle={Findings of the Association for Computational Linguistics: ACL 2023},
  pages={3043--3061},
  year={2023}
}

@article{wei2025streamvln,
  title={Streamvln: Streaming vision-and-language navigation via slowfast context modeling},
  author={Wei, Meng and Wan, Chenyang and Yu, Xiqian and Wang, Tai and Yang, Yuqiang and Mao, Xiaohan and Zhu, Chenming and Cai, Wenzhe and Wang, Hanqing and Chen, Yilun and others},
  journal={arXiv preprint arXiv:2507.05240},
  year={2025}
}

@inproceedings{ma-etal-2022-dorothie,
    title = "{DOROTHIE}: Spoken Dialogue for Handling Unexpected Situations in Interactive Autonomous Driving Agents",
    author = "Ma, Ziqiao  and
      VanDerPloeg, Benjamin  and
      Bara, Cristian-Paul  and
      Huang, Yidong  and
      Kim, Eui-In  and
      Gervits, Felix  and
      Marge, Matthew  and
      Chai, Joyce",
    booktitle = "Findings of the Association for Computational Linguistics: EMNLP 2022",
    month = dec,
    year = "2022",
    address = "Abu Dhabi, United Arab Emirates",
    publisher = "Association for Computational Linguistics",
    url = "https://aclanthology.org/2022.findings-emnlp.354",
    pages = "4800--4822",
}

@inproceedings{Hermann2019LearningTF,
  title        = {Learning to Follow Directions in Street View},
  author       = {Karl Moritz Hermann and Mateusz Malinowski and Piotr Wojciech Mirowski and Andras Banki-Horvath and Keith Anderson and Raia Hadsell},
  booktitle    = {Proceedings of the AAAI Conference on Artificial Intelligence},
  year         = {2020},
  volume       = {34},
  number       = {07},
  pages        = {11773--11780},
  url          = {https://ojs.aaai.org/index.php/AAAI/article/view/6849},
  doi          = {10.1609/aaai.v34i07.6849}
}

@inproceedings{song2025towards,
  title={Towards long-horizon vision-language navigation: Platform, benchmark and method},
  author={Song, Xinshuai and Chen, Weixing and Liu, Yang and Chen, Weikai and Li, Guanbin and Lin, Liang},
  booktitle={Proceedings of the Computer Vision and Pattern Recognition Conference},
  pages={12078--12088},
  year={2025}
}

@inproceedings{khanna2024goat,
  title={Goat-bench: A benchmark for multi-modal lifelong navigation},
  author={Khanna, Mukul and Ramrakhya, Ram and Chhablani, Gunjan and Yenamandra, Sriram and Gervet, Theophile and Chang, Matthew and Kira, Zsolt and Chaplot, Devendra Singh and Batra, Dhruv and Mottaghi, Roozbeh},
  booktitle={Proceedings of the IEEE/CVF Conference on Computer Vision and Pattern Recognition},
  pages={16373--16383},
  year={2024}
}

@inproceedings{he2024mee,
  author = {He, Zongtao and Wang, Liuyi and Chen, Lu and Li, Shu and Yan, Qingqing and Liu, Chengju and Chen, Qijun},
  booktitle = {2024 IEEE/RSJ International Conference on Intelligent Robots and Systems (IROS)},
  title = {Multimodal Evolutionary Encoder for Continuous Vision-Language Navigation},
  year = {2024},
  volume = {},
  number = {},
  pages = {1443-1450},
  doi = {10.1109/IROS58592.2024.10802484},
  issn = {2153-0866},
  month = oct,
}

@article{anderson2018evaluation,
  title={On evaluation of embodied navigation agents},
  author={Anderson, Peter and Chang, Angel and Chaplot, Devendra Singh and Dosovitskiy, Alexey and Gupta, Saurabh and Koltun, Vladlen and Kosecka, Jana and Malik, Jitendra and Mottaghi, Roozbeh and Savva, Manolis and others},
  journal={arXiv preprint arXiv:1807.06757},
  year={2018}
}

@article{hochreiter1997long,
  title={Long short-term memory},
  author={Hochreiter, Sepp and Schmidhuber, J{\"u}rgen},
  journal={Neural computation},
  volume={9},
  number={8},
  pages={1735--1780},
  year={1997},
  publisher={MIT press}
}

@inproceedings{wang2024enhanced,
  title = {Enhanced Language-guided Robot Navigation with Panoramic Semantic Depth Perception and Cross-modal Fusion},
  author = {Wang, Liuyi and Tang, Jiagui and He, Zongtao and Dang, Ronghao and Liu, Chengju and Chen, Qijun},
  booktitle = {2024 IEEE/RSJ International Conference on Intelligent Robots and Systems (IROS)},
  pages = {7726--7733},
  year = {2024},
  doi = {10.1109/IROS58592.2024.10801563},
  organization = {IEEE},
}

@misc{internvla-n1,
  title = {InternVLA-N1: An Open Dual-System Vision-Language Navigation Foundation Model with Learned Latent Plans},
  author = {InternVLA-N1 Team},
  year = {2025},
  howpublished = {Technical report, Shanghai AI Laboratory},
  url = {https://internrobotics.github.io/internvla-n1.github.io/static/pdfs/InternVLA_N1.pdf}
}

@inproceedings{wangbootstrapping,
  title={Bootstrapping Language-Guided Navigation Learning with Self-Refining Data Flywheel},
  author={Wang, Zun and Li, Jialu and Hong, Yicong and Li, Songze and Li, Kunchang and Yu, Shoubin and Wang, Yi and Qiao, Yu and Wang, Yali and Bansal, Mohit and others},
  booktitle={The Thirteenth International Conference on Learning Representations},
  year={2024}
}

@inproceedings{zhang2025embodied,
  title={Embodied navigation foundation model},
  author={Zhang, Jiazhao and Li, Anqi and Qi, Yunpeng and Li, Minghan and Liu, Jiahang and Wang, Shaoan and Liu, Haoran and Zhou, Gengze and Wu, Yuze and Li, Xingxing and others},
  booktitle = {Proceedings of the International Conference on Learning Representations (ICLR)},
  year={2026}
}

@inproceedings{shen2021igibson,
      title={iGibson 1.0: a Simulation Environment for Interactive Tasks in Large Realistic Scenes}, 
      author={Bokui Shen and Fei Xia and Chengshu Li and Roberto Mart\'in-Mart\'in and Linxi Fan and Guanzhi Wang and Claudia P\'erez-D'Arpino and Shyamal Buch and Sanjana Srivastava and Lyne P. Tchapmi and Micael E. Tchapmi and Kent Vainio and Josiah Wong and Li Fei-Fei and Silvio Savarese},
      booktitle={2021 IEEE/RSJ International Conference on Intelligent Robots and Systems (IROS)},
      year={2021},
      pages={accepted},
      organization={IEEE}
}

@inproceedings{li2022igibson,
  title = 	 {iGibson 2.0: Object-Centric Simulation for Robot Learning of Everyday Household Tasks},
  author =       {Li, Chengshu and Xia, Fei and Mart\'in-Mart\'in, Roberto and Lingelbach, Michael and Srivastava, Sanjana and Shen, Bokui and Vainio, Kent Elliott and Gokmen, Cem and Dharan, Gokul and Jain, Tanish and Kurenkov, Andrey and Liu, Karen and Gweon, Hyowon and Wu, Jiajun and Fei-Fei, Li and Savarese, Silvio},
  booktitle = 	 {Proceedings of the 5th Conference on Robot Learning},
  pages = 	 {455--465},
  year = 	 {2022},
  editor = 	 {Faust, Aleksandra and Hsu, David and Neumann, Gerhard},
  volume = 	 {164},
  series = 	 {Proceedings of Machine Learning Research},
  month = 	 {08--11 Nov},
  publisher =    {PMLR},
}

@inproceedings{sima2024drivelm,
  title={Drivelm: Driving with graph visual question answering},
  author={Sima, Chonghao and Renz, Katrin and Chitta, Kashyap and Chen, Li and Zhang, Hanxue and Xie, Chengen and Bei{\ss}wenger, Jens and Luo, Ping and Geiger, Andreas and Li, Hongyang},
  booktitle={European conference on computer vision},
  pages={256--274},
  year={2024},
  organization={Springer}
}

@inproceedings{li2024think2drive,
  title={Think2drive: Efficient reinforcement learning by thinking with latent world model for autonomous driving (in carla-v2)},
  author={Li, Qifeng and Jia, Xiaosong and Wang, Shaobo and Yan, Junchi},
  booktitle={European Conference on Computer Vision},
  pages={142--158},
  year={2024},
  organization={Springer}
}

@article{vasudevan2021talk2nav,
  title={Talk2nav: Long-range vision-and-language navigation with dual attention and spatial memory},
  author={Vasudevan, Arun Balajee and Dai, Dengxin and Van Gool, Luc},
  journal={International Journal of Computer Vision},
  volume={129},
  number={1},
  pages={246--266},
  year={2021},
  publisher={Springer}
}

@inproceedings{misra2018mapping,
  title={Mapping Instructions to Actions in 3D Environments with Visual Goal Prediction},
  author={Misra, Dipendra and Bennett, Andrew and Blukis, Valts and Niklasson, Eyvind and Shatkhin, Max and Artzi, Yoav},
  booktitle={Proceedings of the 2018 Conference on Empirical Methods in Natural Language Processing},
  pages={2667--2678},
  year={2018}
}

@inproceedings{padmakumar2022teach,
  title={Teach: Task-driven embodied agents that chat},
  author={Padmakumar, Aishwarya and Thomason, Jesse and Shrivastava, Ayush and Lange, Patrick and Narayan-Chen, Anjali and Gella, Spandana and Piramuthu, Robinson and Tur, Gokhan and Hakkani-Tur, Dilek},
  booktitle={Proceedings of the AAAI Conference on Artificial Intelligence},
  volume={36},
  number={2},
  pages={2017--2025},
  year={2022}
}

@article{he2021landmark,
  title={Landmark-rxr: Solving vision-and-language navigation with fine-grained alignment supervision},
  author={He, Keji and Huang, Yan and Wu, Qi and Yang, Jianhua and An, Dong and Sima, Shuanglin and Wang, Liang},
  journal={Advances in Neural Information Processing Systems},
  volume={34},
  pages={652--663},
  year={2021}
}

@article{wu2018building,
  title={Building generalizable agents with a realistic and rich 3d environment},
  author={Wu, Yi and Wu, Yuxin and Gkioxari, Georgia and Tian, Yuandong},
  journal={arXiv preprint arXiv:1801.02209},
  year={2018}
}

@article{yan2019cross,
  title={Cross-lingual vision-language navigation},
  author={Yan, An and Wang, Xin Eric and Feng, Jiangtao and Li, Lei and Wang, William Yang},
  journal={arXiv preprint arXiv:1910.11301},
  year={2019}
}

@inproceedings{nguyen2019vision,
  title={Vision-based navigation with language-based assistance via imitation learning with indirect intervention},
  author={Nguyen, Khanh and Dey, Debadeepta and Brockett, Chris and Dolan, Bill},
  booktitle={Proceedings of the IEEE/CVF Conference on Computer Vision and Pattern Recognition},
  pages={12527--12537},
  year={2019}
}

@inproceedings{suhr2019executing,
  title={Executing instructions in situated collaborative interactions},
  author={Suhr, Alane and Yan, Claudia and Schluger, Jack and Yu, Stanley and Khader, Hadi and Mouallem, Marwa and Zhang, Iris and Artzi, Yoav},
  booktitle={Proceedings of the 2019 Conference on Empirical Methods in Natural Language Processing and the 9th International Joint Conference on Natural Language Processing (EMNLP-IJCNLP)},
  pages={2119--2130},
  year={2019}
}

@article{gao2022dialfred,
  title={Dialfred: Dialogue-enabled agents for embodied instruction following},
  author={Gao, Xiaofeng and Gao, Qiaozi and Gong, Ran and Lin, Kaixiang and Thattai, Govind and Sukhatme, Gaurav S},
  journal={IEEE Robotics and Automation Letters},
  volume={7},
  number={4},
  pages={10049--10056},
  year={2022},
  publisher={IEEE}
}

@inproceedings{roh2020conditional,
  title={Conditional driving from natural language instructions},
  author={Roh, Junha and Paxton, Chris and Pronobis, Andrzej and Farhadi, Ali and Fox, Dieter},
  booktitle={Conference on Robot Learning},
  pages={540--551},
  year={2020},
  organization={PMLR}
}

@inproceedings{sriram2019talk,
  title={Talk to the vehicle: Language conditioned autonomous navigation of self driving cars},
  author={Sriram, NN and Maniar, Tirth and Kalyanasundaram, Jayaganesh and Gandhi, Vineet and Bhowmick, Brojeshwar and Krishna, K Madhava},
  booktitle={2019 IEEE/RSJ international conference on intelligent robots and systems (IROS)},
  pages={5284--5290},
  year={2019},
  organization={IEEE}
}

@article{mirowski2019streetlearn,
  title={The streetlearn environment and dataset},
  author={Mirowski, Piotr and Banki-Horvath, Andras and Anderson, Keith and Teplyashin, Denis and Hermann, Karl Moritz and Malinowski, Mateusz and Grimes, Matthew Koichi and Simonyan, Karen and Kavukcuoglu, Koray and Zisserman, Andrew and others},
  journal={arXiv preprint arXiv:1903.01292},
  year={2019}
}

@inproceedings{narayan2019collaborative,
  title={Collaborative dialogue in Minecraft},
  author={Narayan-Chen, Anjali and Jayannavar, Prashant and Hockenmaier, Julia},
  booktitle={Proceedings of the 57th Annual Meeting of the Association for Computational Linguistics},
  pages={5405--5415},
  year={2019}
}

@inproceedings{das2018embodied,
  title={Embodied question answering},
  author={Das, Abhishek and Datta, Samyak and Gkioxari, Georgia and Lee, Stefan and Parikh, Devi and Batra, Dhruv},
  booktitle={Proceedings of the IEEE conference on computer vision and pattern recognition},
  pages={1--10},
  year={2018}
}

@inproceedings{gordon2018iqa,
  title={Iqa: Visual question answering in interactive environments},
  author={Gordon, Daniel and Kembhavi, Aniruddha and Rastegari, Mohammad and Redmon, Joseph and Fox, Dieter and Farhadi, Ali},
  booktitle={Proceedings of the IEEE conference on computer vision and pattern recognition},
  pages={4089--4098},
  year={2018}
}

@ARTICLE{navlab_visnav,
  author={Thorpe, C. and Hebert, M.H. and Kanade, T. and Shafer, S.A.},
  journal={IEEE Transactions on Pattern Analysis and Machine Intelligence}, 
  title={Vision and navigation for the Carnegie-Mellon Navlab}, 
  year={1988},
  volume={10},
  number={3},
  pages={362-373},
  doi={10.1109/34.3900}}

@ARTICLE{Waxman_visnav_alv,
  author={Waxman, A. and LeMoigne, J. and Davis, L. and Srinivasan, B. and Kushner, T. and Eli Liang and Siddalingaiah, T.},
  journal={IEEE Journal on Robotics and Automation}, 
  title={A visual navigation system for autonomous land vehicles}, 
  year={1987},
  volume={3},
  number={2},
  pages={124-141},
  doi={10.1109/JRA.1987.1087089}}

@Inbook{Smith1990,
author="Smith, Randall
and Self, Matthew
and Cheeseman, Peter",
editor="Cox, Ingemar J.
and Wilfong, Gordon T.",
title="Estimating Uncertain Spatial Relationships in Robotics",
bookTitle="Autonomous Robot Vehicles",
year="1990",
publisher="Springer New York",
address="New York, NY",
pages="167--193",
isbn="978-1-4613-8997-2",
doi="10.1007/978-1-4613-8997-2_14",
url="https://doi.org/10.1007/978-1-4613-8997-2_14"
}

@article{Smith1986,
author = {Randall C. Smith and Peter Cheeseman},
title ={On the Representation and Estimation of Spatial Uncertainty},
journal = {The International Journal of Robotics Research},
volume = {5},
number = {4},
pages = {56-68},
year = {1986},
doi = {10.1177/027836498600500404},
URL = {https://doi.org/10.1177/027836498600500404}
}

@inproceedings{moutarlier1989experimental,
  title={An Experimental System for Incremental Environment Modelling by an Autonomous Mobile Robot},
  author={Moutarlier, Philippe and Chatila, Raja},
  booktitle={The First International Symposium on Experimental Robotics I},
  pages={327--346},
  year={1989}
}

@INPROCEEDINGS{Krotkov1989,
  author={Krotkov, E.},
  booktitle={Proceedings, 1989 International Conference on Robotics and Automation}, 
  title={Mobile robot localization using a single image}, 
  year={1989},
  volume={},
  number={},
  pages={978-983 vol.2},
  doi={10.1109/ROBOT.1989.100108}}

@INPROCEEDINGS{1315094,
  author={Nister, D. and Naroditsky, O. and Bergen, J.},
  booktitle={Proceedings of the 2004 IEEE Computer Society Conference on Computer Vision and Pattern Recognition, 2004. CVPR 2004.}, 
  title={Visual odometry}, 
  year={2004},
  volume={1},
  number={},
  pages={I-I},
  doi={10.1109/CVPR.2004.1315094}}

@ARTICLE{4160954,
  author={Davison, Andrew J. and Reid, Ian D. and Molton, Nicholas D. and Stasse, Olivier},
  journal={IEEE Transactions on Pattern Analysis and Machine Intelligence}, 
  title={MonoSLAM: Real-Time Single Camera SLAM}, 
  year={2007},
  volume={29},
  number={6},
  pages={1052-1067},
  doi={10.1109/TPAMI.2007.1049}}

@INPROCEEDINGS{4538852,
  author={Klein, Georg and Murray, David},
  booktitle={2007 6th IEEE and ACM International Symposium on Mixed and Augmented Reality}, 
  title={Parallel Tracking and Mapping for Small AR Workspaces}, 
  year={2007},
  volume={},
  number={},
  pages={225-234},
  doi={10.1109/ISMAR.2007.4538852}}

@InProceedings{lsd_slam,
    author="Engel, Jakob
    and Sch{\"o}ps, Thomas
    and Cremers, Daniel",
    editor="Fleet, David
    and Pajdla, Tomas
    and Schiele, Bernt
    and Tuytelaars, Tinne",
    title="LSD-SLAM: Large-Scale Direct Monocular SLAM",
    booktitle="Computer Vision -- ECCV 2014",
    year="2014",
    publisher="Springer International Publishing",
    address="Cham",
    pages="834--849",
    isbn="978-3-319-10605-2"
}

@ARTICLE{orb_slam2,
  author={Mur-Artal, Raúl and Tardós, Juan D.},
  journal={IEEE Transactions on Robotics}, 
  title={ORB-SLAM2: An Open-Source SLAM System for Monocular, Stereo, and RGB-D Cameras}, 
  year={2017},
  volume={33},
  number={5},
  pages={1255-1262},
  doi={10.1109/TRO.2017.2705103}}

@INPROCEEDINGS{deepvo,
  author={Wang, Sen and Clark, Ronald and Wen, Hongkai and Trigoni, Niki},
  booktitle={2017 IEEE International Conference on Robotics and Automation (ICRA)}, 
  title={DeepVO: Towards end-to-end visual odometry with deep Recurrent Convolutional Neural Networks}, 
  year={2017},
  volume={},
  number={},
  pages={2043-2050},
  doi={10.1109/ICRA.2017.7989236}}

@inproceedings{droid_slam,
 author = {Teed, Zachary and Deng, Jia},
 booktitle = {Advances in Neural Information Processing Systems},
 editor = {M. Ranzato and A. Beygelzimer and Y. Dauphin and P.S. Liang and J. Wortman Vaughan},
 pages = {16558--16569},
 publisher = {Curran Associates, Inc.},
 title = {DROID-SLAM: Deep Visual SLAM for Monocular, Stereo, and RGB-D Cameras},
 volume = {34},
 year = {2021}
}

@article{wang2024grutopia,
  title={Grutopia: Dream general robots in a city at scale},
  author={Wang, Hanqing and Chen, Jiahe and Huang, Wensi and Ben, Qingwei and Wang, Tai and Mi, Boyu and Huang, Tao and Zhao, Siheng and Chen, Yilun and Yang, Sizhe and others},
  journal={arXiv preprint arXiv:2407.10943},
  year={2024}
}

@inproceedings{lin2022adapt,
  title={ADAPT: Vision-Language Navigation with Modality-Aligned Action Prompts},
  author={Lin, Bingqian and Zhu, Yi and Chen, Zicong and Liang, Xiwen and Liu, Jianzhuang and Liang, Xiaodan},
  booktitle={Proceedings of the IEEE/CVF Conference on Computer Vision and Pattern Recognition},
  pages={15396--15406},
  year={2022}
}

@inproceedings{radford2021learning,
  title={Learning transferable visual models from natural language supervision},
  author={Radford, Alec and Kim, Jong Wook and Hallacy, Chris and Ramesh, Aditya and Goh, Gabriel and Agarwal, Sandhini and Sastry, Girish and Askell, Amanda and Mishkin, Pamela and Clark, Jack and others},
  booktitle={International conference on machine learning},
  pages={8748--8763},
  year={2021},
  organization={PmLR}
}

@inproceedings{fang2025hierarchical,
  title={Hierarchical semantic-augmented navigation: Optimal transport and graph-driven reasoning for vision-language navigation},
  author={Fang, Xiang and Fang, Wanlong and Wang, Changshuo},
  booktitle={The Thirty-ninth Annual Conference on Neural Information Processing Systems},
  year={2025}
}

@article{yue2025think,
  title={Think Hierarchically, Act Dynamically: Hierarchical Multi-modal Fusion and Reasoning for Vision-and-Language Navigation},
  author={Yue, Junrong and Zhang, Yifan and Qin, Chuan and Li, Bo and Lie, Xiaomin and Yu, Xinlei and Zhang, Wenxin and Zhao, Zhendong},
  journal={arXiv preprint arXiv:2504.16516},
  year={2025}
}

@inproceedings{zhang2025citynavagent,
  title     = {CityNavAgent: Aerial Vision-and-Language Navigation with Hierarchical Semantic Planning and Global Memory},
  author    = {Zhang, Weichen and Gao, Chen and Yu, Shiquan and Peng, Ruiying and Zhao, Baining and Zhang, Qian and Cui, Jinqiang and Chen, Xinlei and Li, Yong},
  booktitle = {Proceedings of the 63rd Annual Meeting of the Association for Computational Linguistics (ACL)},
  year      = {2025}
}

@article{wang2025clash,
  title={CLASH: Collaborative Large-Small Hierarchical Framework for Continuous Vision-and-Language Navigation},
  author={Wang, Liuyi and He, Zongtao and Li, Jinlong and Qi, Xiaoyan and Hu, Mengxian and Yao, Chenpeng and Liu, Chengju and Chen, Qijun},
  journal={arXiv preprint arXiv:2512.10360},
  year={2025}
}

@article{shi2025smartway,
  title={SmartWay: Enhanced Waypoint Prediction and Backtracking for Zero-Shot Vision-and-Language Navigation},
  author={Shi, Xiangyu and Li, Zerui and Lyu, Wenqi and Xia, Jiatong and Dayoub, Feras and Qiao, Yanyuan and Wu, Qi},
  journal={CoRR},
  year={2025}
}

@article{wen2025ovl,
  title={OVL-MAP: An Online Visual Language Map Approach for Vision-and-Language Navigation in Continuous Environments},
  author={Wen, Shuhuan and Zhang, Ziyuan and Sun, Yuxiang and Wang, Zhiwen},
  journal={IEEE Robotics and Automation Letters},
  year={2025},
  publisher={IEEE}
}

@inproceedings{wang2025dynam3d,
  title={Dynam3D: Dynamic Layered 3D Tokens Empower VLM for Vision-and-Language Navigation},
  author={Wang, Zihan and Lee, Seungjun and Lee, Gim Hee},
  booktitle={Advances in Neural Information Processing Systems},
  year={2025}
}

@inproceedings{hernandez2019agent,
  title={Agent modeling as auxiliary task for deep reinforcement learning},
  author={Hernandez-Leal, Pablo and Kartal, Bilal and Taylor, Matthew E},
  booktitle={Proceedings of the AAAI conference on artificial intelligence and interactive digital entertainment},
  volume={15},
  number={1},
  pages={31--37},
  year={2019}
}

@article{lin2019adaptive,
  title={Adaptive auxiliary task weighting for reinforcement learning},
  author={Lin, Xingyu and Baweja, Harjatin and Kantor, George and Held, David},
  journal={Advances in neural information processing systems},
  volume={32},
  year={2019}
}

@inproceedings{wang2020active,
  title={Active visual information gathering for vision-language navigation},
  author={Wang, Hanqing and Wang, Wenguan and Shu, Tianmin and Liang, Wei and Shen, Jianbing},
  booktitle={European conference on computer vision},
  pages={307--322},
  year={2020},
  organization={Springer}
}

@inproceedings{kuo2023structure,
  title={Structure-encoding auxiliary tasks for improved visual representation in vision-and-language navigation},
  author={Kuo, Chia-Wen and Ma, Chih-Yao and Hoffman, Judy and Kira, Zsolt},
  booktitle={Proceedings of the IEEE/CVF Winter Conference on Applications of Computer Vision},
  pages={1104--1113},
  year={2023}
}

@inproceedings{wang2020environment,
  title={Environment-agnostic multitask learning for natural language grounded navigation},
  author={Wang, Xin Eric and Jain, Vihan and Ie, Eugene and Wang, William Yang and Kozareva, Zornitsa and Ravi, Sujith},
  booktitle={European conference on computer vision},
  pages={413--430},
  year={2020},
  organization={Springer}
}

@inproceedings{dorbala2022clip,
  title={CLIP-Nav: Using CLIP for Zero-Shot Vision-and-Language Navigation},
  author={Dorbala, Vishnu Sashank and Sigurdsson, Gunnar A and Thomason, Jesse and Piramuthu, Robinson and Sukhatme, Gaurav S},
  booktitle={Workshop on Language and Robotics at CoRL 2022}
}

@article{wang2024magic,
  title={Magic: Meta-ability guided interactive chain-of-distillation for effective-and-efficient vision-and-language navigation},
  author={Wang, Liuyi and He, Zongtao and Shen, Mengjiao and Yang, Jingwei and Liu, Chengju and Chen, Qijun},
  journal={IEEE Transactions on Pattern Analysis and Machine Intelligence},
  year={2026},
  publisher={IEEE}
}

@article{lin2025navcot,
  title={Navcot: Boosting llm-based vision-and-language navigation via learning disentangled reasoning},
  author={Lin, Bingqian and Nie, Yunshuang and Wei, Ziming and Chen, Jiaqi and Ma, Shikui and Han, Jianhua and Xu, Hang and Chang, Xiaojun and Liang, Xiaodan},
  journal={IEEE Transactions on Pattern Analysis and Machine Intelligence},
  year={2025},
  publisher={IEEE}
}

@inproceedings{qiao2025open,
  title={Open-nav: Exploring zero-shot vision-and-language navigation in continuous environment with open-source llms},
  author={Qiao, Yanyuan and Lyu, Wenqi and Wang, Hui and Wang, Zixu and Li, Zerui and Zhang, Yuan and Tan, Mingkui and Wu, Qi},
  booktitle={2025 IEEE International Conference on Robotics and Automation (ICRA)},
  pages={6710--6717},
  year={2025},
  organization={IEEE}
}

@inproceedings{xue2025omninav,
  title={Omninav: A unified framework for prospective exploration and visual-language navigation},
  author={Xue, Xinda and Hu, Junjun and Luo, Minghua and Shichao, Xie and Chen, Jintao and Xie, Zixun and Kuichen, Quan and Wei, Guo and Xu, Mu and Chu, Zedong},
  booktitle = {Proceedings of the International Conference on Learning Representations (ICLR)},
  year={2026}
}

@inproceedings{yang2024llm,
  title={Llm-grounder: Open-vocabulary 3d visual grounding with large language model as an agent},
  author={Yang, Jianing and Chen, Xuweiyi and Qian, Shengyi and Madaan, Nikhil and Iyengar, Madhavan and Fouhey, David F and Chai, Joyce},
  booktitle={2024 IEEE International Conference on Robotics and Automation (ICRA)},
  pages={7694--7701},
  year={2024},
  organization={IEEE}
}

@inproceedings{chen2025affordances,
  title={Affordances-oriented planning using foundation models for continuous vision-language navigation},
  author={Chen, Jiaqi and Lin, Bingqian and Liu, Xinmin and Ma, Lin and Liang, Xiaodan and Wong, Kwan-Yee K},
  booktitle={Proceedings of the AAAI Conference on Artificial Intelligence},
  volume={39},
  number={22},
  pages={23568--23576},
  year={2025}
}

@inproceedings{liu2025cvln,
  title={CVLN-Think: Causal Inference with Counterfactual Style Adaptation for Continuous Vision-and-Language Navigation},
  author={Liu, Ruonan and Wu, Shuai and Lin, Di and Zhang, Weidong},
  booktitle={2025 IEEE/RSJ International Conference on Intelligent Robots and Systems (IROS)},
  pages={15299--15305},
  year={2025},
  organization={IEEE}
}

@article{shi2025fastsmartWay,
  title={Fast-SmartWay: Panoramic-Free End-to-End Zero-Shot Vision-and-Language Navigation},
  author={Shi, Xiangyu and Li, Zerui and Qiao, Yanyuan and Wu, Qi},
  journal={arXiv preprint arXiv:2511.00933},
  year={2025}
}

@inproceedings{zeng2025janusvln,
  title     = {JanusVLN: Decoupling Semantics and Spatiality with Dual Implicit Memory for Vision-Language Navigation},
  author    = {Zeng, Shuang and Qi, Dekang and Chang, Xinyuan and Xiong, Feng and Xie, Shichao and Wu, Xiaolong and Liang, Shiyi and Xu, Mu and Wei, Xing},
  booktitle = {Proceedings of the International Conference on Learning Representations (ICLR)},
  year      = {2026}
}

@inproceedings{wang2025think,
  title     = {Aux-Think: Exploring Reasoning Strategies for Data-Efficient Vision-Language Navigation},
  author    = {Wang, Shuo and Wang, Yongcai and Li, Wanting and Cai, Xudong and Wang, Yucheng and Chen, Maiyue and Wang, Kaihui and Su, Zhizhong and Li, Deying and Fan, Zhaoxin},
  booktitle = {Advances in Neural Information Processing Systems},
  year      = {2025}
}

@INPROCEEDINGS{navid4d,
  author={Liu, Haoran and Wan, Weikang and Yu, Xiqian and Li, Minghan and Zhang, Jiazhao and Zhao, Bo and Chen, Zhibo and Wang, Zhongyuan and Zhang, Zhizheng and Wang, He},
  booktitle={2025 IEEE International Conference on Robotics and Automation (ICRA)}, 
  title={Na Vid-4D: Unleashing Spatial Intelligence in Egocentric RGB-D Videos for Vision-and-Language Navigation}, 
  year={2025},
  volume={},
  number={},
  pages={10607-10615},
  doi={10.1109/ICRA55743.2025.11128467}}

@inproceedings{mapnav,
    title = "{M}ap{N}av: A Novel Memory Representation via Annotated Semantic Maps for {VLM}-based Vision-and-Language Navigation",
    author = "Zhang, Lingfeng  and
      Hao, Xiaoshuai  and
      Xu, Qinwen  and
      Zhang, Qiang  and
      Zhang, Xinyao  and
      Wang, Pengwei  and
      Zhang, Jing  and
      Wang, Zhongyuan  and
      Zhang, Shanghang  and
      Xu, Renjing",
    editor = "Che, Wanxiang  and
      Nabende, Joyce  and
      Shutova, Ekaterina  and
      Pilehvar, Mohammad Taher",
    booktitle = "Proceedings of the 63rd Annual Meeting of the Association for Computational Linguistics (Volume 1: Long Papers)",
    month = jul,
    year = "2025",
    address = "Vienna, Austria",
    publisher = "Association for Computational Linguistics",
    doi = "10.18653/v1/2025.acl-long.638",
    pages = "13032--13056",
    ISBN = "979-8-89176-251-0"
}

@article{zhang2025activevln,
  title={Activevln: Towards active exploration via multi-turn rl in vision-and-language navigation},
  author={Zhang, Zekai and Zhu, Weiye and Pan, Hewei and Wang, Xiangchen and Xu, Rongtao and Sun, Xing and Zheng, Feng},
  journal={arXiv preprint arXiv:2509.12618},
  year={2025}
}

@article{qi2025vln,
  title={VLN-R1: Vision-Language Navigation via Reinforcement Fine-Tuning},
  author={Qi, Zhangyang and Zhang, Zhixiong and Yu, Yizhou and Wang, Jiaqi and Zhao, Hengshuang},
  journal={arXiv preprint arXiv:2506.17221},
  year={2025}
}

@article{glossop2025cast,
  title={Cast: Counterfactual labels improve instruction following in vision-language-action models},
  author={Glossop, Catherine and Chen, William and Bhorkar, Arjun and Shah, Dhruv and Levine, Sergey},
  journal={arXiv preprint arXiv:2508.13446},
  year={2025}
}

@article{ding2025adanav,
  title={AdaNav: Adaptive Reasoning with Uncertainty for Vision-Language Navigation},
  author={Ding, Xin and Wei, Jianyu and Yang, Yifan and Jiang, Shiqi and Zhang, Qianxi and Wu, Hao and Jia, Fucheng and Mi, Liang and Yan, Yuxuan and Wang, Weijun and others},
  journal={arXiv preprint arXiv:2509.24387},
  year={2025}
}

@inproceedings{yin2025gc,
  title={GC-VLN: Instruction as Graph Constraints for Training-free Vision-and-Language Navigation},
  author={Yin, Hang and Wei, Haoyu and Xu, Xiuwei and Guo, Wenxuan and Zhou, Jie and Lu, Jiwen},
  booktitle={Conference on Robot Learning},
  pages={1809--1824},
  year={2025},
  organization={PMLR}
}

@article{bhatt2025vln,
  title={Vln-zero: Rapid exploration and cache-enabled neurosymbolic vision-language planning for zero-shot transfer in robot navigation},
  author={Bhatt, Neel P and Yang, Yunhao and Siva, Rohan and Samineni, Pranay and Milan, Daniel and Wang, Zhangyang and Topcu, Ufuk},
  journal={arXiv preprint arXiv:2509.18592},
  year={2025}
}

@inproceedings{kaasene2025following,
  title={Following Route Instructions using Large Vision-Language Models: A Comparison between Low-level and Panoramic Action Spaces},
  author={K{\aa}sene, Vebj{\o}rn and Lison, Pierre},
  booktitle={Proceedings of the 8th International Conference on Natural Language and Speech Processing (ICNLSP-2025)},
  pages={449--463},
  year={2025}
}

@article{cai2025cl,
  title={CL-CoTNav: Closed-Loop Hierarchical Chain-of-Thought for Zero-Shot Object-Goal Navigation with Vision-Language Models},
  author={Cai, Yuxin and He, Xiangkun and Wang, Maonan and Guo, Hongliang and Yau, Wei-Yun and Lv, Chen},
  journal={arXiv preprint arXiv:2504.09000},
  year={2025}
}

@article{windecker2025navitrace,
  title={NaviTrace: Evaluating Embodied Navigation of Vision-Language Models},
  author={Windecker, Tim and Patel, Manthan and Reuss, Moritz and Schwarzkopf, Richard and Cadena, Cesar and Lioutikov, Rudolf and Hutter, Marco and Frey, Jonas},
  journal={arXiv preprint arXiv:2510.26909},
  year={2025}
}

@article{hirose2025omnivla,
  title={OmniVLA: An omni-modal vision-language-action model for robot navigation},
  author={Hirose, Noriaki and Glossop, Catherine and Shah, Dhruv and Levine, Sergey},
  journal={arXiv preprint arXiv:2509.19480},
  year={2025}
}

@inproceedings{visitron,
    title = "{VISITRON}: Visual Semantics-Aligned Interactively Trained Object-Navigator",
    author = "Shrivastava, Ayush  and
      Gopalakrishnan, Karthik  and
      Liu, Yang  and
      Piramuthu, Robinson  and
      Tur, Gokhan  and
      Parikh, Devi  and
      Hakkani-Tur, Dilek",
    editor = "Muresan, Smaranda  and
      Nakov, Preslav  and
      Villavicencio, Aline",
    booktitle = "Findings of the Association for Computational Linguistics: ACL 2022",
    month = may,
    year = "2022",
    address = "Dublin, Ireland",
    publisher = "Association for Computational Linguistics",
    url = "https://aclanthology.org/2022.findings-acl.157/",
    doi = "10.18653/v1/2022.findings-acl.157",
    pages = "1984--1994"
}

@article{subedi2025can,
  title={Can Pretrained Vision-Language Embeddings Alone Guide Robot Navigation?},
  author={Subedi, Nitesh and Haroon, Adam and Ganguly, Shreyan and Tetteh, Samuel TK and Koirala, Prajwal and Fleming, Cody and Sarkar, Soumik},
  journal={arXiv preprint arXiv:2506.14507},
  year={2025}
}

@article{shi2025dagger,
  title={DAgger Diffusion Navigation: DAgger Boosted Diffusion Policy for Vision-Language Navigation},
  author={Shi, Haoxiang and Deng, Xiang and Li, Zaijing and Chen, Gongwei and Wang, Yaowei and Nie, Liqiang},
  journal={arXiv preprint arXiv:2508.09444},
  year={2025}
}

@InProceedings{krantz2023iterative,
    author    = {Krantz, Jacob and Banerjee, Shurjo and Zhu, Wang and Corso, Jason and Anderson, Peter and Lee, Stefan and Thomason, Jesse},
    title     = {Iterative Vision-and-Language Navigation},
    booktitle = {Proceedings of the IEEE/CVF Conference on Computer Vision and Pattern Recognition (CVPR)},
    month     = {June},
    year      = {2023},
    pages     = {14921-14930}
}

@InProceedings{wang2025rethinking,
    author    = {Wang, Liuyi and Xia, Xinyuan and Zhao, Hui and Wang, Hanqing and Wang, Tai and Chen, Yilun and Liu, Chengju and Chen, Qijun and Pang, Jiangmiao},
    title     = {Rethinking the Embodied Gap in Vision-and-Language Navigation: A Holistic Study of Physical and Visual Disparities},
    booktitle = {Proceedings of the IEEE/CVF International Conference on Computer Vision (ICCV)},
    month     = {October},
    year      = {2025},
    pages     = {9455-9465}
}

@inproceedings{adang2025singer,
  title={SINGER: An Onboard Generalist Vision-Language Navigation Policy for Drones},
  author={Adang, Maximilian and Low, JunEn and Shorinwa, Olao and Schwager, Mac},
  booktitle={IROS 2025 Workshop: Open World Navigation in Human-centric Environments}
}

@inproceedings{zhu2021multimodal,
  title={Multimodal text style transfer for outdoor vision-and-language navigation},
  author={Zhu, Wanrong and Wang, Xin and Fu, Tsu-Jui and Yan, An and Narayana, Pradyumna and Sone, Kazoo and Basu, Sugato and Wang, William Yang},
  booktitle={Proceedings of the 16th Conference of the European Chapter of the Association for Computational Linguistics: Main Volume},
  pages={1207--1221},
  year={2021}
}

@inproceedings{schumann-riezler-2022-analyzing,
    title = "Analyzing Generalization of Vision and Language Navigation to Unseen Outdoor Areas",
    author = "Schumann, Raphael  and
      Riezler, Stefan",
    editor = "Muresan, Smaranda  and
      Nakov, Preslav  and
      Villavicencio, Aline",
    booktitle = "Proceedings of the 60th Annual Meeting of the Association for Computational Linguistics (Volume 1: Long Papers)",
    month = may,
    year = "2022",
    address = "Dublin, Ireland",
    publisher = "Association for Computational Linguistics",
    doi = "10.18653/v1/2022.acl-long.518",
    pages = "7519--7532"
}

@inproceedings{loc4plan,
author = {Tian, Huilin and Meng, Jingke and Zheng, Wei-Shi and Li, Yuan-Ming and Yan, Junkai and Zhang, Yunong},
title = {Loc4Plan: Locating Before Planning for Outdoor Vision and Language Navigation},
year = {2024},
isbn = {9798400706868},
publisher = {Association for Computing Machinery},
address = {New York, NY, USA},
doi = {10.1145/3664647.3681518},
booktitle = {Proceedings of the 32nd ACM International Conference on Multimedia},
pages = {4073–4081},
numpages = {9},
location = {Melbourne VIC, Australia},
series = {MM '24}
}

@inproceedings{kim-etal-2020-arramon,
    title = "{A}rra{M}on: A Joint Navigation-Assembly Instruction Interpretation Task in Dynamic Environments",
    author = "Kim, Hyounghun  and
      Zala, Abhaysinh  and
      Burri, Graham  and
      Tan, Hao  and
      Bansal, Mohit",
    editor = "Cohn, Trevor  and
      He, Yulan  and
      Liu, Yang",
    booktitle = "Findings of the Association for Computational Linguistics: EMNLP 2020",
    month = nov,
    year = "2020",
    address = "Online",
    publisher = "Association for Computational Linguistics",
    doi = "10.18653/v1/2020.findings-emnlp.348",
    pages = "3910--3927"
}

@InProceedings{liu2023aerialvln,
    author    = {Liu, Shubo and Zhang, Hongsheng and Qi, Yuankai and Wang, Peng and Zhang, Yanning and Wu, Qi},
    title     = {AerialVLN: Vision-and-Language Navigation for UAVs},
    booktitle = {Proceedings of the IEEE/CVF International Conference on Computer Vision (ICCV)},
    month     = {October},
    year      = {2023},
    pages     = {15384-15394}
}

@InProceedings{Lee_2025_ICCV,
    author    = {Lee, Jungdae and Miyanishi, Taiki and Kurita, Shuhei and Sakamoto, Koya and Azuma, Daichi and Matsuo, Yutaka and Inoue, Nakamasa},
    title     = {CityNav: A Large-Scale Dataset for Real-World Aerial Navigation},
    booktitle = {Proceedings of the IEEE/CVF International Conference on Computer Vision (ICCV)},
    month     = {October},
    year      = {2025},
    pages     = {5912-5922}
}

@ARTICLE{Sanyal2025ASMA,
  author={Sanyal, Sourav and Roy, Kaushik},
  journal={IEEE Robotics and Automation Letters}, 
  title={ASMA: An $\underline{\text{A}}$daptive $\underline{\text{S}}$afety $\underline{\text{M}}$argin $\underline{\text{A}}$lgorithm for Vision-Language Drone Navigation via Scene-Aware Control Barrier Functions},
  year={2025},
  volume={10},
  number={9},
  pages={9232-9239},
  doi={10.1109/LRA.2025.3592138}}

@article{chen2025grad,
  title={Grad-nav++: Vision-language model enabled visual drone navigation with gaussian radiance fields and differentiable dynamics},
  author={Chen, Qianzhong and Gao, Naixiang and Huang, Suning and Low, JunEn and Chen, Timothy and Sun, Jiankai and Schwager, Mac},
  journal={IEEE Robotics and Automation Letters},
  volume={11},
  number={2},
  pages={1418--1425},
  year={2025},
  publisher={IEEE}
}

@inproceedings{li2024vlnvideo,
  title={Vln-video: Utilizing driving videos for outdoor vision-and-language navigation},
  author={Li, Jialu and Padmakumar, Aishwarya and Sukhatme, Gaurav and Bansal, Mohit},
  booktitle={Proceedings of the AAAI Conference on Artificial Intelligence},
  volume={38},
  number={17},
  pages={18517--18526},
  year={2024}
}

@article{liu2024navagent,
  title={Navagent: Multi-scale urban street view fusion for uav embodied vision-and-language navigation},
  author={Liu, Youzhi and Yao, Fanglong and Yue, Yuanchang and Xu, Guangluan and Sun, Xian and Fu, Kun},
  journal={arXiv preprint arXiv:2411.08579},
  year={2024}
}

@article{elnoor2024robot,
  title={Robot navigation using physically grounded vision-language models in outdoor environments},
  author={Elnoor, Mohamed and Weerakoon, Kasun and Seneviratne, Gershom and Xian, Ruiqi and Guan, Tianrui and Jaffar, Mohamed Khalid M and Rajagopal, Vignesh and Manocha, Dinesh},
  journal={arXiv preprint arXiv:2409.20445},
  year={2024}
}

@inproceedings{miao2025towards,
  title={Towards Physically Executable 3D Gaussian for Embodied Navigation},
  author={Miao, Bingchen and Wei, Rong and Ge, Zhiqi and Gao, Shiqi and Zhu, Jingzhe and Wang, Renhan and Tang, Siliang and Xiao, Jun and Tang, Rui and Li, Juncheng and others},
  booktitle = {Proceedings of the International Conference on Learning Representations (ICLR)},
  year={2026}
}

@inproceedings{baghaei-etal-2025-follow,
    title = "Follow the Beaten Path: The Role of Route Patterns on Vision-Language Navigation Agents Generalization Abilities",
    author = "Baghaei, Kourosh T  and
      Pfoser, Dieter  and
      Anastasopoulos, Antonios",
    editor = "Chiruzzo, Luis  and
      Ritter, Alan  and
      Wang, Lu",
    booktitle = "Proceedings of the 2025 Conference of the Nations of the Americas Chapter of the Association for Computational Linguistics: Human Language Technologies (Volume 1: Long Papers)",
    month = apr,
    year = "2025",
    address = "Albuquerque, New Mexico",
    publisher = "Association for Computational Linguistics",
    doi = "10.18653/v1/2025.naacl-long.406",
    pages = "7986--8005",
    ISBN = "979-8-89176-189-6"
}

@inproceedings{uavon,
author = {Xiao, Jianqiang and Sun, Yuexuan and Shao, Yixin and Gan, Boxi and Liu, Rongqiang and Wu, Yanjin and Guan, Weili and Deng, Xiang},
title = {UAV-ON: A Benchmark for Open-World Object Goal Navigation with Aerial Agents},
year = {2025},
isbn = {9798400720352},
publisher = {Association for Computing Machinery},
address = {New York, NY, USA},
doi = {10.1145/3746027.3758251},
booktitle = {Proceedings of the 33rd ACM International Conference on Multimedia},
pages = {13023–13029},
numpages = {7},
location = {Dublin, Ireland},
series = {MM '25}
}

@article{gao2024aerial,
  title={Aerial vision-and-language navigation via semantic-topo-metric representation guided LLM reasoning},
  author={Gao, Yunpeng and Wang, Zhigang and Jing, Linglin and Wang, Dong and Li, Xuelong and Zhao, Bin},
  journal={arXiv preprint arXiv:2410.08500},
  year={2024}
}

@article{xu2025aerial,
  title={Aerial Vision-Language Navigation with a Unified Framework for Spatial, Temporal and Embodied Reasoning},
  author={Xu, Huilin and Liu, Zhuoyang and Luomei, Yixiang and Xu, Feng},
  journal={arXiv preprint arXiv:2512.08639},
  year={2025}
}

@article{cai2025sa,
  title={SA-GCS: Semantic-Aware Gaussian Curriculum Scheduling for UAV Vision-Language Navigation},
  author={Cai, Hengxing and Dong, Jinhan and Rao, Yijie and Deng, Jingcheng and Tan, Jingjun and Chen, Qien and Wang, Haidong and Wang, Zhen and Huang, Shiyu and Sumalee, Agachai and others},
  journal={arXiv preprint arXiv:2508.00390},
  year={2025}
}

@article{zhao2025t,
  title={T-araVLN: Translator for Agricultural Robotic Agents on Vision-and-Language Navigation},
  author={Zhao, Xiaobei and Lyu, Xingqi and Li, Xiang},
  journal={arXiv preprint arXiv:2509.06644},
  year={2025}
}

@article{zhao2025sum,
  title={SUM-AgriVLN: Spatial Understanding Memory for Agricultural Vision-and-Language Navigation},
  author={Zhao, Xiaobei and Lyu, Xingqi and Li, Xiang},
  journal={arXiv preprint arXiv:2510.14357},
  year={2025}
}

@article{zhao2025mde,
  title={MDE-AgriVLN: Agricultural Vision-and-Language Navigation with Monocular Depth Estimation},
  author={Zhao, Xiaobei and Lyu, Xingqi and Chen, Xin and Li, Xiang},
  journal={arXiv preprint arXiv:2512.03958},
  year={2025}
}

@article{wang2025underwatervla,
  title={UnderwaterVLA: Dual-brain Vision-Language-Action architecture for Autonomous Underwater Navigation},
  author={Wang, Zhangyuan and Zhu, Yunpeng and Yan, Yuqi and Tian, Xiaoyuan and Shao, Xinhao and Li, Meixuan and Li, Weikun and Su, Guangsheng and Cui, Weicheng and Fan, Dixia},
  journal={arXiv preprint arXiv:2509.22441},
  year={2025}
}

@article{wen2024vision,
  title={Vision-and-language navigation based on history-aware cross-modal feature fusion in indoor environment},
  author={Wen, Shuhuan and Gong, Simeng and Zhang, Ziyuan and Yu, F Richard and Wang, Zhiwen},
  journal={Knowledge-Based Systems},
  volume={305},
  pages={112610},
  year={2024},
  publisher={Elsevier}
}

@inproceedings{liu2024multiple,
  title={Multiple Visual Features in Topological Map for Vision-and-Language Navigation},
  author={Liu, Ruonan and Kong, Ping and Zhang, Weidong},
  booktitle={2024 IEEE/RSJ International Conference on Intelligent Robots and Systems (IROS)},
  pages={7742--7749},
  year={2024},
  organization={IEEE}
}

@inproceedings{zhu2025history,
  title={From History to Goal: Enhanced Vision-and-Language Navigation with Historical Traceability},
  author={Zhu, Xinguang and Wang, Min and Li, Li and Zhou, Wengang and Li, Houqiang},
  booktitle={2025 IEEE International Conference on Multimedia and Expo (ICME)},
  pages={1--6},
  year={2025},
  organization={IEEE}
}

@inproceedings{kong2025vln,
  title={VLN-KHVR: Knowledge-And-History Aware Visual Representation for Continuous Vision-and-Language Navigation},
  author={Kong, Ping and Liu, Ruonan and Xie, Zongxia and Pang, Zhibo},
  booktitle={2025 IEEE International Conference on Robotics and Automation (ICRA)},
  pages={5236--5243},
  year={2025},
  organization={IEEE}
}

@inproceedings{zhao2024towards,
  title={Towards Coarse-grained Visual Language Navigation Task Planning Enhanced by Event Knowledge Graph},
  author={Zhao, Kaichen and Song, Yaoxian and Zhao, Haiquan and Liu, Haoyu and Li, Tiefeng and Li, Zhixu},
  booktitle={Proceedings of the 33rd ACM International Conference on Information and Knowledge Management},
  pages={3320--3330},
  year={2024}
}

@ARTICLE{tan2025source,
  author={Tan, Mingkui and Chen, Peihao and Zhi, Hongyan and Mai, Jiajie and Rosman, Benjamin and Ji, Dongyu and Zeng, Runhao},
  journal={IEEE Transactions on Multimedia}, 
  title={Source-Free Elastic Model Adaptation for Vision-and-Language Navigation}, 
  year={2025},
  volume={27},
  number={},
  pages={3953-3965},
  doi={10.1109/TMM.2025.3535356}}

@ARTICLE{xu2024hierarchical,
  author={Xu, Ming and Xie, Zilong},
  journal={IEEE Robotics and Automation Letters}, 
  title={Hierarchical Spatial Proximity Reasoning for Vision-and-Language Navigation}, 
  year={2024},
  volume={9},
  number={12},
  pages={10756-10763},
  doi={10.1109/LRA.2024.3477129}}

@inproceedings{wang2023lana,
  title={Lana: A language-capable navigator for instruction following and generation},
  author={Wang, Xiaohan and Wang, Wenguan and Shao, Jiayi and Yang, Yi},
  booktitle={Proceedings of the IEEE/CVF conference on computer vision and pattern recognition},
  pages={19048--19058},
  year={2023}
}

@inproceedings{gopinathan2024spatially,
  title={Spatially-Aware Speaker for Vision-and-Language Navigation Instruction Generation},
  author={Gopinathan, Muraleekrishna and Masek, Martin and Abu-Khalaf, Jumana and Suter, David},
  booktitle={Proceedings of the 62nd Annual Meeting of the Association for Computational Linguistics (Volume 1: Long Papers)},
  pages={13601--13614},
  year={2024}
}

@article{jiangmantis,
  title={Mantis: Interleaved Multi-Image Instruction Tuning},
  author={Jiang, Dongfu and He, Xuan and Zeng, Huaye and Wei, Cong and Ku, Max and Liu, Qian and Chen, Wenhu},
  journal={Transactions on Machine Learning Research},
  year={2024}
}

@inproceedings{wang2024lookahead,
  title={Lookahead exploration with neural radiance representation for continuous vision-language navigation},
  author={Wang, Zihan and Li, Xiangyang and Yang, Jiahao and Liu, Yeqi and Hu, Junjie and Jiang, Ming and Jiang, Shuqiang},
  booktitle={Proceedings of the IEEE/CVF conference on computer vision and pattern recognition},
  pages={13753--13762},
  year={2024}
}

@inproceedings{zhu2023vision,
  title={Vision language navigation with knowledge-driven environmental dreamer},
  author={Zhu, Fengda and Lee, Vincent CS and Chang, Xiaojun and Liang, Xiaodan},
  booktitle={International Joint Conference on Artificial Intelligence 2023},
  pages={1840--1848},
  year={2023},
  organization={Association for the Advancement of Artificial Intelligence (AAAI)}
}

@inproceedings{zhang2024navhint,
  title={Navhint: Vision and language navigation agent with a hint generator},
  author={Zhang, Yue and Guo, Quan and Kordjamshidi, Parisa},
  year={2024},
  organization={Association for Computational Linguistics}
}

@inproceedings{deguchi2024language,
  title={Language to Map: Topological map generation from natural language path instructions},
  author={Deguchi, Hideki and Shibata, Kazuki and Taguchi, Shun},
  booktitle={2024 IEEE International Conference on Robotics and Automation (ICRA)},
  pages={9556--9562},
  year={2024},
  organization={IEEE}
}

@inproceedings{zhao2024over,
  title={OVER-NAV: Elevating Iterative Vision-and-Language Navigation with Open-Vocabulary Detection and StructurEd Representation},
  author={Zhao, Ganlong and Li, Guanbin and Chen, Weikai and Yu, Yizhou},
  booktitle={Proceedings of the IEEE/CVF Conference on Computer Vision and Pattern Recognition},
  pages={16296--16306},
  year={2024}
}

@inproceedings{hong2024only,
  title={Why only text: empowering vision-and-language navigation with multi-modal prompts},
  author={Hong, Haodong and Wang, Sen and Huang, Zi and Wu, Qi and Liu, Jiajun},
  booktitle={Proceedings of the Thirty-Third International Joint Conference on Artificial Intelligence},
  pages={839--847},
  year={2024}
}

@inproceedings{xu2025flame,
  title={Flame: Learning to navigate with multimodal llm in urban environments},
  author={Xu, Yunzhe and Pan, Yiyuan and Liu, Zhe and Wang, Hesheng},
  booktitle={Proceedings of the AAAI Conference on Artificial Intelligence},
  volume={39},
  number={9},
  pages={9005--9013},
  year={2025}
}

@inproceedings{sathyamoorthy2024convoi,
  title={Convoi: Context-aware navigation using vision language models in outdoor and indoor environments},
  author={Sathyamoorthy, Adarsh Jagan and Weerakoon, Kasun and Elnoor, Mohamed and Zore, Anuj and Ichter, Brian and Xia, Fei and Tan, Jie and Yu, Wenhao and Manocha, Dinesh},
  booktitle={2024 IEEE/RSJ International Conference on Intelligent Robots and Systems (IROS)},
  pages={13837--13844},
  year={2024},
  organization={IEEE}
}

@inproceedings{wei2025ground,
  title={Ground slow, move fast: A dual-system foundation model for generalizable vision-and-language navigation},
  author={Wei, Meng and Wan, Chenyang and Peng, Jiaqi and Yu, Xiqian and Yang, Yuqiang and Feng, Delin and Cai, Wenzhe and Zhu, Chenming and Wang, Tai and Pang, Jiangmiao and others},
  booktitle = {Proceedings of the International Conference on Learning Representations (ICLR)},
  year={2026}
}

@inproceedings{zhang2024mg,
  title={MG-VLN: Benchmarking Multi-Goal and Long-Horizon Vision-Language Navigation with Language Enhanced Memory Map},
  author={Zhang, Junbo and Ma, Kaisheng},
  booktitle={2024 IEEE/RSJ International Conference on Intelligent Robots and Systems (IROS)},
  pages={7750--7757},
  year={2024},
  organization={IEEE}
}

@inproceedings{lu2025monovln,
  title={monovln: Bridging the observation gap between monocular and panoramic vision and language navigation},
  author={Lu, Renjie and Zhou, Yu and Cheng, Hao and Meng, Jingke and Zheng, Wei-Shi},
  booktitle={Proceedings of the IEEE/CVF International Conference on Computer Vision},
  pages={9477--9486},
  year={2025}
}

@inproceedings{mildenhall2020nerf,
  title     = {NeRF: Representing Scenes as Neural Radiance Fields for View Synthesis},
  author    = {Mildenhall, Ben and Srinivasan, Pratul P. and Tancik, Matthew and Barron, Jonathan T. and Ramamoorthi, Ravi and Ng, Ren},
  booktitle = {Proceedings of the European Conference on Computer Vision (ECCV)},
  year      = {2020}
}

@article{kerbl2023gaussians,
  title   = {3D Gaussian Splatting for Real-Time Radiance Field Rendering},
  author  = {Kerbl, Bernhard and Kopanas, Georgios and Leimk{\"u}hler, Thomas and Drettakis, George},
  journal = {ACM Transactions on Graphics},
  volume  = {42},
  number  = {4},
  year    = {2023}
}

@article{hu2025astranav,
  title={AstraNav-World: World Model for Foresight Control and Consistency},
  author={Hu, Junjun and Chen, Jintao and Bai, Haochen and Luo, Minghua and Xie, Shichao and Chen, Ziyi and Liu, Fei and Chu, Zedong and Xue, Xinda and Ren, Botao and others},
  journal={arXiv preprint arXiv:2512.21714},
  year={2025}
}

@inproceedings{zhou2025same,
  title={Same: Learning generic language-guided visual navigation with state-adaptive mixture of experts},
  author={Zhou, Gengze and Hong, Yicong and Wang, Zun and Zhao, Chongyang and Bansal, Mohit and Wu, Qi},
  booktitle={Proceedings of the IEEE/CVF International Conference on Computer Vision},
  pages={7794--7807},
  year={2025}
}

@inproceedings{liu2024volumetric,
  title={Volumetric environment representation for vision-language navigation},
  author={Liu, Rui and Wang, Wenguan and Yang, Yi},
  booktitle={Proceedings of the IEEE/CVF conference on computer vision and pattern recognition},
  pages={16317--16328},
  year={2024}
}

@article{zhou2025diccr,
  title={DICCR: Double-gated intervention and confounder causal reasoning for vision-language navigation},
  author={Zhou, Dongming and Deng, Jinsheng and Pang, Zhengbin and Li, Wei},
  journal={Neural Networks},
  volume={184},
  pages={107078},
  year={2025},
  publisher={Elsevier}
}

@misc{openai2022chatgpt,
  title        = {ChatGPT: Optimizing Language Models for Dialogue},
  author       = {OpenAI},
  year         = {2022},
  howpublished = {\url{https://openai.com/blog/chatgpt}}
}

@article{ouyang2022instructgpt,
  title   = {Training Language Models to Follow Instructions with Human Feedback},
  author  = {Ouyang, Long and Wu, Jeffrey and Jiang, Xu and Almeida, Diogo and others},
  journal = {Advances in Neural Information Processing Systems},
  year    = {2022}
}

@inproceedings{devlin2019bert,
  title     = {BERT: Pre-training of Deep Bidirectional Transformers for Language Understanding},
  author    = {Devlin, Jacob and Chang, Ming-Wei and Lee, Kenton and Toutanova, Kristina},
  booktitle = {Proceedings of NAACL-HLT},
  year      = {2019}
}

@inproceedings{ahn2022saycan,
  title     = {SayCan: Grounding Language in Affordances for Collaborative Task Planning},
  author    = {Ahn, Wonjoon and Sax, Alexander and Chen, Yuke and Salakhutdinov, Ruslan and Levine, Sergey},
  booktitle = {Robotics: Science and Systems (RSS)},
  year      = {2022},
  url       = {https://arxiv.org/abs/2204.01691}
}

@inproceedings{brohan2023rt2,
  title={Rt-2: Vision-language-action models transfer web knowledge to robotic control},
  author={Zitkovich, Brianna and Yu, Tianhe and Xu, Sichun and Xu, Peng and Xiao, Ted and Xia, Fei and Wu, Jialin and Wohlhart, Paul and Welker, Stefan and Wahid, Ayzaan and others},
  booktitle={Conference on Robot Learning},
  pages={2165--2183},
  year={2023},
  organization={PMLR}
}

@inproceedings{zheng2024towards,
  title={Towards learning a generalist model for embodied navigation},
  author={Zheng, Duo and Huang, Shijia and Zhao, Lin and Zhong, Yiwu and Wang, Liwei},
  booktitle={Proceedings of the IEEE/CVF Conference on Computer Vision and Pattern Recognition},
  pages={13624--13634},
  year={2024}
}

@article{chen2025constraint,
  title={Constraint-aware zero-shot vision-language navigation in continuous environments},
  author={Chen, Kehan and An, Dong and Huang, Yan and Xu, Rongtao and Su, Yifei and Ling, Yonggen and Reid, Ian and Wang, Liang},
  journal={IEEE Transactions on Pattern Analysis and Machine Intelligence},
  year={2025},
  publisher={IEEE}
}

@InProceedings{Xu_2025_ICCV,
    author    = {Xu, Peiran and Gong, Xicheng and Mu, Yadong},
    title     = {NavQ: Learning a Q-Model for Foresighted Vision-and-Language Navigation},
    booktitle = {Proceedings of the IEEE/CVF International Conference on Computer Vision (ICCV)},
    month     = {October},
    year      = {2025},
    pages     = {6327-6341}
}

@inproceedings{zhang-etal-2025-vision-language,
    title = "Vision-and-Language Navigation with Analogical Textual Descriptions in {LLM}s",
    author = "Zhang, Yue  and
      Ma, Tianyi  and
      Wang, Zun  and
      Qiao, Yanyuan  and
      Kordjamshidi, Parisa",
    editor = "Christodoulopoulos, Christos  and
      Chakraborty, Tanmoy  and
      Rose, Carolyn  and
      Peng, Violet",
    booktitle = "Proceedings of the 2025 Conference on Empirical Methods in Natural Language Processing",
    month = nov,
    year = "2025",
    address = "Suzhou, China",
    publisher = "Association for Computational Linguistics",
    doi = "10.18653/v1/2025.emnlp-main.759",
    pages = "15017--15025",
    ISBN = "979-8-89176-332-6",
}

@inproceedings{cho2014learning,
  title     = {Learning Phrase Representations using {RNN} Encoder--Decoder for Statistical Machine Translation},
  author    = {Cho, Kyunghyun and van Merri{\"e}nboer, Bart and Gulcehre, Caglar and Bahdanau, Dzmitry and Bougares, Fethi and Schwenk, Holger and Bengio, Yoshua},
  booktitle = {Proceedings of the 2014 Conference on Empirical Methods in Natural Language Processing (EMNLP)},
  year      = {2014},
  pages     = {1724--1734}
}

@inproceedings{ross2011reduction,
  title     = {A Reduction of Imitation Learning and Structured Prediction to No-Regret Online Learning},
  author    = {Ross, St{\'e}phane and Gordon, Geoffrey J. and Bagnell, J. Andrew},
  booktitle = {Proceedings of the 14th International Conference on Artificial Intelligence and Statistics (AISTATS)},
  year      = {2011},
  pages     = {627--635}
}

@misc{lin2025vlnversebenchmarkvisionlanguagenavigation,
      title={VLNVerse: A Benchmark for Vision-Language Navigation with Versatile, Embodied, Realistic Simulation and Evaluation}, 
      author={Sihao Lin and Zerui Li and Xunyi Zhao and Gengze Zhou and Liuyi Wang and Rong Wei and Rui Tang and Juncheng Li and Hanqing Wang and Jiangmiao Pang and Anton van den Hengel and Jiajun Liu and Qi Wu},
      year={2025},
      eprint={2512.19021},
      archivePrefix={arXiv},
      primaryClass={cs.CV},
      url={https://arxiv.org/abs/2512.19021}, 
}

@inproceedings{zhu2017target,
  title={Target-driven visual navigation in indoor scenes using deep reinforcement learning},
  author={Zhu, Yuke and Mottaghi, Roozbeh and Kolve, Eric and Lim, Joseph J and Gupta, Abhinav and Fei-Fei, Li and Farhadi, Ali},
  booktitle={2017 IEEE international conference on robotics and automation (ICRA)},
  pages={3357--3364},
  year={2017},
  organization={IEEE}
}

@article{chaplot2020object,
  title={Object goal navigation using goal-oriented semantic exploration},
  author={Chaplot, Devendra Singh and Gandhi, Dhiraj Prakashchand and Gupta, Abhinav and Salakhutdinov, Russ R},
  journal={Advances in Neural Information Processing Systems},
  volume={33},
  pages={4247--4258},
  year={2020}
}

@article{cui2025generating,
  title={Generating Vision-Language Navigation Instructions Incorporated Fine-Grained Alignment Annotations},
  author={Cui, Yibo and Xie, Liang and Zhao, Yu and Sun, Jiawei and Yin, Erwei},
  journal={Information Fusion},
  pages={104107},
  year={2025},
  publisher={Elsevier}
}

@article{xiong2026sfco,
  title={SFCo-Nav: Efficient Zero-Shot Visual Language Navigation via Collaboration of Slow LLM and Fast Attributed Graph Alignment},
  author={Xiong, Chaoran and Wei, Litao and Hu, Xinhao and Ma, Kehui and Xia, Ziyi and Jiang, Zixin and Sun, Zhen and Pei, Ling},
  journal={arXiv preprint arXiv:2603.01477},
  year={2026}
}

@article{xin2026agentvln,
  title={AgentVLN: Towards Agentic Vision-and-Language Navigation},
  author={Xin, Zihao and Li, Wentong and Jiang, Yixuan and Huang, Ziyuan and Wang, Bin and Li, Piji and Zhu, Jianke and Qin, Jie and Huang, Shengjun},
  journal={arXiv preprint arXiv:2603.17670},
  year={2026}
}

@article{chu2026abot,
  title={Abot-n0: Technical report on the vla foundation model for versatile embodied navigation},
  author={Chu, Zedong and Xie, Shichao and Wu, Xiaolong and Shen, Yanfen and Luo, Minghua and Wang, Zhengbo and Liu, Fei and Leng, Xiaoxu and Hu, Junjun and Yin, Mingyang and others},
  journal={arXiv preprint arXiv:2602.11598},
  year={2026}
}

@misc{cvpr2026decovln,
      title={DecoVLN: Decoupling Observation, Reasoning, and Correction for Vision-and-Language Navigation}, 
      author={Zihao Xin and Wentong Li and Yixuan Jiang and Bin Wang and Runming Cong and Jie Qin and Shengjun Huang},
      year={2026},
      eprint={2603.13133},
      archivePrefix={arXiv},
      primaryClass={cs.RO},
      url={https://arxiv.org/abs/2603.13133}, 
}

@article{zhang2024vision,
  title={Vision-and-Language Navigation Today and Tomorrow: A Survey in the Era of Foundation Models},
  author={Zhang, Yue and Ma, Ziqiao and Li, Jialu and Qiao, Yanyuan and Wang, Zun and Chai, Joyce and Wu, Qi and Bansal, Mohit and Kordjamshidi, Parisa},
  journal={Transactions on Machine Learning Research},
  year={2024}
}

@article{gao2024vision,
  title={Vision-Language Navigation with Embodied Intelligence: A Survey},
  author={Gao, Peng and Wang, Peng and Gao, Feng and Wang, Fei and Yuan, Ruyue},
  journal={arXiv preprint arXiv:2402.14304},
  year={2024}
}

@inproceedings{wang2025safety,
  title={Safety of embodied navigation: a survey},
  author={Wang, Zixia and Hu, Jia and Mu, Ronghui},
  booktitle={Proceedings of the Thirty-Fourth International Joint Conference on Artificial Intelligence},
  pages={10714--10722},
  year={2025}
}

@article{yao2025aeroverse,
  title={AeroVerse-Review: Comprehensive survey on aerial embodied vision-and-language navigation},
  author={Yao, Fanglong and Liu, Youzhi and Zhang, Wenyi and Zhu, Zhengqiu and Li, Chenglong and Liu, Nayu and Hu, Peng and Yue, Yuanchang and Wei, Kaiwen and He, Xin and others},
  journal={The Innovation Informatics},
  volume={1},
  number={1},
  pages={100015--1},
  year={2025},
  publisher={The Innovation Informatics}
}

@article{kawaharazuka2025vla,
  title={Vision-Language-Action Models for Robotics: A Review Towards Real-World Applications},
  author={Kawaharazuka, Kento and Oh, Jihoon and Yamada, Jun and Posner, Ingmar and Zhu, Yuke},
  journal={IEEE Access},
  volume={13},
  pages={162467--162504},
  year={2025},
  doi={10.1109/ACCESS.2025.3609980}
}

@article{ma2026survey,
  title={A Survey on Vision--Language--Action Models for Embodied AI},
  author={Ma, Yueen and Song, Zixing and Zhuang, Yuzheng and Hao, Jianye and King, Irwin},
  journal={IEEE Transactions on Neural Networks and Learning Systems},
  year={2026},
  publisher={IEEE}
}

@article{lin2023advances,
  title={Advances in embodied navigation using large language models: A survey},
  author={Lin, Jinzhou and Gao, Han and Feng, Xuxiang and Xu, Rongtao and Wang, Changwei and Zhang, Man and Guo, Li and Xu, Shibiao},
  journal={arXiv preprint arXiv:2311.00530},
  year={2023}
}

@inproceedings{Jeong_2025_BMVC,
  author    = {SeongJun Jeong and Gi-Cheon Kang and Seongho Choi and Joochan Kim and Byoung-Tak Zhang},
  title     = {Continual Vision-and-Language Navigation},
  booktitle = {36th British Machine Vision Conference 2025, {BMVC} 2025, Sheffield, UK, November 24-27, 2025},
  publisher = {BMVA},
  year      = {2025},
  url       = {https://bmva-archive.org.uk/bmvc/2025/assets/papers/Paper_84/paper.pdf}
}

@article{li2024vision,
  title   = {Vision-Language Navigation with Continual Learning},
  author  = {Li, Zhiyuan and Lv, Yanfeng and Tu, Ziqin and Shang, Di and Qiao, Hong},
  journal = {arXiv preprint arXiv:2409.02561},
  year    = {2024}
}

@inproceedings{wang2026allday,
  title     = {All-day Multi-scenes Lifelong Vision-and-Language Navigation with Tucker Adaptation},
  author    = {Wang, Xudong and Li, Gan and Liu, Zhiyu and Wang, Yao and Liu, Lianqing and Han, Zhi},
  booktitle = {International Conference on Learning Representations},
  year      = {2026}
}

@inproceedings{jiang2026m3e,
  title     = {M$^3$E: Continual Vision-and-Language Navigation via Mixture of Macro and Micro Experts},
  author    = {Jiang, Yongliang and Zhang, Huaidong and Luo, Xuandi and He, Shengfeng},
  booktitle = {International Conference on Learning Representations},
  year      = {2026}
}

@article{zhang2026qwen,
  title={Qwen-RobotNav Technical Report: A Scalable Navigation Model Designed for an Agentic Navigation System},
  author={Zhang, Jiazhao and Zhou, Gengze and Yin, Hale and Huang, Yiyang and Lei, Zixing and Peng, Qihang and Yuan, Haoqi and Zhang, Jie and Guo, Xudong and Chen, Xiaoyue and others},
  journal={arXiv preprint arXiv:2606.18112},
  year={2026}
}

\end{document}